\documentclass{article}
\pdfoutput=1 

\PassOptionsToPackage{numbers,sort&compress}{natbib}
\usepackage[preprint]{neurips_2026}
\usepackage[utf8]{inputenc}
\usepackage[T1]{fontenc}
\usepackage{hyperref}
\usepackage{url}
\usepackage{booktabs}
\usepackage{amsfonts}
\usepackage{amsmath}
\usepackage{amssymb}
\usepackage{nicefrac}
\usepackage{microtype}
\usepackage{xcolor}
\usepackage{graphicx}
\usepackage{longtable}
\usepackage{array}
\usepackage{tabularx}
\usepackage{caption}
\usepackage{multirow}
\makeatletter
\renewcommand{\@noticestring}{}
\renewcommand\section{\@startsection{section}{1}{\z@}{-2.0ex plus -0.25ex minus -0.2ex}{1.0ex plus 0.15ex}{\normalfont\Large\bfseries}}
\renewcommand\subsection{\@startsection{subsection}{2}{\z@}{-1.5ex plus -0.2ex minus -0.15ex}{0.8ex plus 0.1ex}{\normalfont\large\bfseries}}
\renewcommand\subsubsection{\@startsection{subsubsection}{3}{\z@}{-1.15ex plus -0.15ex minus -0.1ex}{0.65ex plus 0.1ex}{\normalfont\normalsize\bfseries}}
\makeatother
\renewcommand{\arraystretch}{1.08}
\title{WuYu-EnvLE-Bench: A Benchmark for Evaluating Large Language Models in Environmental Law Enforcement}

\author{%
  \makebox[\textwidth][c]{%
    Ziliang Yang\textsuperscript{1,2,*}
    \quad
    Yi Zhang\textsuperscript{3,*}
    \quad
    Kaijun Lin\textsuperscript{3}
    \quad
    Jiachao Ke\textsuperscript{3}
    \quad
    Haihong Xu\textsuperscript{4,\(\dagger\)}
    \quad
    Zongguo Wen\textsuperscript{3,\(\dagger\)}
  }
  \\[0.7em]
  \makebox[\textwidth][c]{%
    \normalfont\rmfamily\fontsize{10}{11}\selectfont
    \textsuperscript{1}\,
    School of Environment, Tsinghua University
  }
  \\
  \makebox[\textwidth][c]{%
    \normalfont\rmfamily\fontsize{10}{11}\selectfont
    \textsuperscript{2}\,
    College of Economics and Management,
    Beijing University of Technology
  }
  \\
  \makebox[\textwidth][c]{%
    \normalfont\rmfamily\fontsize{10}{11}\selectfont
    \textsuperscript{3}\,
    State Key Laboratory of Iron and Steel Industry Environmental Protection,
    School of Environment, Tsinghua University
  }
  \\
  \makebox[\textwidth][c]{%
    \normalfont\rmfamily\fontsize{10}{11}\selectfont
    \textsuperscript{4}\,
    Appraisal Center for Environmental Engineering,
    Ministry of Ecology and Environment
  }
  \\[0.45em]
  \makebox[\textwidth][c]{%
    \normalfont\ttfamily\fontsize{9}{11}\selectfont
    wenzg@tsinghua.edu.cn
  }
  \\[0.3em]
  \makebox[\textwidth][c]{%
    \normalfont\rmfamily\fontsize{9}{11}\selectfont
    \textsuperscript{*}Equal contribution.
    \qquad
    \textsuperscript{\(\dagger\)}Corresponding authors.
  }
}

\begin{document}
\maketitle

\begin{abstract}
Large language models (LLMs) are increasingly considered for environmental enforcement, but their ability to produce traceable enforcement decisions remains unclear. We introduce WuYu-EnvLE-Bench, a benchmark built from real enforcement cases, regulatory standards, and practitioner review. It contains 2,521 benchmark instances, 14 tasks, and 12 pollution-medium subdomains across pre-enforcement, in-enforcement, and post-enforcement workflows. Using Absolute Environmental Enforcement Score (AES) and Intelligent Enforcement Index (IEI), we evaluate open-source and closed-source LLMs across capability, response quality, and resource efficiency. Results show that LLMs perform well on rule-bounded tasks but remain unreliable in evidence-chain construction, contradiction detection, multi-source integration, and procedural judgment. Model scaling also shows diminishing returns: medium-sized models approach leading models in structured tasks, while larger models do not reliably overcome evidence-reasoning bottlenecks. WuYu-EnvLE-Bench highlights the need for evidence-grounded, rule-aware, and task-adaptive enforcement reasoning.
\end{abstract}

\section{Introduction}

Environmental law enforcement is a high-stakes regulatory process that converts complex environmental incidents into executable, traceable, and reviewable administrative decisions \citep{gray20115}. Unlike general question answering or isolated legal reasoning, environmental enforcement requires the joint interpretation of pollution facts, monitoring records, emission standards, evidence materials, statutory rules, and procedural requirements. Enforcement officers must identify risks, plan inspections, collect and fix evidence, determine violations, apply legal provisions, and exercise enforcement discretion \citep{oecd202417}. This makes environmental enforcement a technically grounded and legally constrained decision-making problem, rather than a pure text understanding task \citep{hurst202110}.

Large language models have shown strong potential in knowledge-intensive and reasoning-intensive domains, including legal analysis, scientific question answering, and decision support \citep{guha20233,wang202420,phan202618}. In principle, this makes LLMs attractive candidates for environmental enforcement assistance: they can extract facts from citizen complaints, summarize monitoring reports, match case facts with legal provisions, generate inspection plans, and draft enforcement documents. However, environmental enforcement requires models not only produce plausible text, but also maintain factual consistency, identify legally relevant evidence, follow procedural logic, and generate outputs that can support accountable administrative decisions \citep{magesh202514}. Whether existing LLMs satisfy these requirements remains largely untested, as existing environmental LLM evaluations mainly focus on environmental science knowledge rather than the full lifecycle of environmental enforcement tasks \citep{he20258,zhang202523}.

This gap is especially important because environmental enforcement is a lifecycle process \citep{gunningham20116}. Pre-enforcement tasks focus on risk discovery, complaint verification, public-opinion monitoring, and non-site warning. In-enforcement tasks involve inspection task planning, multi-evidence extraction, and inquiry analysis. Post-enforcement tasks require enforcement decision, justification assessment and result reasoning \citep{gray20115}. These stages differ substantially in information availability, reasoning structure, and operational objective. A model with high average performance may still be unsuitable for deployment if it fails in evidence-sensitive or procedure-sensitive stages. Therefore, the first question we study is:

\par\noindent\textbf{RQ1: What are the strengths and weaknesses of LLMs across the full lifecycle of environmental law enforcement?}\par

A second challenge concerns deployability. Recent LLM research often assumes that larger models provide stronger task performance \citep{xiao202521}, yet environmental enforcement is usually deployed under resource constraints, especially in local enforcement agencies, mobile inspection systems, and long-running regulatory platforms. In this setting, the most capable model is not necessarily the most useful one. If medium-sized open-source models can approach large closed-source models on structured enforcement tasks, then further scaling may bring limited practical benefit relative to its cost. This motivates the second research question:

\par\noindent\textbf{RQ2: Does environmental law enforcement exhibit diminishing returns from model scaling?}\par

To address these questions, we introduce WuYu-EnvLE-Bench, a benchmark for evaluating LLMs in environmental enforcement. It organizes 2,521 benchmark instances derived from real environmental enforcement cases into a two-dimensional framework of enforcement stage and pollution medium, covering 14 tasks and 12 pollution-medium subdomains. We evaluate models of different scales, sources, and reasoning modes, and propose AES and IEI to measure both enforcement capability and deployment-oriented value. The results reveal two main findings. First, current LLMs can support rule-bounded tasks with explicit facts and standardized outputs, but remain unreliable in evidence-chain reasoning, contradiction verification, multi-source integration, and procedural judgment. Second, model scaling shows clear diminishing returns: medium-sized models can approach leading models in structured tasks, whereas larger models do not necessarily overcome evidence-reasoning bottlenecks. These findings suggest that environmental enforcement LLMs should prioritize evidence-grounded, rule-aware, and task-adaptive reasoning over parameter scaling alone. Figure 1 summarizes the benchmark framework.

\par\noindent\begin{minipage}{\linewidth}
  \centering
  \includegraphics[width=\linewidth,height=0.72\textheight,keepaspectratio]{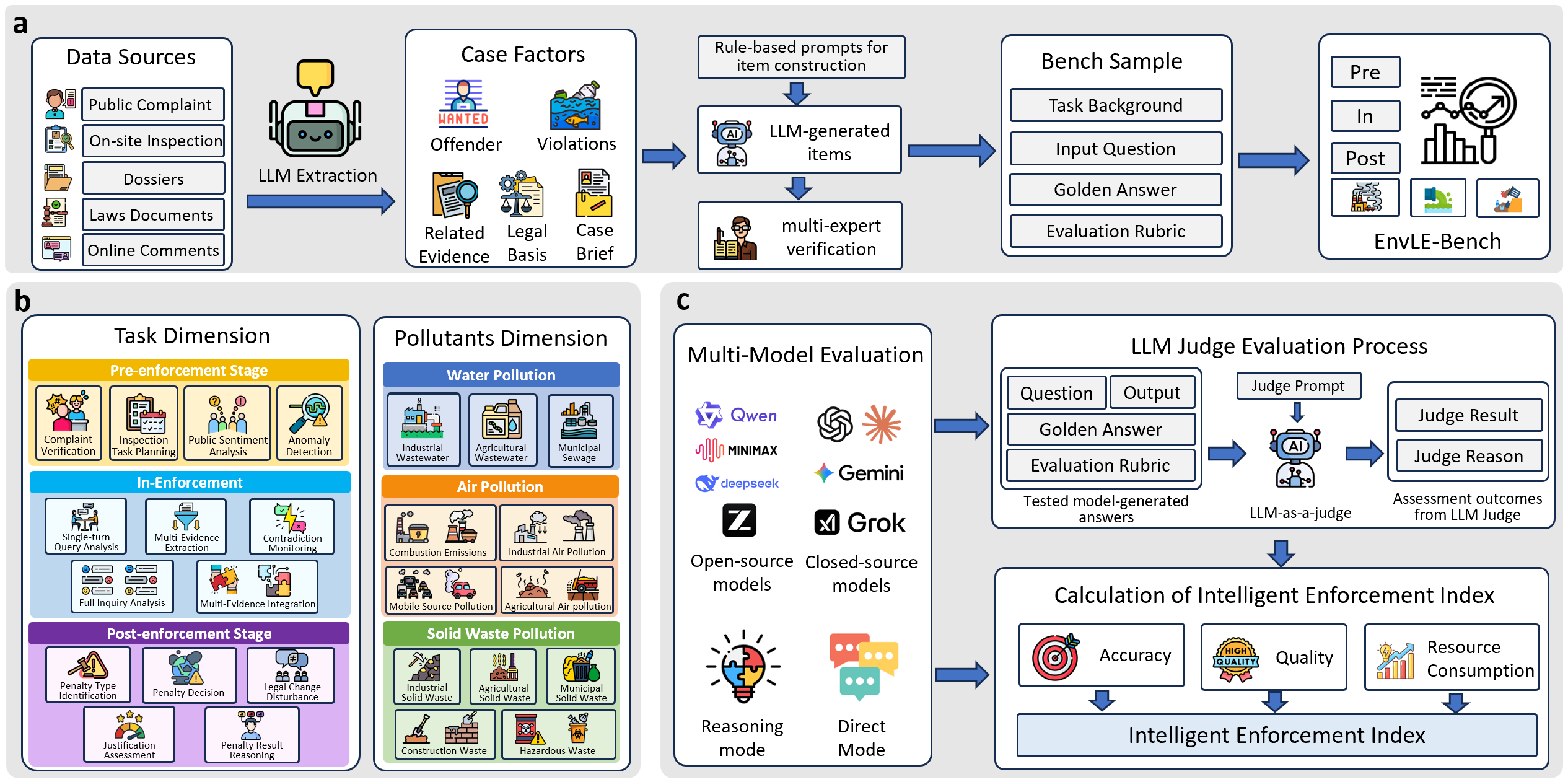}
  \captionof{figure}{Overview of WuYu-EnvLE-Bench. Panel (a) shows the data-to-benchmark pipeline with expert validation, panel (b) maps the task-pollutant matrix across enforcement stages and domains, and panel (c) illustrates the multi-model evaluation and the Intelligent Enforcement Index derived from accuracy, quality, and resource consumption.}
  \label{fig:main-1-overview-of-wuyu-envle-bench-panel-a-shows-the-data-t}
\end{minipage}\par\vspace{0.65\baselineskip}

\section{Related works}

\subsection{Legal Domain Benchmarks}

General legal benchmarks mainly aim to evaluate large language models' mastery of general legal knowledge and foundational legal reasoning abilities. \citet{guha20233} developed LEGALBENCH, a landmark collaborative framework consisting of 162 tasks organized around diverse forms of legal reasoning, thereby providing a systematic tool for measuring LLM performance across different types of legal reasoning. Meanwhile, \citet{fei20242} proposed LawBench, which evaluates legal capabilities from the perspective of cognitive psychology by decomposing legal competence into three levels: memorization, understanding, and application. Through 20 diverse tasks, LawBench assesses whether models possess reliable legal knowledge and whether they can apply such knowledge in realistic legal scenarios. Further, \citet{li202411} introduced LexEval, which proposes a more fine-grained six-dimensional legal cognitive taxonomy, namely the LexAbility Taxonomy, covering memorization, understanding, logic inference, discrimination, generation, and ethics. Based on this taxonomy, LexEval constructs a large-scale Chinese legal evaluation dataset containing 14,150 questions.

Task-specific legal benchmarks further examine particular mechanisms of complex legal reasoning. \citet{zhang202422} focused on the presumption of innocence in criminal judgment prediction. In LJPIV, they construct an adversarial dataset with innocent verdict labels based on the tripartite reasoning framework of criminal law, namely elements of the offense, unlawfulness, and culpability, thereby improving models' ability to handle exculpatory reasoning. To address the dynamic evolution of legal systems, \citet{han20267} introduced LawShift, a benchmark for legal judgment prediction under statutory revisions. By adopting a metamorphic testing framework, LawShift evaluates whether models can adapt their reasoning logic to legal updates across 31 fine-grained types of statutory changes. In addition, \citet{li202512} explored agent-based applications in legal scenarios through LegalAgentBench. By providing 37 external knowledge interaction tools and 300 complex tasks, LegalAgentBench evaluates the comprehensive ability of models to perform multi-step retrieval, tool use, and complex legal document drafting.

Taken together, these studies provide important tools for evaluating LLMs' foundational legal understanding. However, their primary focus remains on legal text comprehension, general statutory application, and judicial reasoning. In environmental enforcement, models must not only understand legal norms, but also transform pollution facts, monitoring data, emission standards, and evidence materials into executable enforcement judgments. Therefore, relying solely on existing legal benchmarks is insufficient for evaluating models' strengths and weaknesses across the full lifecycle of environmental enforcement. A dedicated and comprehensive benchmark for environmental enforcement is thus needed to systematically diagnose the capability boundaries of LLMs in this domain.

\subsection{Environmental Domain Benchmarks}

In terms of compliance assessment and practical regulatory review, existing benchmarks have provided preliminary explorations of environmental legal applications. \citet{wang202519} proposed HSE-Bench, which focuses on health, safety, and environment compliance assessment. This benchmark also integrates an IRAC-based reasoning flow, and its data sources include regulatory texts, court cases, professional examinations, and field monitoring videos, aiming to simulate realistic compliance decision-making processes. However, HSE-Bench differs from precise environmental enforcement in that it mainly emphasizes general HSE behavioral compliance, such as personal protective equipment use and workplace safety, rather than quantitative enforcement centered on pollutant discharge standards and exceedance determination. For the processing of procedural legal documents, \citet{meyur202415} constructed NEPAQuAD, the first comprehensive question-answering dataset based on Environmental Impact Statement documents under the U.S. National Environmental Policy Act. Using the MAPLE evaluation framework, they tested model performance in long-document processing and regulatory reasoning for environmental review.

In terms of knowledge reserves and interdisciplinary challenges, existing studies have established relatively comprehensive environmental science evaluation systems. \citet{guo20254} developed the ELLE benchmark, which contains 1,130 question-answer pairs covering 16 subfields, including environmental chemistry, toxicology, and management. It evaluates models' ability to generate environmental solutions from the dimensions of professionalism, clarity, and feasibility. To address interdisciplinary challenges, \citet{zhang202523} constructed EnvBench, consisting of 4,998 tasks, and ChatEnv, an instruction dataset containing approximately 112 million characters. Based on these resources, they developed EnvGPT through instruction fine-tuning, significantly improving model performance in analysis, reasoning, and calculation across five core environmental themes, including climate change and water resource management. In addition, \citet{he20258} introduced ESGenius, a benchmark for evaluating LLMs on environmental, social, and governance (ESG) and sustainability knowledge. ESGenius combines expert-validated multiple-choice questions with a curated corpus of authoritative ESG frameworks, standards, reports, and recommendation documents, thereby extending environmental-domain evaluation to sustainability-oriented knowledge and source-grounded question answering.

These environmental science benchmarks mainly focus on models' mastery, explanation, and solution-generation abilities regarding environmental knowledge, providing important references for understanding the knowledge foundation of LLMs in the environmental domain. However, environmental enforcement requires models to embed professional knowledge into administrative enforcement procedures and complete the transformation from scientific facts to legal judgments. Existing environmental science benchmarks have not sufficiently examined these enforcement-critical capabilities, nor have they evaluated the relationship between model capability and deployment resources. Therefore, it is necessary to construct a dedicated environmental enforcement benchmark that simultaneously evaluates technical fact understanding, legal rule application, evidence-chain reasoning, and resource efficiency. Such a benchmark can provide a systematic basis for determining the applicability boundaries of LLMs in real-world enforcement scenarios.

\section{WuYu-EnvLE-Bench}

\subsection{WuYu-EnvLE-Bench construction}

WuYu-EnvLE-Bench is designed to evaluate whether large language models can transform pollution facts, evidentiary materials, regulatory standards, and administrative procedures into reliable environmental enforcement judgments. Environmental enforcement requires models to operate within a practical regulatory workflow, where each conclusion must be supported by identifiable facts, legally relevant evidence, applicable rules, and procedurally executable reasoning \citep{ministryofecologyandenvironmentofthepeoplesrepublicofchina202316}.

The benchmark was constructed from real environmental enforcement materials, including citizen complaint records, on-site inspection records, administrative penalty decisions, monitoring reports, photographs, inquiry transcripts, and relevant laws, standards, and regulations. These materials cover major enforcement scenarios involving water pollution, air pollution, and solid waste pollution, and preserve the enforcement chain from risk discovery to case investigation, violation determination and penalty decision-making. All case materials were anonymized and reorganized before being used for task construction.

The construction process follows a case-to-task pipeline. First, core enforcement facts were extracted from raw materials to form standardized case briefs, including the regulated entity, violation behavior, time and location, pollution medium, evidentiary materials, monitoring results, legal basis, and enforcement outcome. Second, task-specific samples were generated from these case briefs according to different enforcement requirements, such as complaint validity screening, inspection task planning, evidence extraction, evidence-chain construction, legal applicability analysis, penalty-type classification and penalty result reasoning. Third, each sample was paired with a gold standard and an evaluation rubric. The gold standard defines the expected enforcement judgment or reference response, while the rubric specifies the key facts, evidentiary elements, legal reasoning requirements, and output constraints used for scoring. Finally, all samples, gold standards, and rubrics were manually reviewed to ensure factual consistency, legal applicability, procedural reasonableness, and scoring completeness.

Each benchmark instance therefore consists of four components: a task input, a case brief, a gold standard, and an evaluation rubric. During evaluation, the candidate model response is assessed by an LLM-as-a-Judge framework \citep{phan202618}. For open-ended tasks, the judge receives the task input, candidate response, case brief, gold standard, and evaluation rubric, and then produces a structured evaluation result. The unified judge prompt, adjudication procedure, and output schema are provided in Appendix C.3.2. The evaluation focuses on four core dimensions: conclusion correctness, key information coverage, professional reasoning quality, and enforcement practicality. This design allows WuYu-EnvLE-Bench to evaluate not only whether a model reaches the correct conclusion, but also whether its reasoning is supported by evidence, consistent with enforcement logic, and usable in regulatory practice.

This construction and evaluation design enables WuYu-EnvLE-Bench to preserve the authenticity of real enforcement cases while supporting standardized and scalable model assessment. More importantly, it avoids judging model outputs only by surface-level textual similarity, and instead evaluates whether models can identify enforcement-relevant facts, apply legal rules, construct valid evidentiary relations, and generate outputs that support subsequent regulatory action.

\subsection{Task taxonomy and data statistics}

WuYu-EnvLE-Bench is organized along two dimensions: enforcement stage and pollution medium. The enforcement-stage dimension reflects the procedural structure of environmental enforcement, covering pre-enforcement, in-enforcement, and post-enforcement tasks. The pollution-medium dimension captures the professional heterogeneity of regulated objects, covering water pollution, air pollution, and solid waste pollution. Together, these two dimensions form a task--domain matrix that enables the benchmark to evaluate both lifecycle-specific capability and cross-domain generalization.

Along the enforcement-stage dimension, pre-enforcement tasks mainly involve risk discovery and clue verification, such as complaint screening, public-opinion monitoring, and early warning. In-enforcement tasks focus on inspection, evidence extraction, evidence-chain construction, violation identification, and legal applicability analysis. Post-enforcement tasks involve penalty discretion, legal-change analysis and penalty result reasoning. This structure allows the benchmark to diagnose whether model performance varies across different stages of the enforcement workflow.

Along the pollution-medium dimension, water, air, and solid waste cases represent different evidentiary and regulatory structures. Water pollution tasks emphasize discharge pathways, receiving water bodies, and monitoring indicators; air pollution tasks involve production processes, emission conditions, facility operation, and diffusion contexts; solid waste tasks focus on material attributes, storage, transfer, disposal, and responsibility tracing. This dimension allows WuYu-EnvLE-Bench to test whether models can generalize across heterogeneous environmental enforcement scenarios rather than only perform well in a single pollution medium.

WuYu-EnvLE-Bench contains 2,521 environmental enforcement cases, covering 14 representative task types and 12 pollution-medium subdomains. As shown in Figure 2 panel (a), the task distribution covers the full enforcement lifecycle, with pre-enforcement, in-enforcement, and post-enforcement tasks accounting for 29.3\%, 35.6\%, and 35.1\% of the benchmark, respectively. As shown in Figure 2 panel (b), the pollution-media distribution includes water pollution, air pollution, and solid waste pollution, accounting for 26.4\%, 32.7\%, and 40.9\%, respectively. The outer rings further divide these categories into specific task types and pollution-medium subdomains, ensuring broad coverage of real-world enforcement scenarios. Detailed task design rules are provided in Appendix B.

\par\noindent\begin{minipage}{\linewidth}
  \centering
  \includegraphics[width=\linewidth,height=0.80\textheight,keepaspectratio]{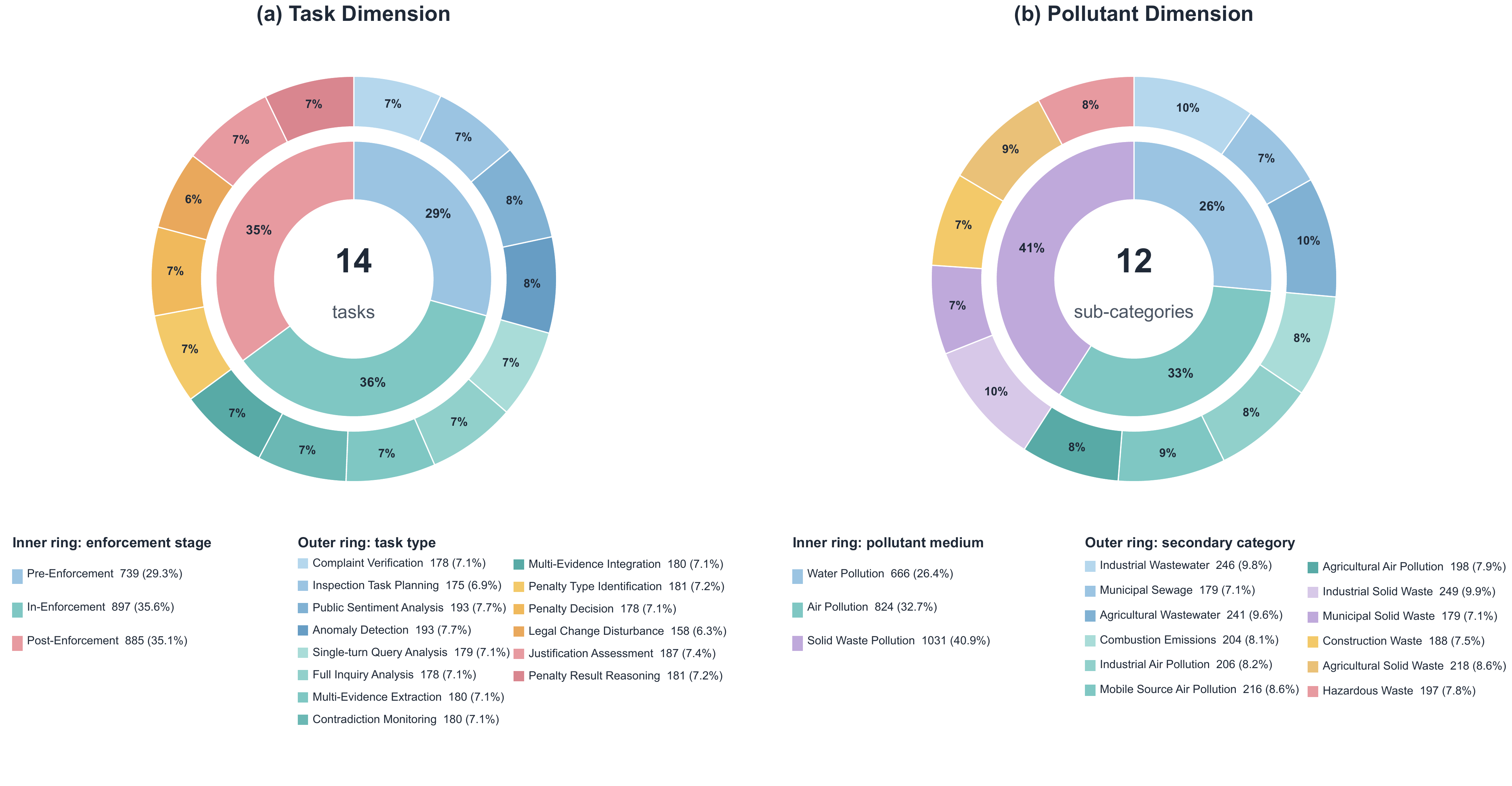}
  \captionof{figure}{Data statistics of WuYu-EnvLE-Bench. Panel (a) summarizes the distribution across pre-enforcement, in-enforcement, and post-enforcement stages and task types; panel (b) summarizes the distribution across water pollution, air pollution, and solid waste pollution subcategories.}
  \label{fig:main-3-data-statistics-of-wuyu-envle-bench-panel-a-summarize}
\end{minipage}\par\vspace{0.65\baselineskip}

Overall, this two-dimensional design allows WuYu-EnvLE-Bench to evaluate LLMs from both procedural and professional perspectives. The enforcement-stage dimension supports the analysis of model strengths and weaknesses across the regulatory lifecycle, while the pollution-medium dimension tests whether these capabilities remain stable across different environmental objects and evidentiary systems.

\subsection{Evaluation metrics}

Existing evaluations of large language models commonly rely on correctness-oriented metrics, such as average accuracy in multitask knowledge benchmarks and pass rates in code-generation benchmarks, to summarize model capability \citep{hendrycks20219}. However, in real-world environmental enforcement scenarios, task correctness alone is insufficient to fully reflect the practical value of a model \citep{liang202313}. On the one hand, different models may achieve similar task performance while differing substantially in parameter scale, inference cost, and deployment resource requirements \citep{xiao202521}. On the other hand, environmental enforcement requires models not only to provide correct answers, but also to present reliable factual support, clear reasoning processes, and professional expressions consistent with enforcement norms \citep{curtis20201}. Therefore, raw task-level metrics cannot be directly compared across tasks. To enable unified capability assessment, each task is first converted into a normalized Task Score according to its task-specific evaluation protocol:

\[
\mathrm{TaskScore}_{m,t}\in[0,1].
\]

where $\mathrm{TaskScore}_{m,t}$ denotes the normalized score of model $m$ on task $t$. For rule-based decision tasks, the task score is derived from answer correctness or step-level judgment consistency. For open-ended enforcement tasks, the task score is obtained through the LLM-as-a-Judge framework described above. Detailed task-specific scoring rules are provided in Appendix C.

To measure overall environmental enforcement capability, we define the Absolute Environmental Enforcement Score (AES). Since WuYu-EnvLE-Bench covers multiple enforcement stages with different operational functions, we first calculate the average task score within each stage:

\[
S_{m,g}=\frac{1}{|\mathcal{T}_{g}|}\sum_{t\in\mathcal{T}_{g}}\mathrm{TaskScore}_{m,t}.
\]

where $\mathcal{T}_{g}$ denotes the set of tasks belonging to enforcement stage $g$. The stage-level scores are then aggregated with equal weights:

\[
\widetilde{\mathrm{AES}}_{m}=\frac{1}{n}\sum_{g=1}^{n}S_{m,g}.
\]

Here, $n$ denotes the number of enforcement stages. In this study, $n=3$, corresponding to pre-enforcement, in-enforcement, and post-enforcement. The reported AES is obtained by mapping from the normalized capability space to a 0--100 scale for readability. A higher AES indicates stronger overall capability across the full environmental enforcement lifecycle.

However, task performance alone is insufficient for evaluating the practical value of a model in real enforcement scenarios. Two models may achieve similar AES but differ substantially in response quality, parameter scale, inference cost, and deployment feasibility. Environmental enforcement also requires model outputs to be not only correct, but also reliable, explainable, and professionally usable. Therefore, we further define the Intelligent Enforcement Index (IEI), which integrates enforcement capability, response quality, and resource efficiency.

Specifically, we first define a capability base:

\[
\mathrm{CB}_{m}=w_{a}\cdot\widetilde{\mathrm{AES}}_{m}+w_{q}\cdot Q_{m}.
\]

where $Q_{m}$ denotes the model-level Quality Factor, which summarizes the reliability, explainability, and professionalism of model outputs. The parameters $w_{a}$ and $w_{q}$ represent the weights assigned to task capability and response quality. In this study, we adopt an equal-weight setting, assigning the same importance to task performance and response quality.

The Intelligent Enforcement Index is then defined as:

\[
\mathrm{IEI}_{m}=\mathrm{CB}_{m}\cdot\mathrm{RE}_{m}.
\]

where $\mathrm{RE}_{m}$ denotes the model-level Resource Efficiency. For open-source models, resource efficiency is estimated based on parameter scale; for closed-source models, it is estimated based on inference cost during evaluation. Resource efficiency is used as an adjustment factor rather than the primary determinant of model capability.

AES and IEI serve different analytical purposes. AES directly measures a model's environmental enforcement capability and is therefore used for cross-model, cross-stage, and cross-task performance analysis. IEI further incorporates response quality and resource burden, and is used to examine deployment-oriented value and the marginal effect of model scaling. Since open-source and closed-source models use different resource proxies, IEI is mainly interpreted within each model category rather than as a universal ranking across all models. The detailed calculation of Task Scores, Quality Factor, Resource Efficiency, and IEI is provided in Appendix C.

\section{Results}

\subsection{Overall Performance}

To obtain an overall view of LLM capability and deployment value in environmental enforcement, we evaluate models from two complementary perspectives. AES is used to measure absolute enforcement capability across stages and pollution media, while IEI further incorporates resource efficiency to assess whether higher capability can be achieved with reasonable deployment cost. The evaluated models are grouped into closed-source LLMs, open-source LLMs, and open-source small language models (SLMs), allowing comparison of capability ceilings, open-source competitiveness, and scale-related efficiency.

\par\noindent\begin{minipage}{\linewidth}
  \centering
  \includegraphics[width=0.85\linewidth,height=0.612\textheight,keepaspectratio]{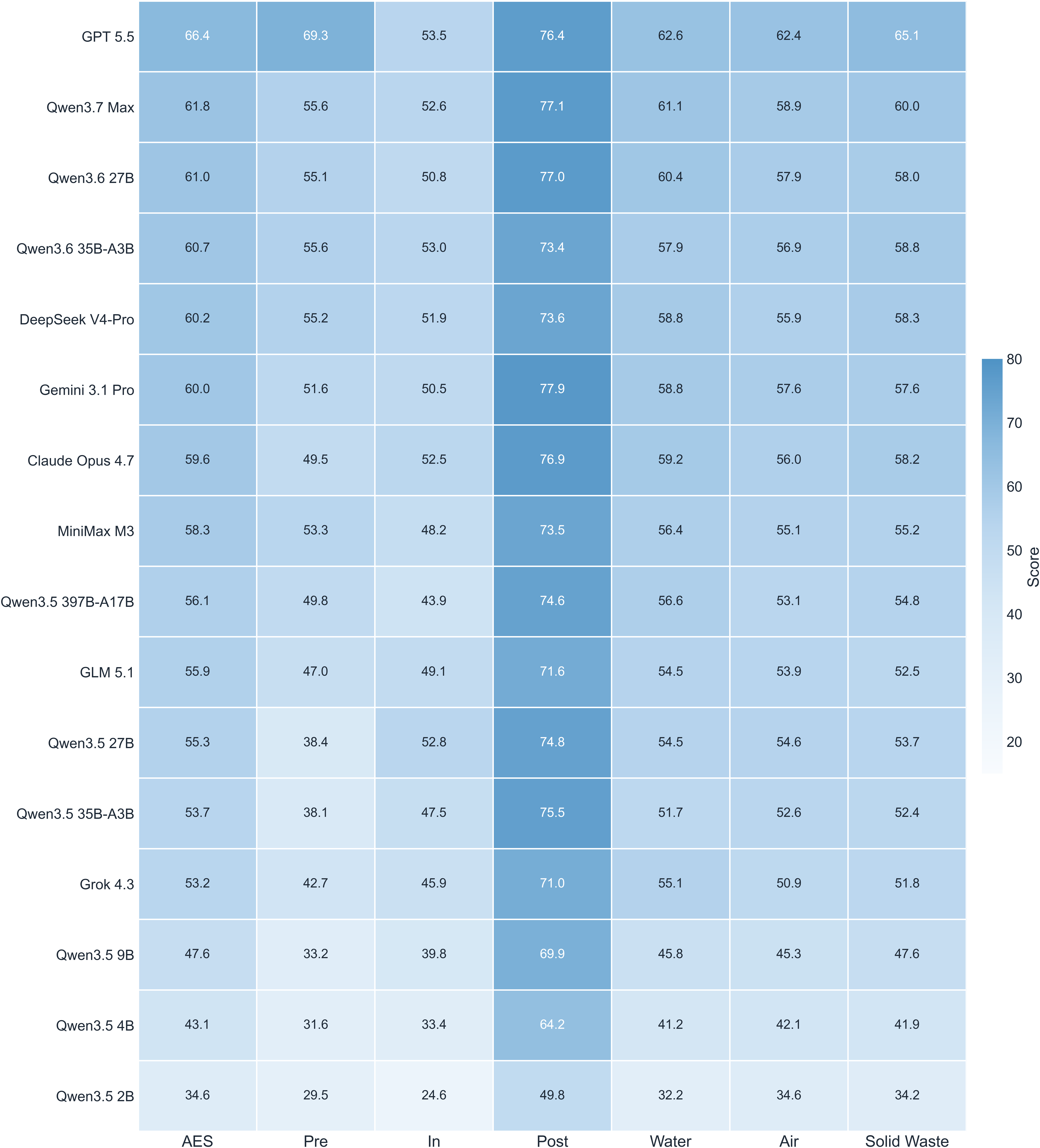}
  \captionof{figure}{Overall AES and dimension-level performance across models, enforcement stages, and pollution media.}
  \label{fig:main-4-overall-absolute-environmental-enforcement-score-aes-}
\end{minipage}\par\vspace{0.65\baselineskip}

Figure 3 shows that current LLMs have developed considerable environmental enforcement capability, but their performance is clearly uneven across enforcement stages. Most models achieve their highest scores in the post-enforcement stage and lower scores in the in-enforcement stage. For example, GPT-5.5 scores 69.3, 53.5, and 76.4 in the pre-enforcement, in-enforcement, and post-enforcement stages, respectively, while Qwen3.6-27B scores 55.1, 50.8, and 77.0. In contrast, differences across pollution media are relatively limited. GPT-5.5 obtains 62.6, 62.4, and 65.1 in water, air, and solid waste domains, while Qwen3.6-27B obtains 60.4, 57.9, and 58.0. This suggests that model performance is shaped more by task structure and reasoning requirements than by whether the case concerns water pollution, air pollution, or solid waste pollution. These results further show that closed-source LLMs still provide the highest capability ceiling, with GPT-5.5 achieving the highest AES, followed by Qwen3.7 Max. However, several open-source LLMs have approached this level: Qwen3.6-27B, Qwen3.6-35B-A3B, and DeepSeek V4-Pro reach AES scores of 61.0, 60.7, and 60.2, respectively. By contrast, open-source SLMs remain clearly limited, with Qwen3.5-2B, Qwen3.5-4B, and Qwen3.5-9B scoring 34.6, 43.1, and 47.6. The IEI results in Figure 4 provide direct evidence that absolute enforcement capability and deployment-oriented value begin to diverge once resource burden is taken into account. Among open-source models, Qwen3.6-27B and Qwen3.6-35B-A3B achieve the highest IEI scores of 56.22 and 54.30, respectively, exceeding the open-source mean of 48.20 by 8.02 and 6.10 points. By contrast, the substantially larger Qwen3.5-397B-A17B and DeepSeek V4-Pro achieve IEI scores of only 49.16 and 48.85, although their AES values remain within the competitive range. The pattern is therefore not that smaller models are universally preferable. Rather, IEI increases markedly from the 2B–9B range to the 27B–35B range, after which further parameter expansion produces little or even negative deployment return. A similar divergence is observed among closed-source models. Grok 4.3 and Qwen3.7 Max achieve the highest IEI scores of 42.13 and 41.54, whereas GPT-5.5 reaches only 34.21 despite having the largest AES bubble. Its absolute capability advantage is therefore insufficient to offset its substantially higher inference cost. These results indicate that IEI identifies a capability–resource frontier rather than reproducing a conventional capability ranking: models gain high IEI scores only when additional capability is sufficiently large to justify the additional deployment burden.

\par\noindent\begin{minipage}{\linewidth}
  \centering
  \includegraphics[width=0.85\linewidth,height=0.612\textheight,keepaspectratio]{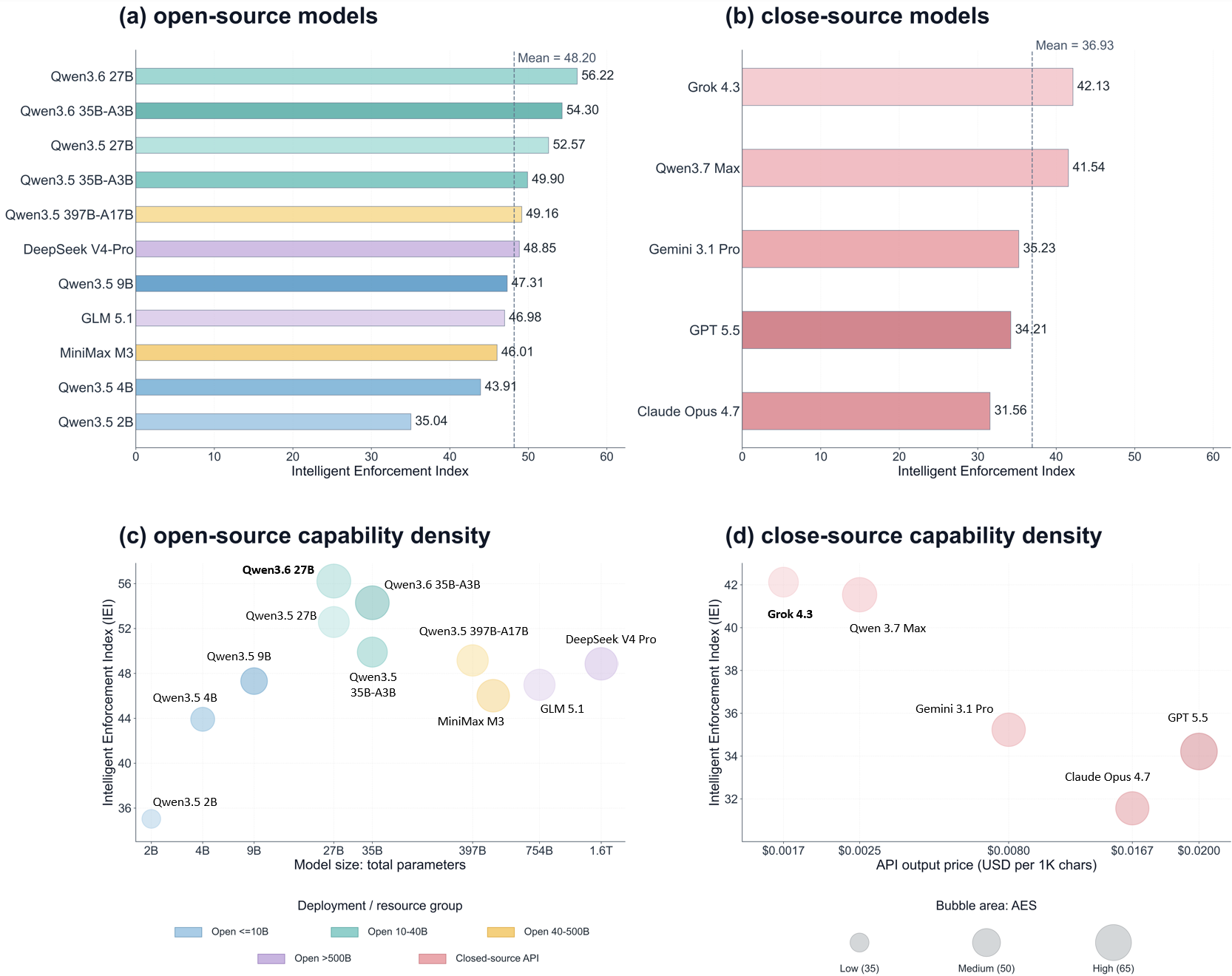}
  \captionof{figure}{Model Intelligent Enforcement Index (IEI) comparison. Panel (a) ranks open-source models by IEI; panel (b) ranks closed-source models by IEI; panel (c) plots open-source capability density against model scale and AES; and panel (d) shows the corresponding capability-density trade-off for closed-source models.}
  \label{fig:main-5-model-intelligent-enforcement-index-iei-comparison-pa}
\end{minipage}\par\vspace{0.65\baselineskip}

Overall, the combined evidence from AES and IEI reveals a capability threshold in environmental enforcement. The largest improvement occurs when models move from the 2B–9B range into the 27B–35B range, where absolute capability rises substantially and deployment efficiency reaches its highest level. Beyond this range, additional model scale produces only limited gains in AES, while the associated parameter or inference-cost burden increasingly offsets those gains, resulting in flat or declining IEI. This rank reversal appears in both model categories: the model with the highest absolute capability is not necessarily the model with the highest deployment-oriented value, while several medium-scale systems occupy the capability–resource frontier.This pattern is particularly relevant to environmental enforcement, which requires sustained, high-frequency operation under limited computational and budgetary resources.  Therefore, the central result is that environmental enforcement deployment should distinguish capability ceilings from operational efficiency: models provide greater practical value only when their additional enforcement capability is sufficient to justify the corresponding increase in resource burden.

\subsection{Lifecycle Analysis of Environmental Enforcement}

Environmental enforcement is a lifecycle process in which different stages involve different information states and decision requirements. Pre-enforcement tasks mainly deal with incomplete clues, risk signals, and preliminary screening; in-enforcement tasks require the construction and verification of violation facts from heterogeneous evidence; post-enforcement tasks are usually based on more established facts and focus on rule application, penalty reasoning, and enforcement decision generation. Therefore, lifecycle analysis helps reveal whether LLM capability remains stable across different enforcement functions, or whether it changes with the evidentiary and procedural demands of each stage.

\begin{table}[t]
\centering
\caption{Top 5 models performance across the environmental law enforcement lifecycle.}
\label{tab:task-radar-top5-aes}

\begingroup
\setlength{\heavyrulewidth}{1pt}
\setlength{\lightrulewidth}{0.5pt}

\resizebox{\textwidth}{!}{%
\begin{tabular}{
    >{\centering\arraybackslash}p{2.5cm}
    l
    rrrrr
}
\toprule
Stage & Task & GPT 5.5 & Qwen3.7 Max & Qwen3.6 27B
& Qwen3.6 35B-A3B & DeepSeek V4-Pro \\
\midrule

\multirow{4}{2.5cm}{\centering Pre-Enforcement}
& Complaint Verification
& 63.9 & 56.1 & 53.3 & 52.8 & 83.9 \\

& Inspection Task Planning
& 86.9 & 49.4 & 47.2 & 46.6 & 27.8 \\

& Public Sentiment Analysis
& 55.8 & 51.2 & 52.2 & 54.4 & 45.8 \\

& Anomaly Detection
& 70.4 & 65.4 & 67.8 & 68.6 & 63.1 \\

\midrule

\multirow{5}{2.5cm}{\centering In-Enforcement}
& Single-turn Query Analysis
& 63.9 & 61.7 & 62.8 & 65.0 & 61.9 \\

& Full Inquiry Analysis
& 56.1 & 57.5 & 53.6 & 59.7 & 61.4 \\

& Multi-Evidence Extraction
& 97.3 & 84.1 & 69.2 & 66.5 & 72.0 \\

& Contradiction Monitoring
& 26.4 & 29.1 & 27.5 & 24.2 & 22.5 \\

& Multi-Evidence Integration
& 23.6 & 30.8 & 40.7 & 49.5 & 41.8 \\

\midrule

\multirow{5}{2.5cm}{\centering Post-Enforcement}
& Penalty Type Identification
& 88.9 & 88.9 & 89.4 & 88.3 & 82.8 \\

& Penalty Decision
& 55.9 & 62.7 & 67.8 & 50.8 & 57.1 \\

& Legal Change Disturbance
& 76.9 & 75.0 & 69.9 & 77.6 & 71.2 \\

& Justification Assessment
& 75.6 & 73.7 & 75.0 & 70.4 & 74.8 \\

& Penalty Result Reasoning
& 84.6 & 85.3 & 83.1 & 80.1 & 82.1 \\

\bottomrule
\end{tabular}%
}
\endgroup
\end{table}

Table 1 reveals clear stage-specific differences in model performance. In the pre-enforcement stage, performance varies substantially across tasks and models. DeepSeek V4-Pro achieves the highest score in Complaint Verification at 83.9, markedly outperforming the other models, while GPT-5.5 leads Anomaly Detection with 70.4. Inspection Task Planning exhibits an even more pronounced capability gap: GPT-5.5 reaches 86.9, whereas the remaining models score between 27.8 and 49.4, indicating that transforming preliminary clues into structured inspection priorities, evidence-collection procedures, and inquiry plans remains difficult for most models. The in-enforcement stage presents the strongest overall bottleneck. Models perform relatively consistently on Single-turn Query Analysis and Full Inquiry Analysis, with scores mostly ranging from the mid-50s to mid-60s. GPT-5.5 also achieves 97.3 in Multi-Evidence Extraction, substantially exceeding the other models. However, performance declines sharply in tasks requiring cross-evidence reasoning. Contradiction Monitoring remains weak across all five models, with scores ranging only from 22.5 to 29.1, while Multi-Evidence Integration reaches a maximum of 49.5. By contrast, the post-enforcement stage shows the most stable performance. Penalty Type Identification scores remain between 82.8 and 89.4, Penalty Result Reasoning exceeds 80 for all five models, and both Legal Change Disturbance and Justification Assessment generally remain around 70 or above. Penalty Decision shows greater variation, ranging from 50.8 to 67.8, suggesting that discretionary judgment remains more difficult than standardized penalty classification. 

The stage difference is therefore not caused simply by the chronological position of a task in the enforcement process. Rather, it reflects the underlying structure of information and reasoning. Post-enforcement tasks are easier for LLMs because the relevant facts are usually explicit, the applicable rules are well bounded, and the output forms are relatively standardized. The weakness of LLMs is most concentrated in the in-enforcement stage, but it is not limited to that stage; whenever a task requires multi-source evidence association, factual consistency verification, or procedural organization, model reliability declines. This suggests that current LLMs are stronger in knowledge retrieval, rule application, and normative expression, while their main limitation lies in evidence-based fact determination and procedure-constrained reasoning under complex enforcement conditions. Detailed task-level performance for each model is provided in Appendix D.

\subsection{Overall Analysis Across Pollution media}

\begin{table}[t]
\centering
\caption{Top 5 models performance across pollution-medium subdomains.}
\label{tab:pollutant-top5-aes}

\begingroup
\setlength{\heavyrulewidth}{1pt}     
\setlength{\lightrulewidth}{0.5pt}   
\renewcommand{\arraystretch}{1.05}

\resizebox{\textwidth}{!}{%
\begin{tabular}{
    >{\centering\arraybackslash}m{2.8cm}
    l
    rrrrr
}
\toprule
Pollutant medium
& Category
& GPT 5.5
& Qwen3.7 Max
& Qwen3.6 27B
& Qwen3.6 35B-A3B
& DeepSeek V4-Pro \\

\midrule

\multirow[c]{3}{2.8cm}{\centering Water Pollution}
& Industrial Wastewater
& 62.7 & 59.9 & 58.3 & 57.1 & 56.1 \\

& Municipal Sewage
& 59.4 & 57.8 & 57.8 & 56.1 & 57.3 \\

& Agricultural Wastewater
& 65.6 & 65.6 & 65.3 & 60.5 & 63.1 \\

\midrule

\multirow[c]{4}{2.8cm}{\centering Air Pollution}
& Combustion Emissions
& 59.4 & 60.5 & 60.1 & 58.5 & 54.4 \\

& Industrial Air Pollution
& 62.9 & 58.0 & 58.8 & 57.8 & 55.2 \\

& Mobile Source Air Pollution
& 63.2 & 59.7 & 58.1 & 57.3 & 57.3 \\

& Agricultural Air Pollution
& 64.1 & 57.4 & 54.6 & 54.1 & 56.8 \\

\midrule

\multirow[c]{5}{2.8cm}{\centering Solid Waste Pollution}
& Industrial Solid Waste
& 63.0 & 59.9 & 57.8 & 56.8 & 58.1 \\

& Municipal Solid Waste
& 61.1 & 60.1 & 57.4 & 55.1 & 57.1 \\

& Construction Waste
& 67.7 & 60.6 & 61.2 & 62.2 & 59.9 \\

& Agricultural Solid Waste
& 70.2 & 62.0 & 57.9 & 63.1 & 61.7 \\

& Hazardous Waste
& 63.5 & 57.4 & 55.8 & 56.6 & 54.6 \\

\bottomrule
\end{tabular}%
}

\endgroup
\end{table}

To examine whether LLM capability generalizes across pollution media, we compare model performance across water pollution, air pollution, and solid waste subdomains. These domains differ in evidentiary form and regulatory logic, thus this analysis tests whether model performance is primarily shaped by pollution type itself or by the reasoning structure required by each enforcement scenario.

Table 2 shows that model performance varies more substantially across pollution-medium subdomains than across the three broad pollution media. In water pollution, agricultural wastewater is the strongest subdomain for all five models, with GPT-5.5 and Qwen3.7 Max both reaching 65.6 and Qwen3.6-27B closely following at 65.3. Municipal sewage is comparatively more challenging, with the best score reaching only 59.4 and the remaining models concentrated between 56.1 and 57.8. In air pollution, performance differs more clearly across subdomains. Qwen3.7 Max achieves the highest score of 60.5 in combustion emissions, whereas GPT-5.5 leads industrial, mobile-source, and agricultural air pollution with scores of 62.9, 63.2, and 64.1, respectively. The cross-model gap is particularly pronounced in agricultural air pollution, where scores range from 54.1 to 64.1, indicating less stable performance under heterogeneous operating conditions and less standardized evidentiary structures. Solid waste pollution exhibits the greatest internal variation. Municipal solid waste remains relatively difficult, with a leading score of 61.1, while construction waste and agricultural solid waste yield substantially higher results, led by GPT-5.5 at 67.7 and 70.2. Agricultural solid waste also shows the widest cross-model spread, ranging from 57.9 to 70.2. In hazardous waste, GPT-5.5 achieves the highest score of 63.5, while the other models range from 54.6 to 57.4.

These results indicate that the pollution medium itself is not the primary source of model bias. Rather, performance differences arise from how evidence is organized within each enforcement scenario. Models perform more reliably when pollutant facts can be mapped onto stable indicators, typical objects, or clear regulatory pathways. They become less stable when violation facts must be inferred from contextual phenomena, facility operation, uncertain clues, or multi-stage transfer chains. Therefore, the key distinction is not simply water versus air versus solid waste, but whether the task requires straightforward rule matching or complex evidence-based fact construction. This finding supports the broader conclusion that the capability boundaries of current LLMs are shaped more by task structure and evidentiary complexity than by pollution medium alone. Complete model performance across the pollution-media dimension is provided in Appendix D.

\subsection{Cross-Analysis of Enforcement Tasks and Pollution-medium subdomains}

To further identify where model capability succeeds or fails, we analyze the interaction between enforcement tasks and pollution-medium subdomains. This cross-analysis is necessary because the same task may require different evidentiary reasoning across pollutant scenarios, while the same pollution-medium subdomain may become easy or difficult depending on the task type. We therefore compute group-level performance score for open-source LLMs, closed-source LLMs, and open-source SLMs under each task--pollutant combination, in order to identify stable high-performance regions and systematic bottlenecks.

\par\noindent\begin{minipage}{\linewidth}
  \centering
  \includegraphics[width=\linewidth,height=0.72\textheight,keepaspectratio]{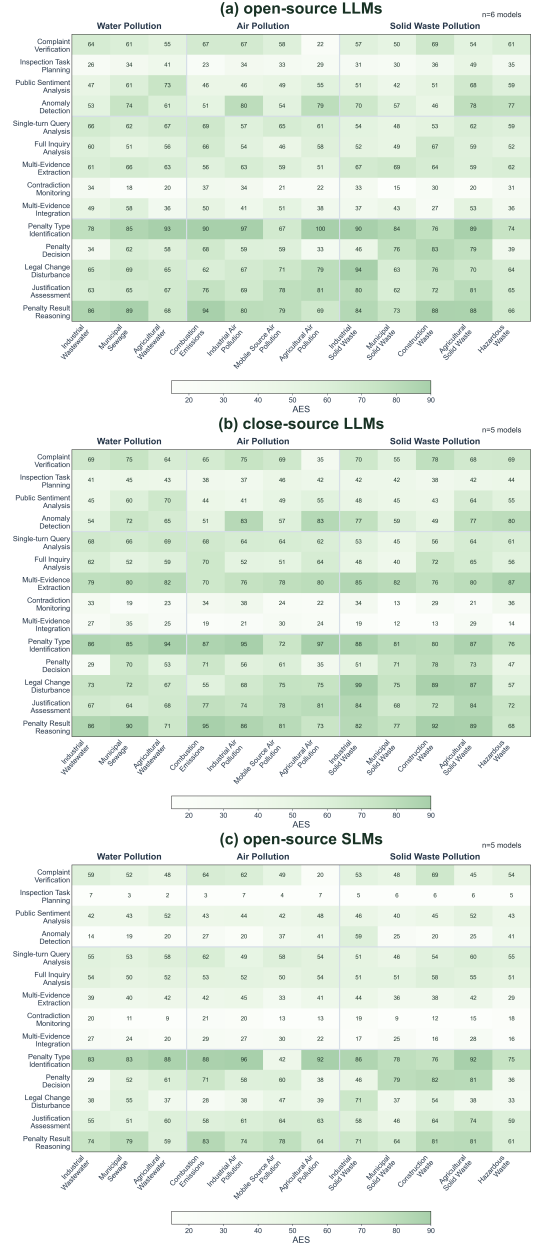}
  \captionof{figure}{Cross-analysis of enforcement tasks and pollution-medium subdomains. Panel (a) reports average AES scores for open-source LLMs, panel (b) reports average AES scores for closed-source LLMs, and panel (c) reports average AES scores for open-source SLMs.}
  \label{fig:main-14-cross-analysis-of-enforcement-tasks-and-pollution-me}
\end{minipage}\par\vspace{0.65\baselineskip}

Figure 5 shows that high-score regions are not tied to a single pollution medium, but repeatedly appear when the task structure is rule-bounded. In Penalty Type Identification, open-source large, closed-source large, and open-source SLMs score 90, 87, and 88 in combustion emissions; 97, 95, and 96 in industrial air pollution; and 93, 94, and 88 in agricultural wastewater. Penalty Result Reasoning shows the same cross-domain stability, with the three groups scoring 89, 90, and 79 in municipal sewage; and 88, 92, and 81 in construction waste. By contrast, low-score regions recur when the task requires procedural or evidentiary reasoning. In Inspection Task Planning, open-source SLMs score below 10 in most pollution-medium subdomains, while open-source and closed-source LLMs mostly remain around 30-40. Contradiction Monitoring remains weak in municipal sewage, agricultural wastewater, and municipal solid waste, with group scores of 18, 19, and 11; 20, 23, and 20; and 15, 13, and 9, respectively. Multi-Evidence Integration is also difficult even for closed-source LLMs, which score only 27, 18, and 12 in industrial wastewater, combustion emissions, and construction waste.

These results show that the capability boundary of environmental enforcement models is produced by the interaction between task structure and evidentiary complexity. The same pollution-medium subdomain can be relatively easy in a rule-matching task but become difficult in an evidence-reasoning task; conversely, the same task can remain difficult across different pollution media when it requires inspection task planning, contradiction detection, or cross-material integration. Therefore, current LLMs can generalize relatively well when enforcement reasoning follows a stable path from violation fact identification to legal-rule matching and penalty conclusion generation. However, they become unstable when they must coordinate fact determination, evidence organization, legal application, and procedural reasoning within the same task. This indicates that the central bottleneck of intelligent environmental enforcement is not regulatory knowledge recall, but evidence-grounded and procedure-constrained reasoning.

\subsection{Effects of Thinking Mode on Environmental Enforcement Capability}

From a general reasoning perspective, LLM's thinking mode is often expected to improve complex reasoning, but environmental enforcement requires not only explicit reasoning steps but also stable control over facts, evidence, and procedural constraints. We therefore compare the same models' AES score under Thinking and No-thinking modes across model categories, task types, and pollution-medium subdomains. The goal is to examine whether explicit reasoning can serve as a general enhancement strategy for environmental enforcement, or whether its value depends on model type and task structure.

\par\noindent\begin{minipage}{\linewidth}
  \centering
  \includegraphics[width=\linewidth,height=0.72\textheight,keepaspectratio]{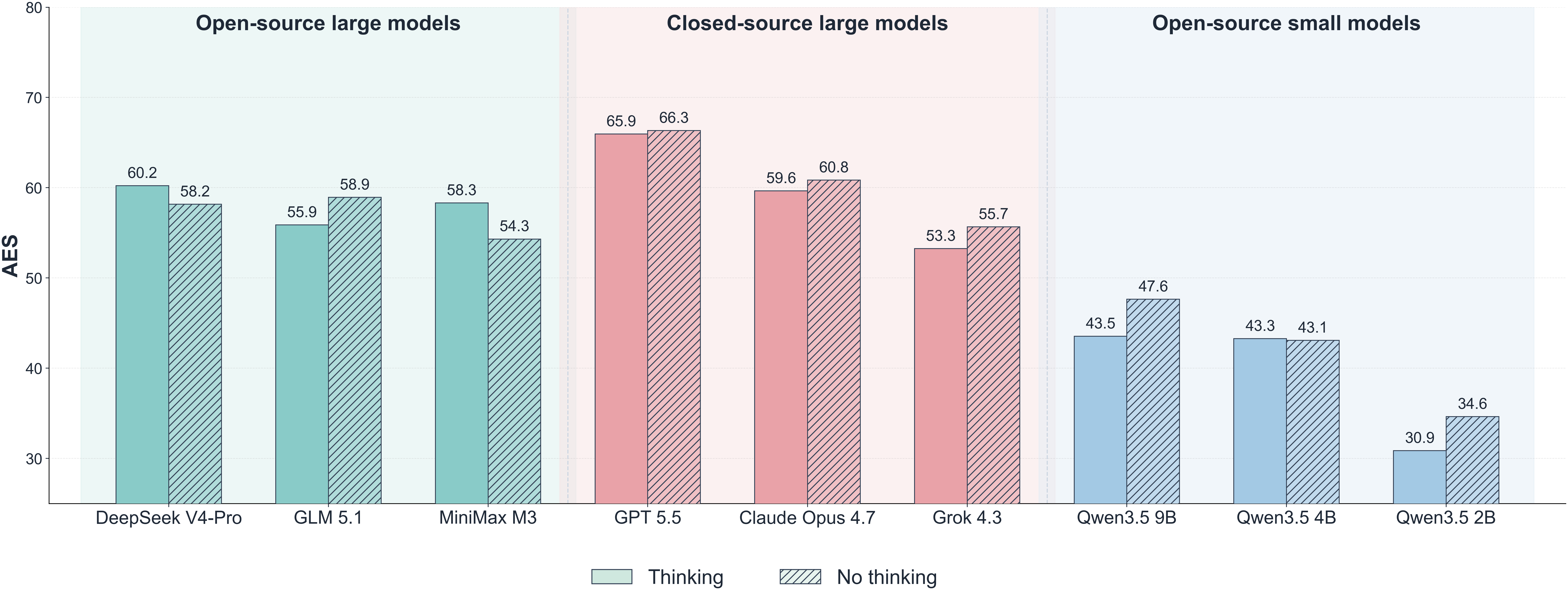}
  \captionof{figure}{Effect of Thinking mode on overall AES by models}
  \label{fig:main-15-effect-of-thinking-mode-on-overall-aes-by-models}
\end{minipage}\par\vspace{0.65\baselineskip}

Figure 6 shows that the overall effect of Thinking mode differs clearly across model categories. Among open-source LLMs, Thinking produces mixed results: DeepSeek V4-Pro improves from 58.2 to 60.2 and MiniMax M3 from 54.3 to 58.3, whereas GLM 5.1 declines from 58.9 to 55.9. This indicates that Thinking can help some open-source LLMs, but only when the model already has sufficient reasoning stability. For closed-source LLMs, Thinking does not bring overall gains: GPT-5.5 decreases from 66.3 to 65.9, Claude Opus 4.7 from 60.8 to 59.6, and Grok 4.3 from 55.7 to 53.3. For open-source SLMs, the effect is more fragile, with Qwen3.5-9B and Qwen3.5-2B declining by 4.1 and 3.7 points.

\par\noindent\begin{minipage}{\linewidth}
  \centering
  \includegraphics[width=\linewidth,height=0.72\textheight,keepaspectratio]{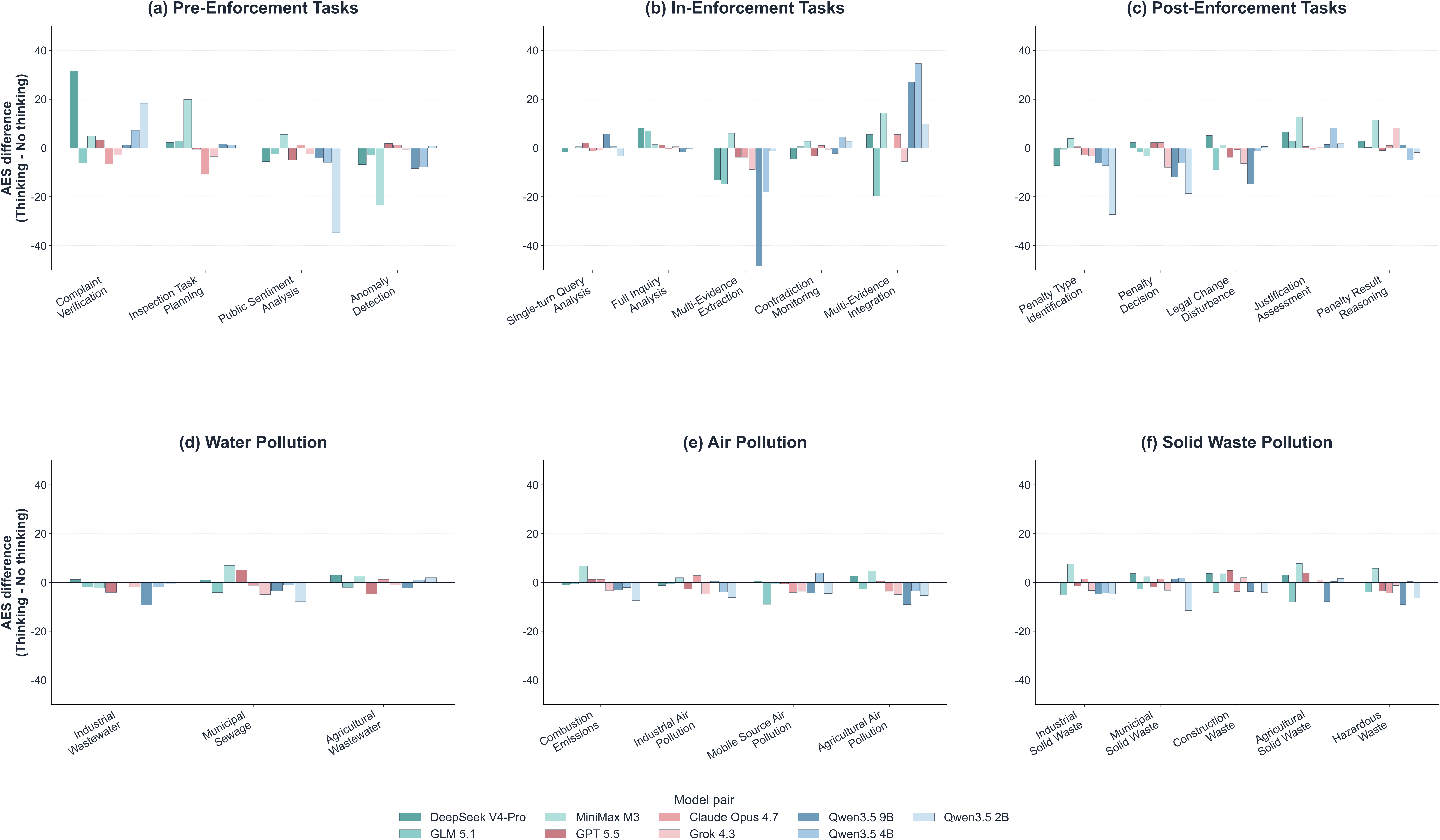}
  \captionof{figure}{Task- and domain-level AES changes under Thinking mode. Panels (a)-(c) show score differences for pre-enforcement, in-enforcement, and post-enforcement tasks; panels (d)-(f) show score differences across water, air, and solid-waste pollution subdomains.}
  \label{fig:main-16-task-and-domain-level-aes-changes-under-thinking-mod}
\end{minipage}\par\vspace{0.65\baselineskip}

Figure 7 further shows that these differences are mainly task-dependent. Thinking improves tasks that benefit from explicit organization of judgment conditions, such as Complaint Verification, Full Inquiry Analysis, Inspection Task Planning, Multi-Evidence Integration, Justification Assessment, and Penalty Result Reasoning. For example, DeepSeek V4-Pro gains 31.7 points in Complaint Verification and 8.1 points in Full Inquiry Analysis, while MiniMax M3 gains 12--20 points in several chain-reasoning tasks. However, Thinking reduces performance in tasks with more complex, heterogeneous, or weakly structured inputs. In Multi-Evidence Extraction, DeepSeek V4-Pro, GLM 5.1, GPT-5.5, and Claude Opus 4.7 decline by 13.2, 14.8, 3.9, and 3.9 points, and Qwen3.5-9B drops from 64.8 to 16.5. Similar declines appear in Public Sentiment Analysis and Anomaly Detection, such as MiniMax M3 decreasing by 23.3 points in Anomaly Detection and Qwen3.5-2B decreasing by 34.7 points in Public Sentiment Risk Warning.

These two figures together suggest that Thinking mode is not a general-purpose enhancement mechanism. Its value depends on both model category and task structure. For open-source LLMs, Thinking can provide useful intermediate organization when the model has enough baseline capability to keep the reasoning process stable. For closed-source LLMs, the marginal benefit is limited, likely because their default responses already contain strong implicit reasoning and alignment. For open-source SLMs, Thinking is more likely to amplify unstable reasoning and weaken output consistency. More importantly, the task-level pattern explains why gains and losses occur. When the input is relatively clear, structured, or can be organized around explicit verification criteria, Thinking helps models identify missing elements, arrange reasoning steps, and express procedural logic. When the input is complex, multi-source, or weakly structured, as in evidence extraction, public sentiment warning, and anomaly detection, Thinking may amplify the model's tendency toward self-explanation, making the output less faithful to the original signals and less reliable for enforcement use. Therefore, reasoning mode in environmental enforcement should be task-adaptive rather than enabled by default.

\subsection{Case Analysis}

The preceding sections identify model capability differences from aggregate performance, enforcement stages, pollution media, task–pollutant interactions, and Thinking mode. However, these score-level results do not fully show how models succeed or fail in concrete enforcement responses. This section therefore uses representative cases to explain the mechanisms behind three recurring patterns: why rule‑based enforcement decisions are easier than evidence‑chain reasoning, why medium‑sized models can approach the capability boundary in structured tasks, and why Thinking mode is only useful when the reasoning structure is controllable. The full task prompts, gold annotations, raw model outputs, and judge rationales are provided in Appendix E.

\par\noindent\begin{minipage}{\linewidth}
  \centering
  \includegraphics[width=\linewidth,height=0.72\textheight,keepaspectratio]{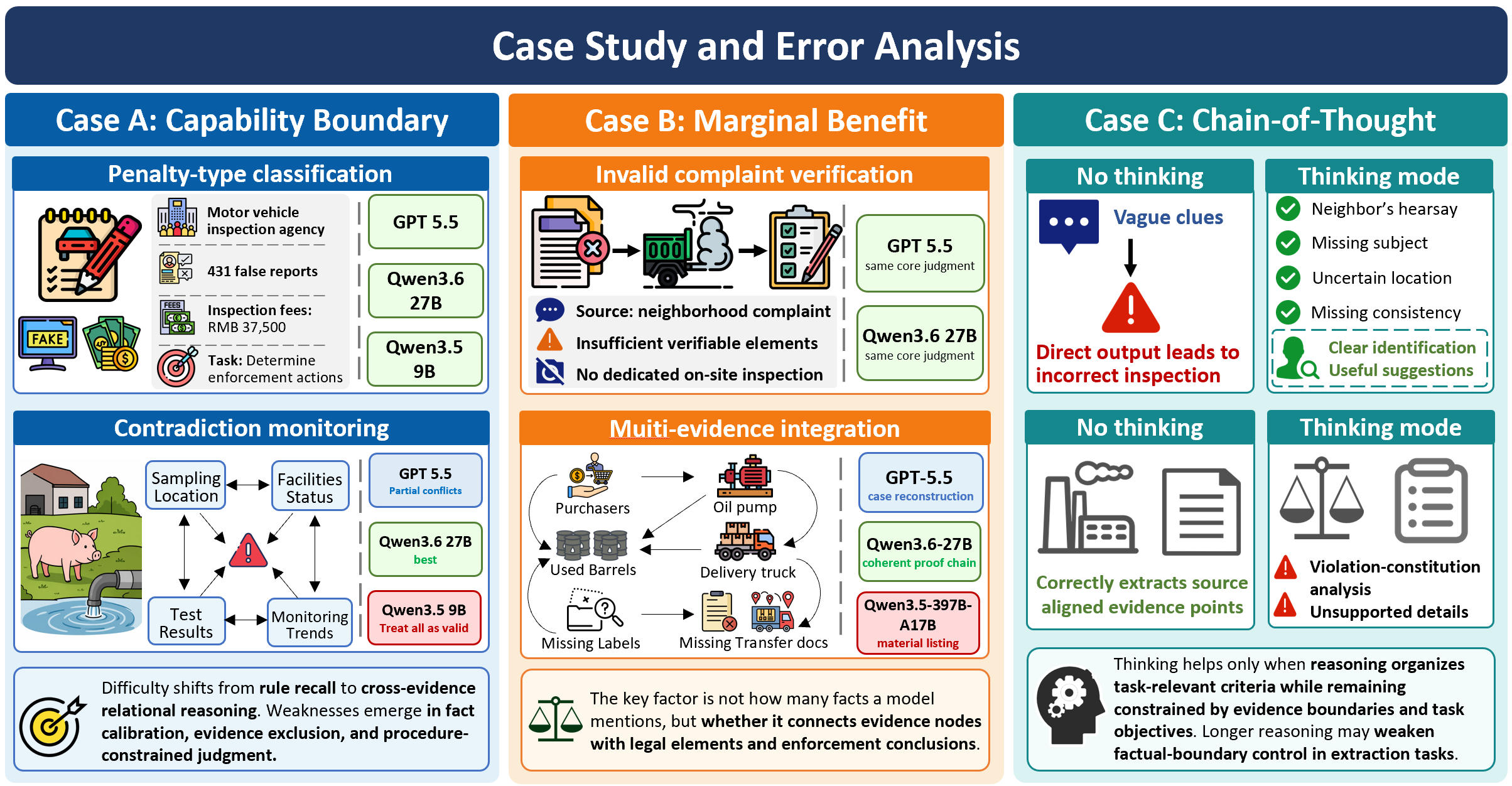}
  \captionof{figure}{Representative cases across WuYu-EnvLE-Bench tasks}
  \label{fig:main-17-penalty-type-classification-in-a-vehicle-emission-in}
\end{minipage}\par\vspace{0.65\baselineskip}

Figure 8 summarizes three representative qualitative cases that illustrate the main capability boundaries identified above. \textbf{Case A} illustrates \emph{the contrast between rule-bounded decision-making and evidence-chain reasoning}. In the penalty-type classification setting, models can stably map false vehicle inspection reports to a fine and inspection fees to confiscation of illegal gains. However, in the agricultural wastewater contradiction-monitoring case, performance drops sharply because models must jointly assess sampling location, facility operation status, rapid-test procedure, and evidence credibility. This case suggests that the difficulty of environmental enforcement does not begin with legal-rule recall. Rather, the task shifts from single-fact recognition to cross-evidence relational reasoning, where models remain weak in fact calibration, evidence exclusion, and procedure-constrained judgment.

\textbf{Case B} illustrates \emph{the marginal effect of model scaling} in structured enforcement tasks. In an invalid mobile-source air-pollution complaint, GPT-5.5 and Qwen3.6-27B reach the same core judgment: the clue lacks sufficient verifiable elements and should enter general risk screening rather than a dedicated on-site inspection. Their differences mainly lie in expression completeness and reasoning organization. In the hazardous-waste multi-evidence integration case, however, larger scale or MoE architecture does not automatically produce better evidence-chain reasoning. GPT-5.5 and Qwen3.5-397B-A17B MoE cover many factual elements, but their outputs remain closer to material listing or case reconstruction. By contrast, Qwen3.6-27B connects the purchaser, oil-pumping tools, small truck, used-oil barrels, HW08 identification, missing labels, transfer materials, and repeated sales into a more coherent proof chain. This case indicates that the decisive factor is not how many facts a model can mention, but whether it can connect evidence nodes with legal elements and enforcement conclusions.

\textbf{Case C} illustrates \emph{the task-dependent effect of Thinking mode}. In an invalid complaint verification task, Thinking helps DeepSeek V4-Pro identify hearsay sources, missing responsible subjects, uncertain location, and absent time-frequency information, preventing the model from turning weak clues directly into inspection actions. In an industrial air-pollution evidence extraction task, however, Thinking causes Qwen3.5-9B to drift from source-aligned extraction toward violation-constitution analysis and to introduce unsupported details. This case shows that Thinking mode is useful only when explicit reasoning helps organize task-relevant judgment conditions while remaining constrained by evidence boundaries and task objectives. In source-attribution and evidence-extraction tasks, longer reasoning may instead weaken factual-boundary control.

Overall, current LLMs are more reliable in rule application and standardized enforcement expression, especially when facts are explicit, legal consequences are stable, and output formats are fixed. Their limitations become more evident when tasks require cross-evidence calibration, evidence exclusion, multi-source integration, and procedure-constrained judgment. Therefore, reliable reasoning in environmental enforcement should not be understood as longer reasoning alone, but as reasoning constrained by evidence boundaries, task objectives, and enforcement procedures.

\section{Discussion}

Based on the assessment results, we discuss two central questions: where current LLMs can provide reliable support across the environmental enforcement lifecycle, and whether further model scaling continues to produce meaningful enforcement gains. The results show that model capability in this domain is shaped less by general language ability alone than by task structure, evidentiary complexity, and the degree to which reasoning can be constrained by enforcement rules and procedures.

\par\noindent\textbf{RQ1: What are the strengths and weaknesses of LLMs across the environmental law enforcement lifecycle?}\par

Current LLMs show useful but task-dependent capability in environmental enforcement. They perform more reliably when facts are explicit, rules are well bounded, and outputs are standardized, such as in penalty-type classification, penalty-result reasoning, legal-change disturbance, and anomaly identification. Performance declines when tasks shift from rule application to evidence-based fact construction, including cross-material consistency verification, evidence sufficiency judgment, and procedural reasoning. This weakness is most concentrated in the in-enforcement stage, but it is not limited to that stage; across water, air, and solid waste domains, performance differences are driven more by task structure and evidentiary complexity than by pollution type itself. Reasoning mode also needs task adaptation rather than default activation. Explicit reasoning can help identify information gaps and organize judgment chains, but in evidence extraction, strict-format output, and source-attribution tasks, excessive reasoning may cause factual-boundary drift. Therefore, reliable reasoning in environmental enforcement is reasoning constrained by evidence boundaries, task objectives, and enforcement procedures. Current LLMs are better suited for rule matching, material organization, and document assistance than for replacing enforcement officers in fact determination, evidence review, or procedural judgment.

\par\noindent\textbf{RQ2: Does environmental law enforcement exhibit diminishing returns from model scaling?}\par

Current LLMs also show a clear marginal effect of model capability in environmental enforcement. Scaling from small to medium-sized models substantially improves terminology understanding, context handling, instruction following, and basic reasoning, but further parameter expansion does not bring proportional enforcement gains once models reach a medium or large scale. At that point, the main bottleneck shifts from general knowledge coverage to evidence-relation modeling, factual-boundary control, and procedural reasoning. Therefore, the highest capability score does not necessarily indicate the highest deployment value, especially in enforcement scenarios requiring long-term operation, large-scale invocation, and controllable costs. Medium-sized models may already provide sufficient capability for rule-bounded and fact-explicit tasks, whereas larger models mainly improve expression completeness and local robustness without necessarily overcoming evidence-chain reasoning limitations. Future development should therefore move beyond parameter scaling and focus on domain-rule representation, source-grounded evidence modeling, contradiction detection, and procedure-constrained reasoning.

Overall, WuYu-EnvLE-Bench reveals that current LLMs are useful but bounded tools for environmental enforcement. Their strengths lie in rule-bounded and standardized tasks, while their main limitations emerge in evidence-based fact construction, contradiction verification, and procedure-constrained reasoning. At the same time, model scaling shows clear diminishing returns once basic domain competence is reached. Future environmental enforcement LLMs should therefore move from scale-centered improvement toward evidence-grounded, rule-aware, and procedure-constrained optimization.

\section{Conclusion}

This study introduced WuYu-EnvLE-Bench, a benchmark for evaluating large language models in environmental law enforcement. Built from real enforcement cases, regulatory standards, and practitioner review, the benchmark covers the full enforcement workflow across pre-enforcement, in-enforcement, and post-enforcement stages, as well as water pollution, air pollution, and solid waste scenarios. By combining AES for capability assessment with IEI for deployment-oriented evaluation, WuYu-EnvLE-Bench provides a systematic basis for examining not only whether models can complete enforcement tasks, but also whether their capability is reliable, explainable, and resource-efficient.

Our results show that current LLMs can provide useful support in rule-bounded and standardized enforcement tasks, especially when facts are explicit, legal rules are clear, and outputs follow stable formats. However, their reliability decreases substantially when tasks require evidence-chain construction, contradiction verification, multi-source material integration, or procedure-constrained judgment. These limitations are most visible in the in-enforcement stage, where models must transform heterogeneous evidence into reviewable violation facts. The results also reveal clear diminishing returns from model scaling: medium-sized models can approach leading models in structured tasks, while larger models do not necessarily deliver proportional gains in evidence-based reasoning or deployment value.

Overall, WuYu-EnvLE-Bench shows that LLMs are promising but bounded tools for environmental enforcement. They are better suited for rule matching, material organization, preliminary analysis, and document assistance than for autonomous fact determination or penalty discretion. Future environmental enforcement models should therefore move beyond scale-centered improvement and focus on domain-rule representation, evidence-grounded reasoning, contradiction detection, procedural constraints, and task-adaptive inference. This direction is essential for developing LLM-based systems that can support accountable, traceable, and practically deployable environmental enforcement.

\section{Acknowledgment}

The authors gratefully acknowledge the financial support from the Jing-Jin-Ji Regional Integrated Environmental Improvement - National Science and Technology Major Project (No.2025ZD1208400). We also gratefully acknowledge the experts from the Appraisal Center for Environmental Engineering, Ministry of Ecology and Environment, for their valuable guidance and professional insights.

\clearpage

\bibliographystyle{unsrtnat}
\bibliography{references}

\clearpage
\appendix
\numberwithin{figure}{section}
\numberwithin{table}{section}
\section{Ethical Considerations}

The source materials used to construct WuYu-EnvLE-Bench were derived from ecological and environmental enforcement documents, regulatory texts, and case materials that were either publicly available or processed under appropriate research-use constraints. To mitigate privacy risks, all case materials were anonymized before benchmark construction and before being submitted to non-locally deployed large language models. Identifiable information, including enterprise names, personal names, exact addresses, contact details, identification numbers, social credit codes, and precise geographic coordinates, was removed or replaced with functional descriptions. At the same time, enforcement-relevant facts, evidentiary structures, legal applicability conditions, and regulatory outcomes were retained to preserve the validity of the evaluation tasks.

WuYu-EnvLE-Bench is designed to evaluate the capability boundaries of large language models in environmental law enforcement scenarios. Its purpose is to support research on more reliable, interpretable, and deployable enforcement-assistance models. We do not suggest that large language models can independently conduct environmental enforcement, determine violations, or make administrative penalty decisions without human verification. In real-world applications, model outputs should only be used as auxiliary decision-support information and must be reviewed by qualified enforcement professionals.

\section{Task Design}

\subsection*{B.1 Pre-enforcement}

\subsubsection*{B.1.1 Complaint Verification}

\subsubsection*{B.1.1.1 Motivation}

Citizen complaints are an important source of violation clues in environmental enforcement and represent a typical risk-identification task in the pre-enforcement stage. Unlike standardized materials such as administrative penalty decisions or monitoring reports, complaints are often submitted through phone calls, online messages, or petition materials. They are usually colloquial, emotional, and incomplete, often lacking clear information about location, responsible subject, time, or direct evidence.

In real enforcement practice, officers must quickly determine whether a complaint has sufficient verification value. Sending vague, unlocatable, or non-specific complaints directly to on-site inspection would waste limited enforcement resources and delay responses to high-value clues. Therefore, this task evaluates whether large language models can assess the validity of citizen complaints. The model is required not only to judge whether a complaint is verifiable, but also to extract key elements such as time, location, subject, behavior, consequence, and source reliability, and to identify remaining information gaps. This task mainly tests information extraction, vague clue assessment, verifiability judgment, and inspection preparation capabilities in the pre-enforcement stage.

\subsubsection*{B.1.1.2 Data Construction}

This task is constructed from real environmental enforcement cases. For each case, we extract a standardized Case Brief containing the responsible subject, pollution behavior, location, temporal information, environmental impact, and enforcement outcome, and then rewrite it into citizen complaint texts that resemble phone reports, online messages, or petition statements. To generate samples with different levels of verifiability, we control two factors: information completeness and linguistic vagueness. Key elements such as location, subject, discharge time, and pollution phenomenon may be retained or removed to form valid, invalid, or low-verifiability complaints, while formal case facts are rewritten into more colloquial and uncertain expressions. Each sample is paired with gold annotations, including the validity label, key information elements, a verifiability score of 1--5, information gaps, and, for valid complaints, inspection priorities, evidence collection suggestions, and inquiry points.

\subsubsection*{B.1.1.3 Prompt}

The evaluated model receives the task prompt shown in Figure B.1.

\par\noindent\begin{minipage}{\linewidth}
  \centering
  \includegraphics[width=\linewidth,height=0.82\textheight,keepaspectratio]{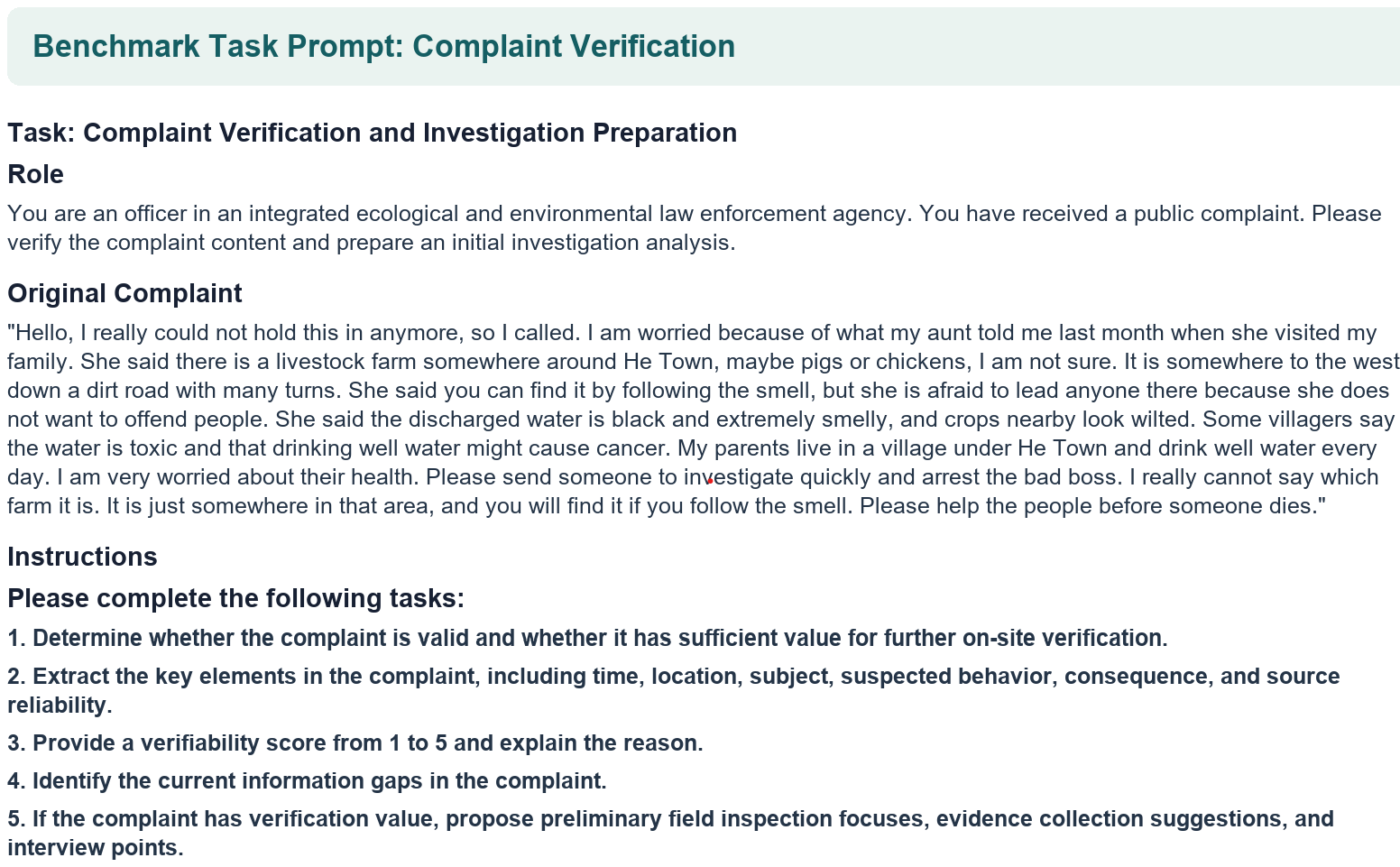}
  \captionof{figure}{Complaint verification benchmark task prompt.}
  \label{fig:appendix-b-1-complaint-verification-benchmark-task-prompt}
\end{minipage}\par\vspace{0.65\baselineskip}

The model is required to produce a complete verification-preparation analysis, rather than only outputting a binary judgment of ``valid'' or ``invalid.'' The evaluation therefore considers not only whether the validity judgment is correct, but also whether the model covers key complaint information, identifies missing elements, designs a reasonable verification plan, and provides practically executable enforcement suggestions. Inspection priorities, evidence collection suggestions, and inquiry points are used as important criteria for assessing whether the model can translate complaint clues into actionable enforcement plans.

\subsubsection*{B.1.1.4 Evaluation}

This task uses an LLM-as-a-Judge evaluation framework. The judge model receives the task input, candidate response, Case Brief, Gold Answer, and Evaluation Rubric, and evaluates whether the model correctly identifies enforcement-relevant information from vague and incomplete complaint texts.

The evaluation focuses on four aspects: conclusion correctness, key information coverage, professional reasoning quality, and enforcement executability. Conclusion correctness measures whether the model correctly determines the validity and verification value of the complaint. Key information coverage measures whether the model accurately extracts time, location, subject, behavior, consequence, and source reliability. Professional reasoning quality evaluates whether the verifiability score and explanation follow the logic of environmental enforcement screening. Enforcement executability assesses whether the model identifies critical information gaps and provides actionable inspection suggestions when the complaint is valid.

\par\noindent\begin{minipage}{\linewidth}
  \centering
  \includegraphics[width=\linewidth,height=0.82\textheight,keepaspectratio]{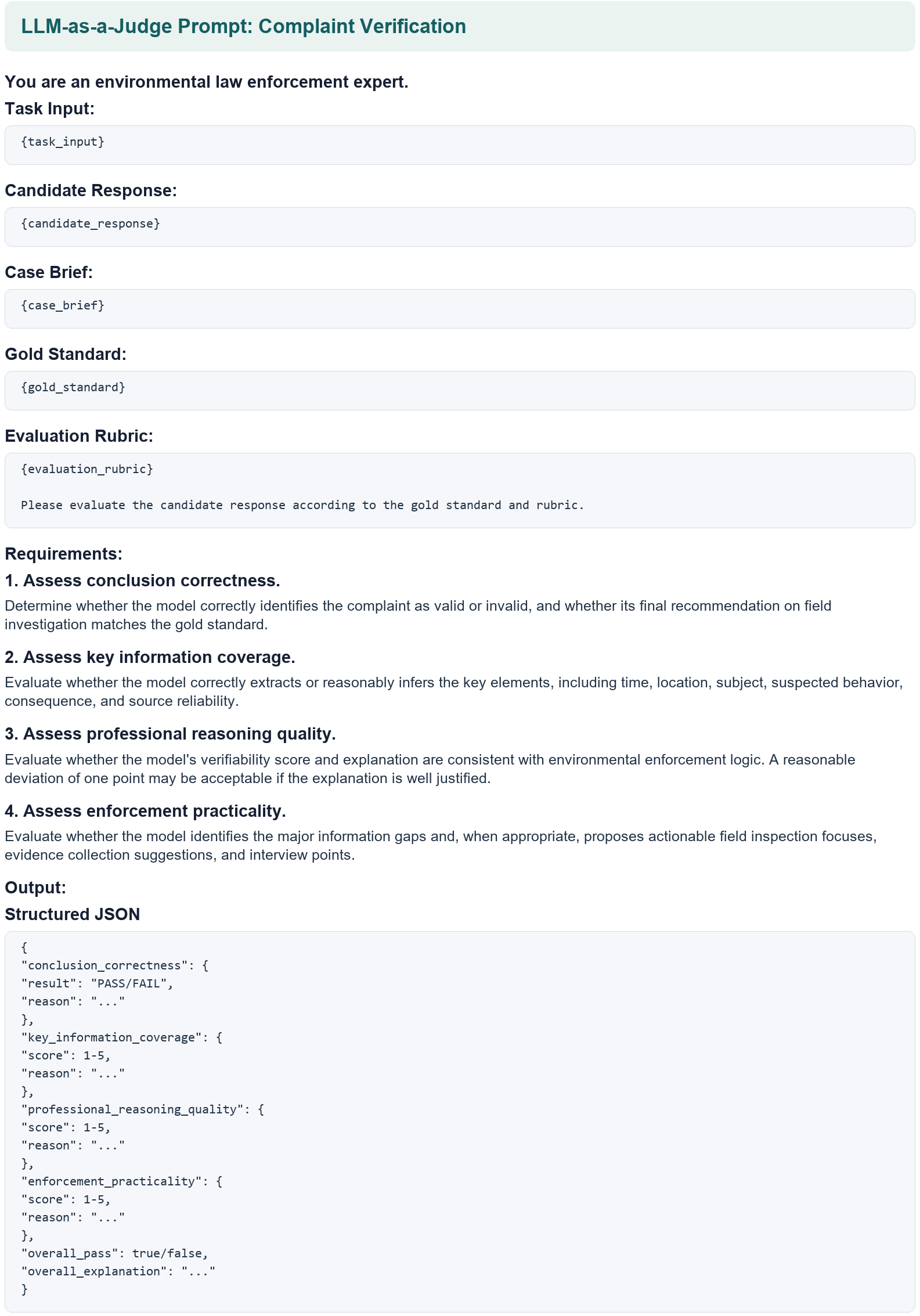}
  \captionof{figure}{Complaint verification LLM-as-a-Judge prompt.}
  \label{fig:appendix-b-2-complaint-verification-llm-as-a-judge-prompt}
\end{minipage}\par\vspace{0.65\baselineskip}

\subsubsection*{B.1.2 Inspection Task Planning}

\subsubsection*{B.1.2.1 Motivation}

Inspection task planning is a key step after a citizen complaint is judged to have verification value. In real environmental enforcement, complaint clues often provide only pollution phenomena and approximate locations. Before conducting an on-site inspection, enforcement officers need to determine inspection priorities, prepare evidence collection methods, and identify relevant interview targets. An incomplete inspection plan may lead to missing key evidence, overlooking important facilities, selecting inappropriate sampling points, or failing to question relevant persons, thereby affecting violation determination and subsequent penalty procedures.

This task evaluates whether models can transform valid complaint clues into concrete, systematic, and executable on-site inspection plans. It mainly tests the model's ability to decompose enforcement tasks, identify evidentiary needs, and organize field investigation procedures.

\subsubsection*{B.1.2.2 Data Construction}

This task is constructed from samples that were judged as valid in B.1.1 Complaint Verification. These complaints contain sufficient information, such as a relatively clear location, responsible subject, pollution behavior, or environmental consequence, to support further on-site inspection.For each valid complaint, we combine the complaint text with its corresponding Case Brief to construct a gold-standard inspection plan. The gold answer is organized as a structured field inspection checklist, including three components: Inspection Focus, Evidence Suggestions, and Interview Points. Inspection Focus specifies the facilities, discharge pathways, key locations, and environmentally sensitive targets that should be checked. Evidence Suggestions specify the samples, photographs, videos, documents, ledgers, electronic data, and other materials that should be collected. Interview Points specify the persons to be questioned and the key facts to be verified.

The generated gold answers are further revised according to the investigation logic of the original cases to ensure professional validity and practical executability. Each sample finally contains a structured gold checklist and an evaluation rubric for judging the candidate model output.

\subsubsection*{B.1.2.3 Prompt}

The evaluated model receives a valid complaint and is required to generate an executable on-site inspection plan.

\par\noindent\begin{minipage}{\linewidth}
  \centering
  \includegraphics[width=\linewidth,height=0.82\textheight,keepaspectratio]{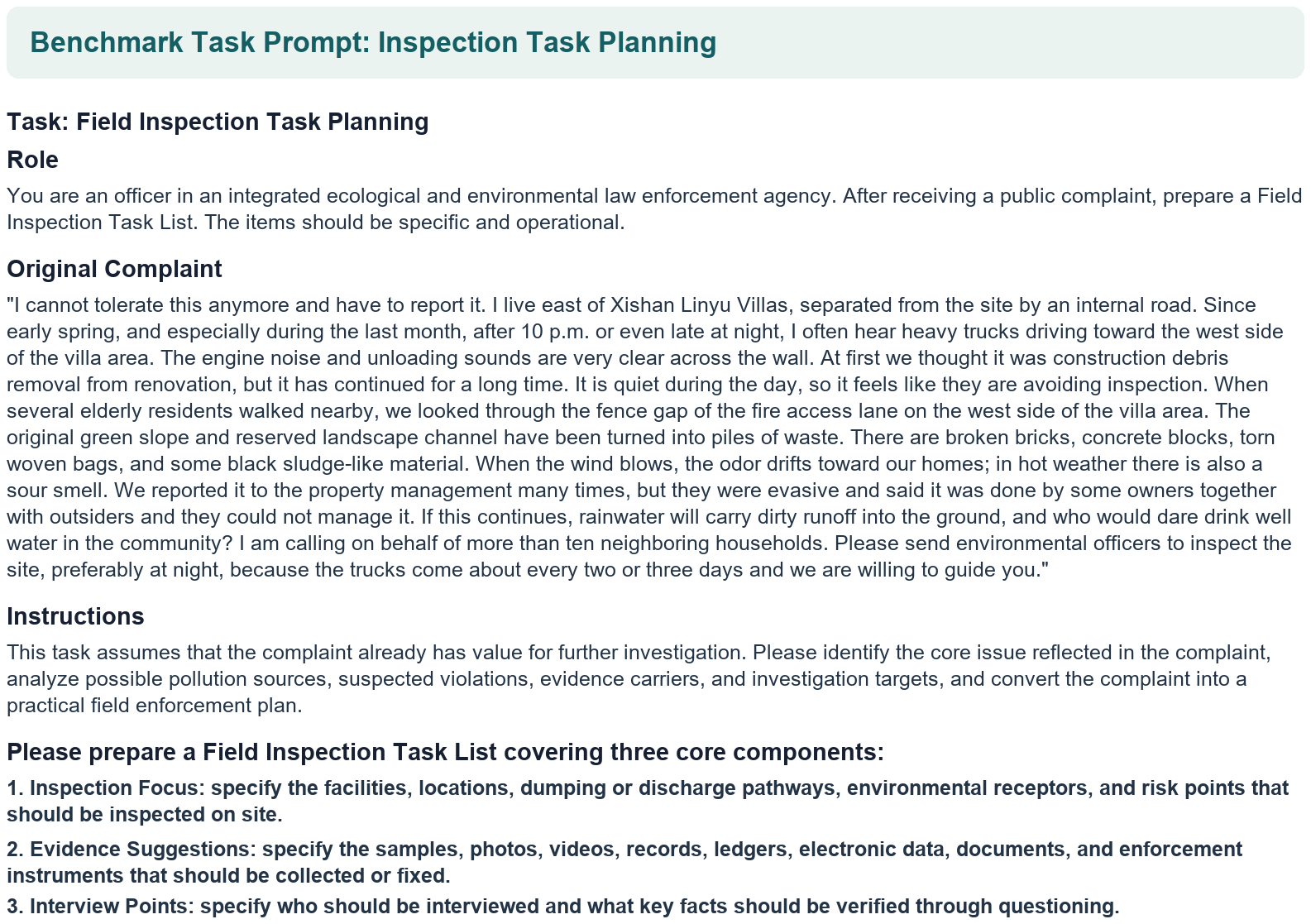}
  \captionof{figure}{Field inspection task planning benchmark prompt.}
  \label{fig:appendix-b-3-field-inspection-task-planning-benchmark-prompt}
\end{minipage}\par\vspace{0.65\baselineskip}

Unlike Complaint Verification, this task assumes that the complaint already has further investigation value. The model must identify the core issue in the complaint, infer possible pollution sources, violation behaviors, evidentiary carriers, and investigation targets, and convert them into concrete enforcement actions. The required output includes Inspection Focus, Evidence Suggestions, and Interview Points.

\subsubsection*{B.1.2.4 Evaluation}

This task is evaluated using an LLM-as-a-Judge framework. The judge model receives the complaint text, candidate response, Case Brief, gold inspection checklist, and evaluation rubric, and assesses whether the generated plan is appropriate, specific, and executable. The evaluation focuses on three dimensions: Inspection Focus, Evidence Suggestions, and Interview Points. Inspection Focus measures whether the model identifies key facilities, locations, discharge pathways, pollution sources, and sensitive targets. Evidence Suggestions evaluates whether the model proposes sufficient evidence collection measures, including sampling, photos or videos, records, ledgers, permits, electronic data, and procedural requirements such as sampling location and chain of custody. Interview Points assesses whether the model identifies relevant inquiry targets and designs questions around discharge behavior, facility operation, ledger authenticity, environmental impact, and responsibility. The complete Judge Prompt is shown below.

\par\noindent\begin{minipage}{\linewidth}
  \centering
  \includegraphics[width=\linewidth,height=0.82\textheight,keepaspectratio]{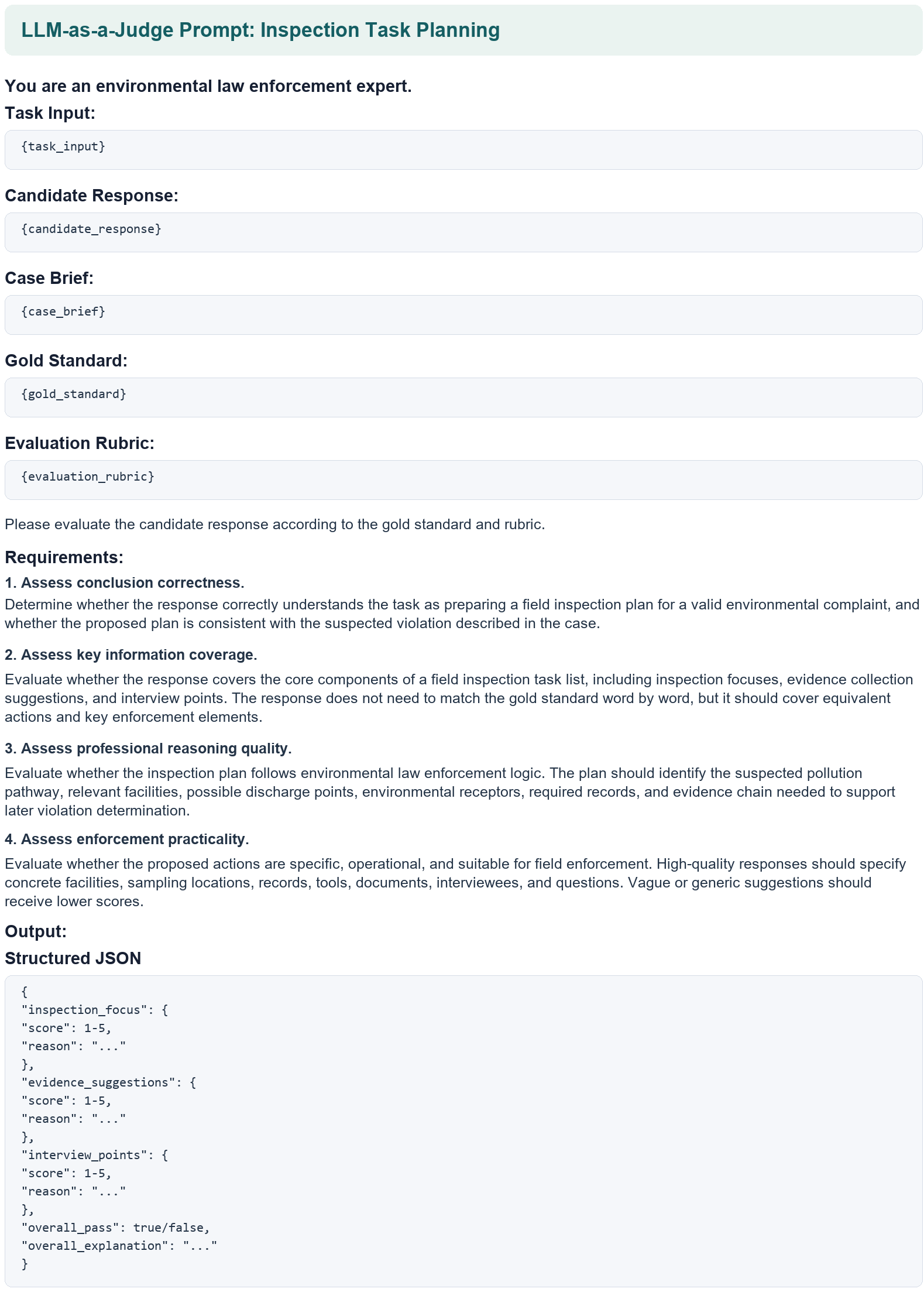}
  \captionof{figure}{Field inspection task planning LLM-as-a-Judge prompt.}
  \label{fig:appendix-b-4-field-inspection-task-planning-llm-as-a-judge-p}
\end{minipage}\par\vspace{0.65\baselineskip}

\subsubsection*{B.1.3 Public Sentiment Analysis}

\subsubsection*{B.1.3.1 Motivation}

Public sentiment analysis provides an important early-warning signal for non-site environmental enforcement. Online discussions may reveal pollution clues before formal complaints or inspections, but such clues are often scattered across posts, comments, follow-up replies, and reposts. These materials are typically noisy, fragmented, emotional, and mixed with irrelevant or misleading information. This task evaluates whether models can identify environmental violation risks from noisy public sentiment texts, distinguish strong and weak evidence, integrate dispersed clues, and recommend appropriate warning responses.

\subsubsection*{B.1.3.2 Data Construction}

This task uses aggregated public sentiment materials, including main posts, comments, and follow-up replies, to simulate noisy online environmental discussions. To reflect the high-noise and low-signal nature of real public sentiment data, the samples include verifiable risk clues, weak supporting signals, misleading information, emotional complaints, and irrelevant content. Based on real case briefs and online discussion patterns, we generate public sentiment texts with temporal progression, cross-comment corroboration, partial disagreement, and noise interference. The final aggregated text is annotated with gold answers, including risk existence, risk level, core conclusion, strong evidence, weak evidence, and recommended response action. Since key information may be distributed across multiple comments rather than stated in the main post, the task requires models to integrate time, location, subject, behavior, and environmental impact across the full discussion thread.

\subsubsection*{B.1.3.3 Prompt}

The evaluated model receives aggregated public sentiment materials and is required to produce an environmental risk alert analysis.

\par\noindent\begin{minipage}{\linewidth}
  \centering
  \includegraphics[width=\linewidth,height=0.82\textheight,keepaspectratio]{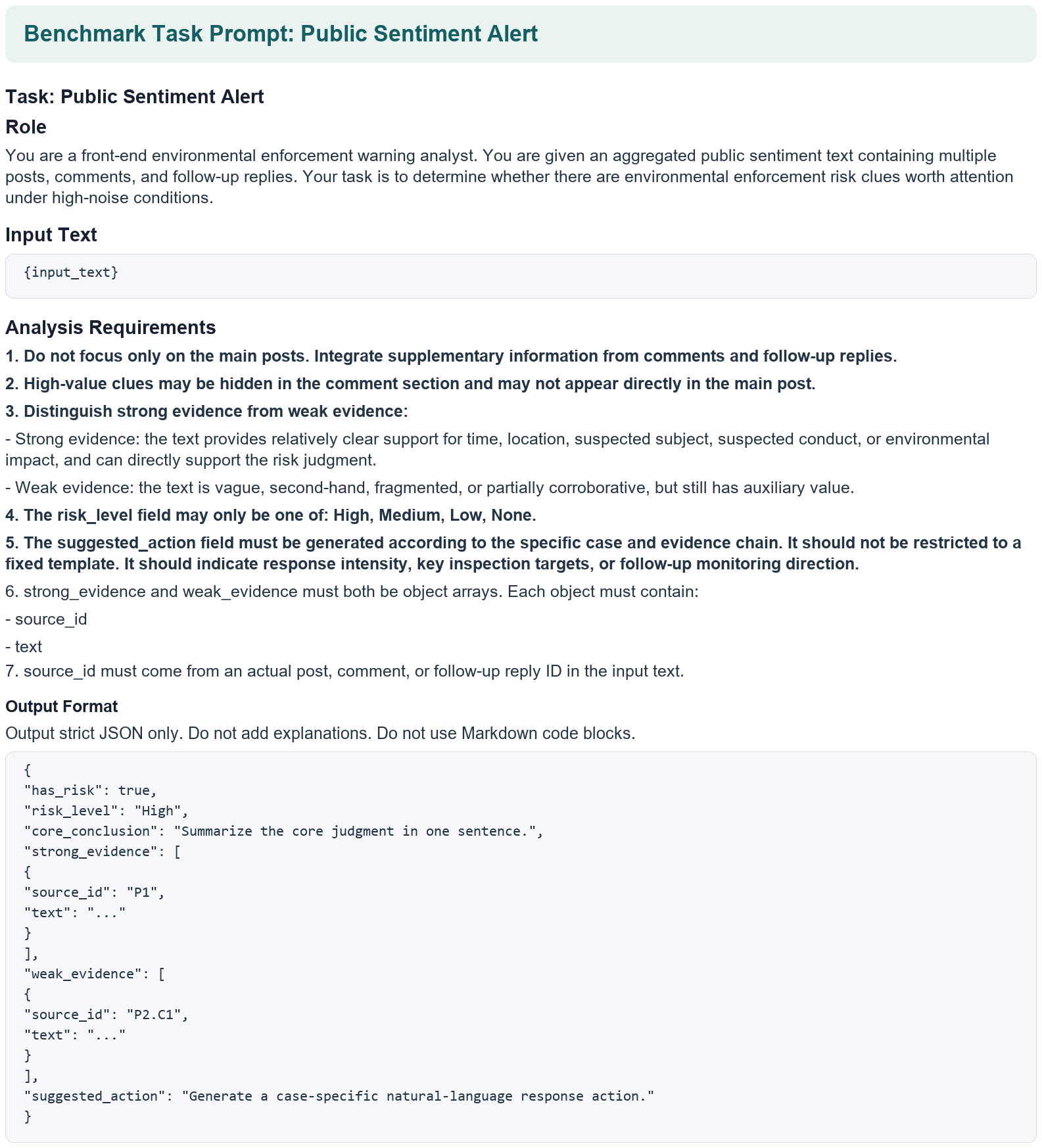}
  \captionof{figure}{Public sentiment alert benchmark task prompt.}
  \label{fig:appendix-b-5-public-sentiment-alert-benchmark-task-prompt}
\end{minipage}\par\vspace{0.65\baselineskip}

The model output includes risk existence, risk level, core conclusion, strong evidence, weak evidence, and recommended response action. The task evaluates whether the model can identify meaningful environmental risk signals from noisy online text and convert them into appropriate non-site enforcement warnings.

\clearpage
\par\noindent\begin{minipage}{\linewidth}
  \centering
  \includegraphics[width=\linewidth,height=0.80\textheight,keepaspectratio]{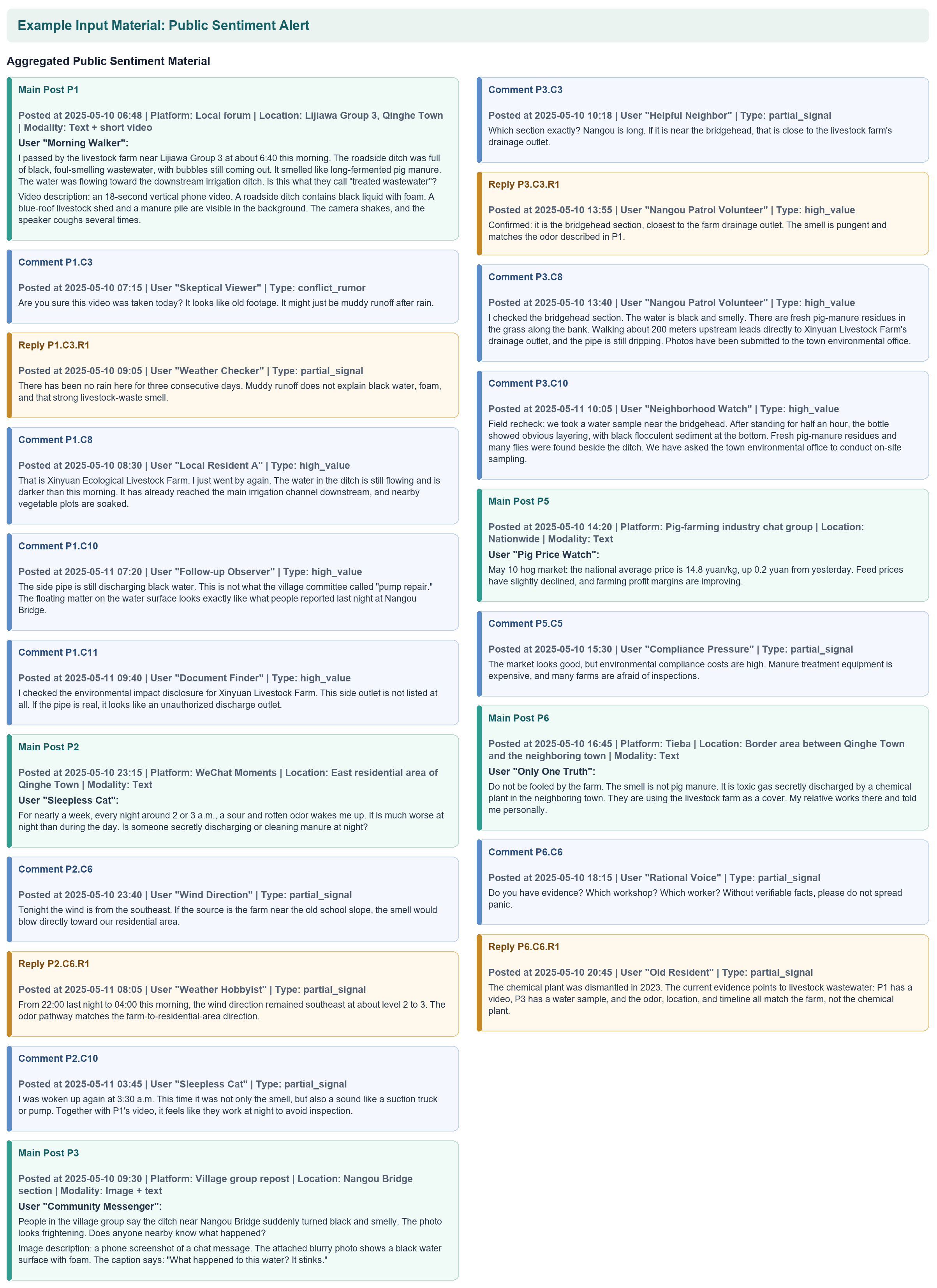}
  \captionof{figure}{Public sentiment alert example input material.}
  \label{fig:appendix-b-6-public-sentiment-alert-example-input-material}
\end{minipage}\par\vspace{0.65\baselineskip}

\subsubsection*{B.1.3.4 Evaluation}

This task is evaluated using an LLM-as-a-Judge framework. The judge model receives the aggregated public sentiment text, candidate response, Gold Answer, and Evaluation Rubric, and evaluates the response from five aspects: risk identification, risk level, core conclusion, evidence recall, and response action.Risk identification assesses whether the model correctly determines whether an environmental risk exists. Risk level assesses whether the model correctly classifies the severity as high, medium, low, or none. Core conclusion assesses whether the model accurately summarizes the main risk fact, including the subject, behavior, time or frequency, pollution medium, and environmental impact, without adding unsupported facts. Evidence recall assesses whether the model identifies the key strong evidence in the gold answer, such as clear time, location, image or video descriptions, and corroborating comments. Response action assesses whether the proposed enforcement response matches the risk level, such as immediate on-site inspection for high-risk cases, focused monitoring for medium-risk cases, continued observation for low-risk cases, or no action for irrelevant noise. The complete Judge Prompt is shown below.

\par\noindent\begin{minipage}{\linewidth}
  \centering
  \includegraphics[width=\linewidth,height=0.82\textheight,keepaspectratio]{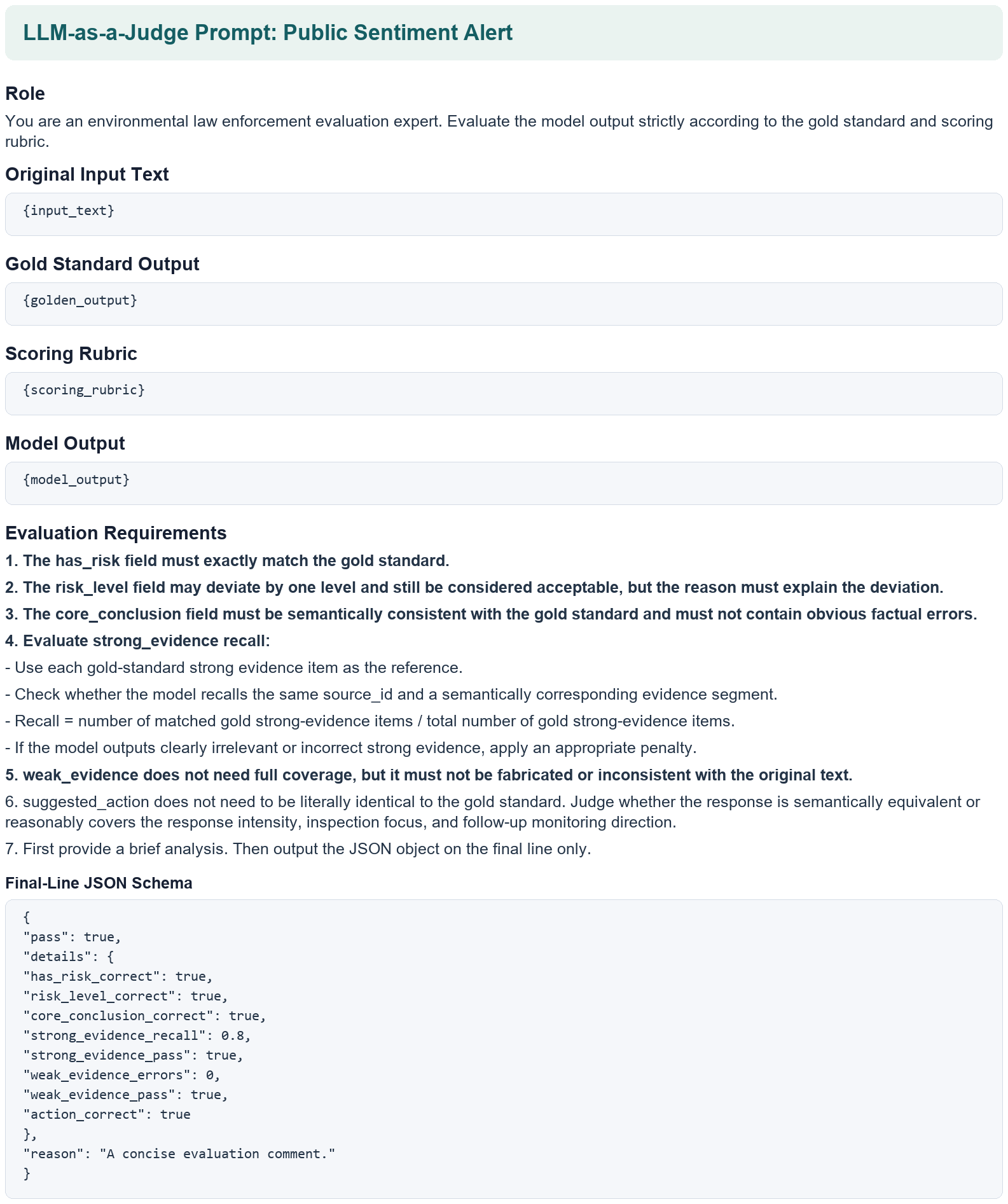}
  \captionof{figure}{Public sentiment alert LLM-as-a-Judge prompt.}
  \label{fig:appendix-b-7-public-sentiment-alert-llm-as-a-judge-prompt}
\end{minipage}\par\vspace{0.65\baselineskip}

\subsubsection*{B.1.4 Anomaly Detection}

\subsubsection*{B.1.4.1 Motivation}

Non-site environmental enforcement relies on multi-source regulatory data, such as automatic monitoring records, electricity and water consumption data, video records, hazardous waste manifests, and transportation trajectories. Although these data enable continuous supervision, manual review is inefficient and may fail to detect complex abnormal patterns. Enterprises may evade supervision through diluted sampling, parameter manipulation, constant-value simulation, falsified transport trajectories, or manifest mismatches. This task evaluates whether models can identify abnormal patterns, logical conflicts, and potential data falsification from time-series and multi-source enforcement data, and propose relevant follow-up verification actions.

\subsubsection*{B.1.4.2 Data Construction}

This task is constructed through case-driven multi-source anomaly simulation. For each case, we extract enterprise background, production process, discharge or waste-generation characteristics, and regulatory clues from the Case Brief, and then generate scenario-specific monitoring materials for water pollution, air pollution, or solid/hazardous waste management. Water samples cover indicators such as COD, ammonia nitrogen, total phosphorus, total nitrogen, pH, and flow rate; air samples cover SO$_2$, NOx, particulate matter, oxygen content, flue gas velocity, temperature, pressure, and VOCs/NMHC; solid and hazardous waste samples cover waste generation, inventory, transfer manifests, transport trajectories, weighbridge records, and disposal receipts. The simulated anomalies include sudden pollutant drops, long-term constant values, flow--concentration inconsistencies, parameter manipulation, abnormal treatment-facility operation, fixed-station/mobile-monitoring conflicts, and mismatches among manifests, trajectories, weights, and inventory. Each sample contains enterprise background information and timestamped multi-source data with both normal and abnormal segments, requiring models to identify anomalies through trend comparison and cross-source verification. For large time-series tables, we use HTML format to improve structural stability and require XML-style output tags to support consistent parsing.

\subsubsection*{B.1.4.3 Prompt}

The evaluated model receives continuous monitoring data and multi-source regulatory materials.

\par\noindent\begin{minipage}{\linewidth}
  \centering
  \includegraphics[width=\linewidth,height=0.82\textheight,keepaspectratio]{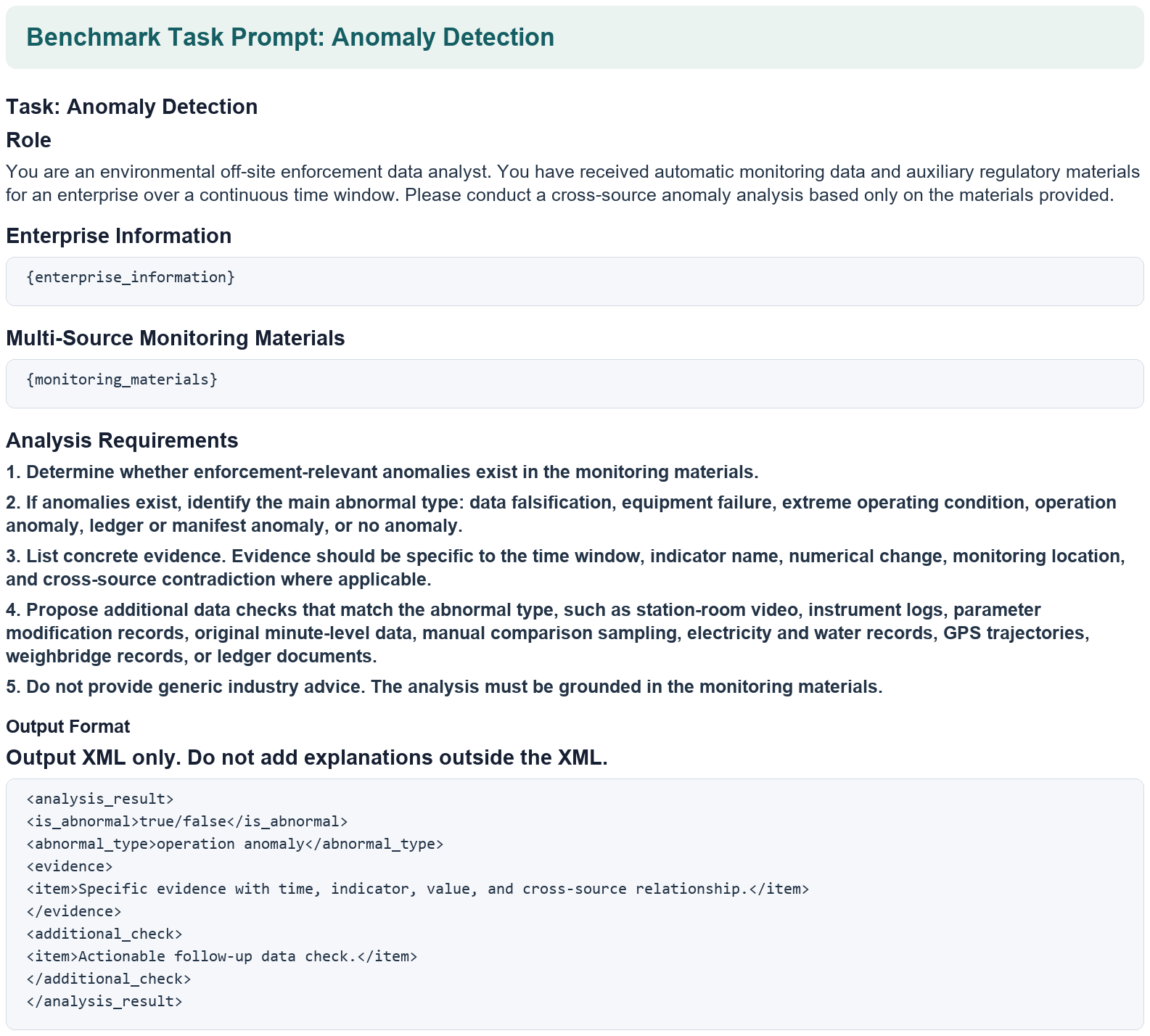}
  \captionof{figure}{Anomaly detection benchmark task prompt.}
  \label{fig:appendix-b-8-anomaly-detection-benchmark-task-prompt}
\end{minipage}\par\vspace{0.65\baselineskip}

\par\noindent\begin{minipage}{\linewidth}
  \centering
  \includegraphics[width=\linewidth,height=0.82\textheight,keepaspectratio]{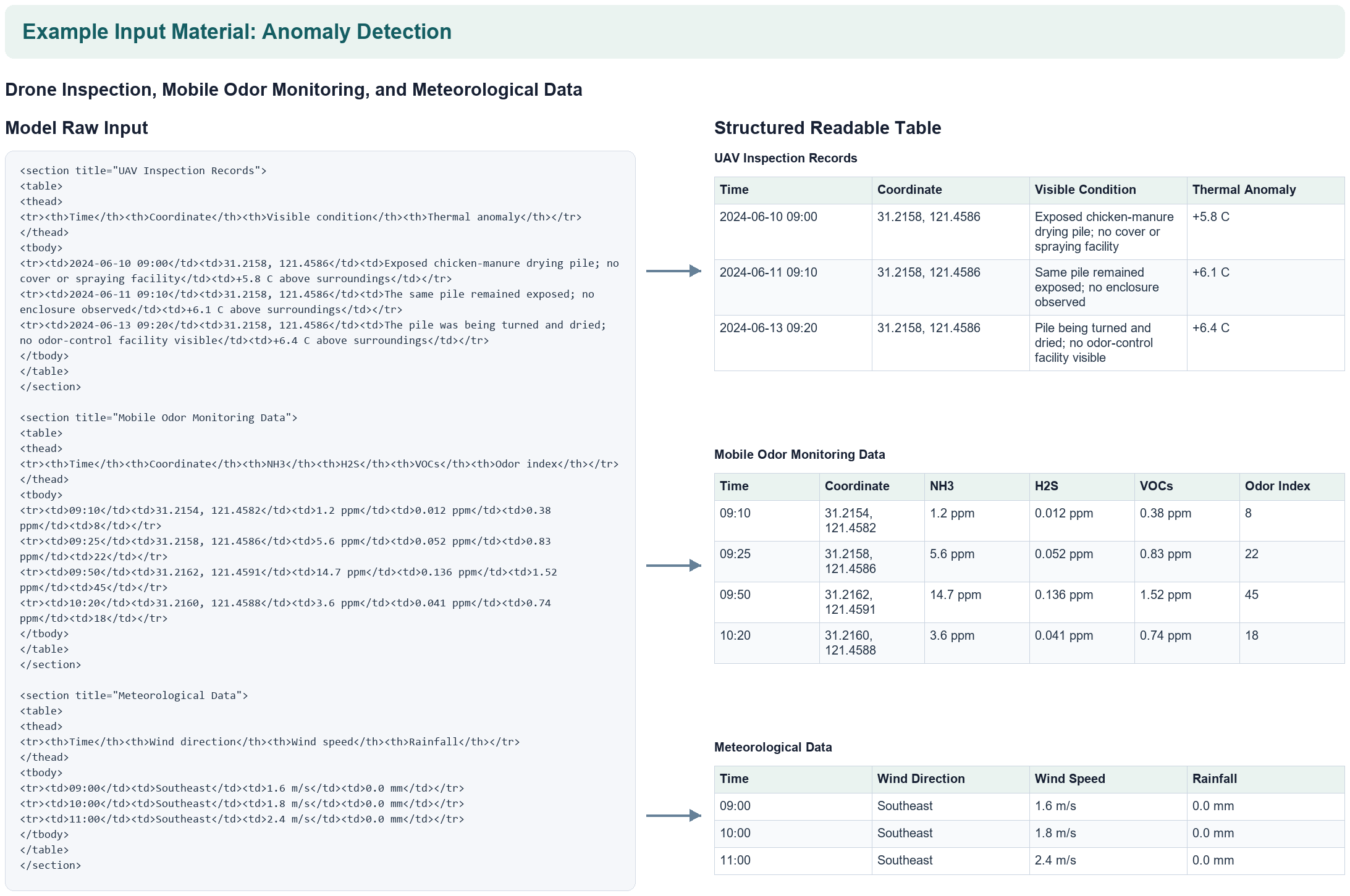}
  \captionof{figure}{Anomaly detection example input material.}
  \label{fig:appendix-b-9-anomaly-detection-example-input-material}
\end{minipage}\par\vspace{0.65\baselineskip}

The model must make judgments based on the provided data rather than general industry knowledge. The output includes whether an anomaly exists, the anomaly type, supporting evidence, and follow-up verification suggestions. Evidence should specify the relevant time window, indicator names, numerical changes, and cross-source conflicts. Verification suggestions should match the anomaly type, such as checking station-room videos, parameter modification logs, raw minute-level data, weighbridge records, GPS trajectories, or pollution-control facility operation logs.

\subsubsection*{B.1.4.4 Evaluation}

This task is evaluated using an LLM-as-a-Judge framework. The judge model receives the Case Brief, task prompt, candidate response, Gold Answer, and Evaluation Rubric, and evaluates the response along four dimensions. Anomaly judgment assesses whether the model correctly determines the presence of an anomaly. Anomaly type accuracy assesses whether the model identifies the core anomaly type, such as data falsification, parameter manipulation, equipment failure, abnormal operation, or manifest--trajectory mismatch. Evidence localization assesses whether the model identifies the key time window, indicator combination, and cross-source conflict; this is the central criterion for determining whether the model truly understands the time-series data. Follow-up verification relevance assesses whether the proposed verification actions match the evidence chain, such as retrieving parameter logs, station-room videos, manual comparison monitoring, electricity or water records, manifests, trajectories, or weighbridge data.

\par\noindent\begin{minipage}{\linewidth}
  \centering
  \includegraphics[width=\linewidth,height=0.82\textheight,keepaspectratio]{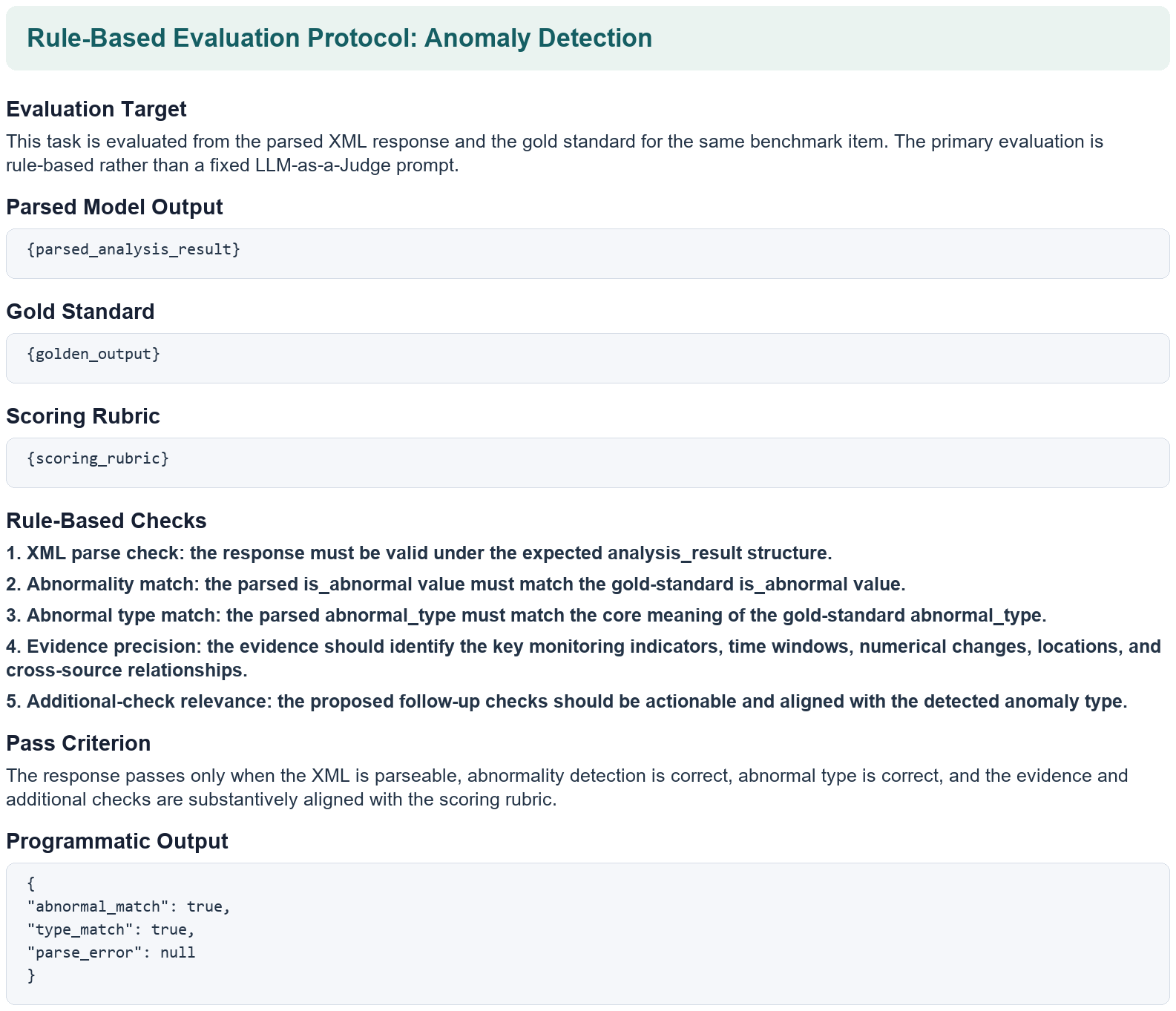}
  \captionof{figure}{Anomaly detection LLM-as-a-Judge prompt.}
  \label{fig:appendix-b-10-anomaly-detection-llm-as-a-judge-prompt}
\end{minipage}\par\vspace{0.65\baselineskip}

\subsection*{B.2 In-enforcement}

\subsubsection*{B.2.1 Single-turn Query Analysis}

\subsubsection*{B.2.1.1 Motivation}

Enforcement questioning is an important step in fixing violation facts, identifying responsible parties, and completing evidence chains during on-site environmental enforcement. In practice, respondents may conceal facts, evade key questions, provide misleading explanations, or weaken liability through vague expressions such as ``I do not remember,'' ``it was temporary,'' or ``only a small amount.'' This task evaluates whether models can identify abnormal responses from a single question--answer pair and classify the abnormality type. It corresponds to a real-time questioning alert scenario, where the model assists enforcement officers in detecting concealment, deception, evasion, or misleading tendencies and adjusting follow-up questions accordingly.

\subsubsection*{B.2.1.2 Data Construction}

This task is constructed from real environmental enforcement case briefs. For each case, we extract core facts, including the responsible subject, violation behavior, key facilities, time and location, evidence materials, and final determination, and then generate a single-round question--answer pair between an enforcement officer and a respondent. The questions focus on key violation elements, such as facility operation, wastewater discharge, solid waste storage, hazardous waste transfer, or environmental acceptance procedures. The samples include both normal and abnormal responses. Abnormal responses may deny evidence-supported facts, avoid key time or quantity information, describe violations as temporary, or guide the officer toward an incomplete explanation, while normal responses are consistent with case facts and directly answer the question. Each sample is annotated with whether an abnormality exists, the abnormality type, supporting rationale, and, when necessary, suggested follow-up questions. The abnormality types include vagueness, evasiveness, concealment, deception, and misleading response.

\subsubsection*{B.2.1.3 Prompt}

The evaluated model receives the case brief, prior dialogue context, the officer's question, and the respondent's answer, and must determine whether the single response is abnormal.

\par\noindent\begin{minipage}{\linewidth}
  \centering
  \includegraphics[width=\linewidth,height=0.82\textheight,keepaspectratio]{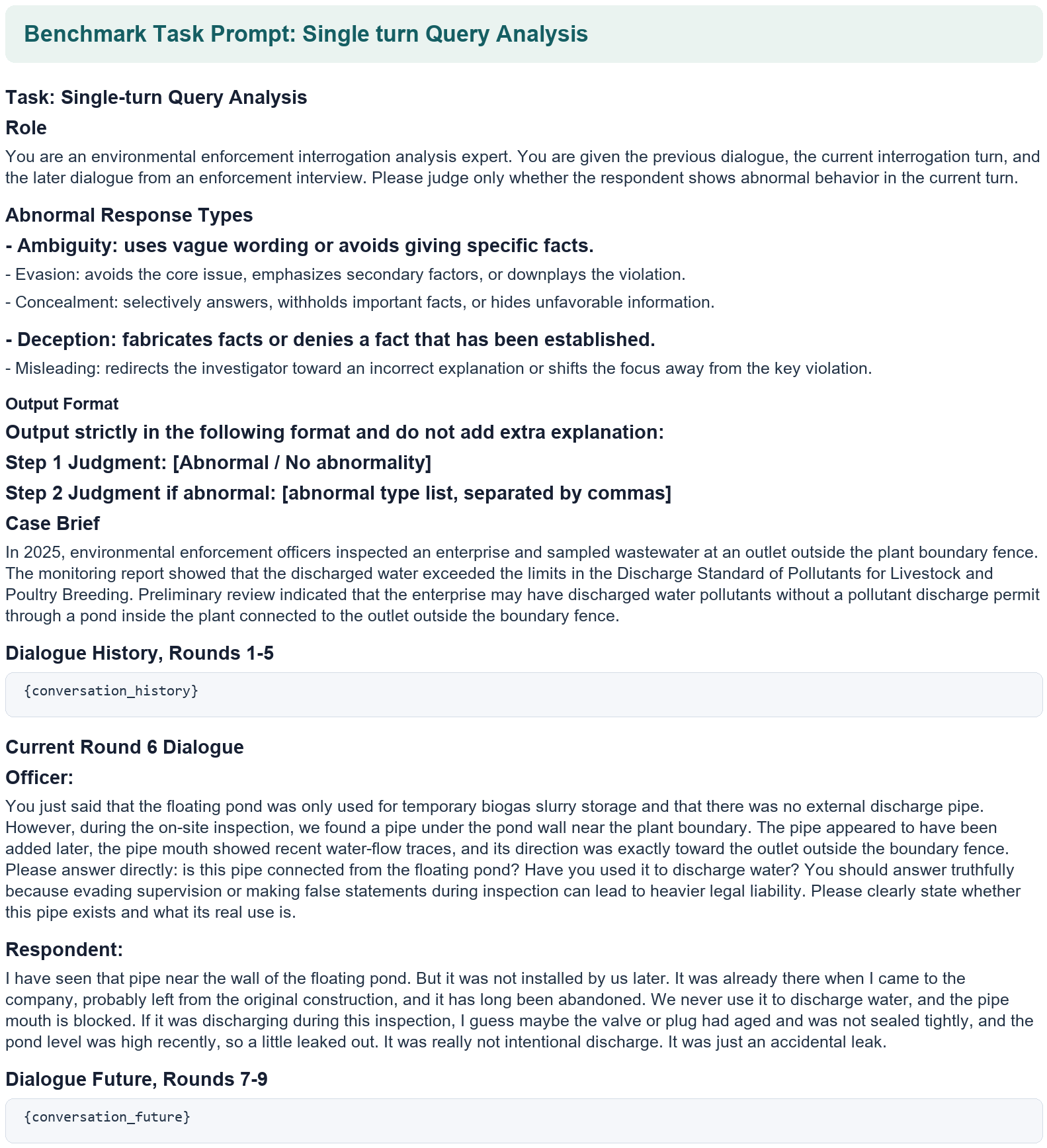}
  \captionof{figure}{Single-turn query analysis benchmark task prompt.}
  \label{fig:appendix-b-11-single-turn-query-analysis-benchmark-task-prom}
\end{minipage}\par\vspace{0.65\baselineskip}

The task focuses on whether the model can identify deviations from case facts, avoidance of the question, conflicts with known evidence, or language strategies that may mislead the investigation. The full prior dialogue is omitted here for brevity because complete questioning materials are provided in B.2.2.

\subsubsection*{B.2.1.4 Evaluation}

This task is evaluated using an LLM-as-a-Judge framework. The judge model receives the case brief, single-round question--answer pair, gold answer, candidate response, and evaluation rubric, and assesses the response from four dimensions: conclusion correctness, key information coverage, professional reasoning quality, and enforcement executability. Conclusion correctness measures whether the model correctly identifies the presence or absence of an abnormal response. Key information coverage measures whether the model captures central abnormal points such as time evasion, fact denial, quantity minimization, responsibility shifting, or conflict with evidence. Professional reasoning quality evaluates whether the abnormality type follows enforcement questioning logic, avoiding both over-classification of normal uncertainty and under-classification of evidence-conflicting statements. Enforcement executability assesses whether the model's explanation can support follow-up questioning or statement fixation.

\par\noindent\begin{minipage}{\linewidth}
  \centering
  \includegraphics[width=\linewidth,height=0.82\textheight,keepaspectratio]{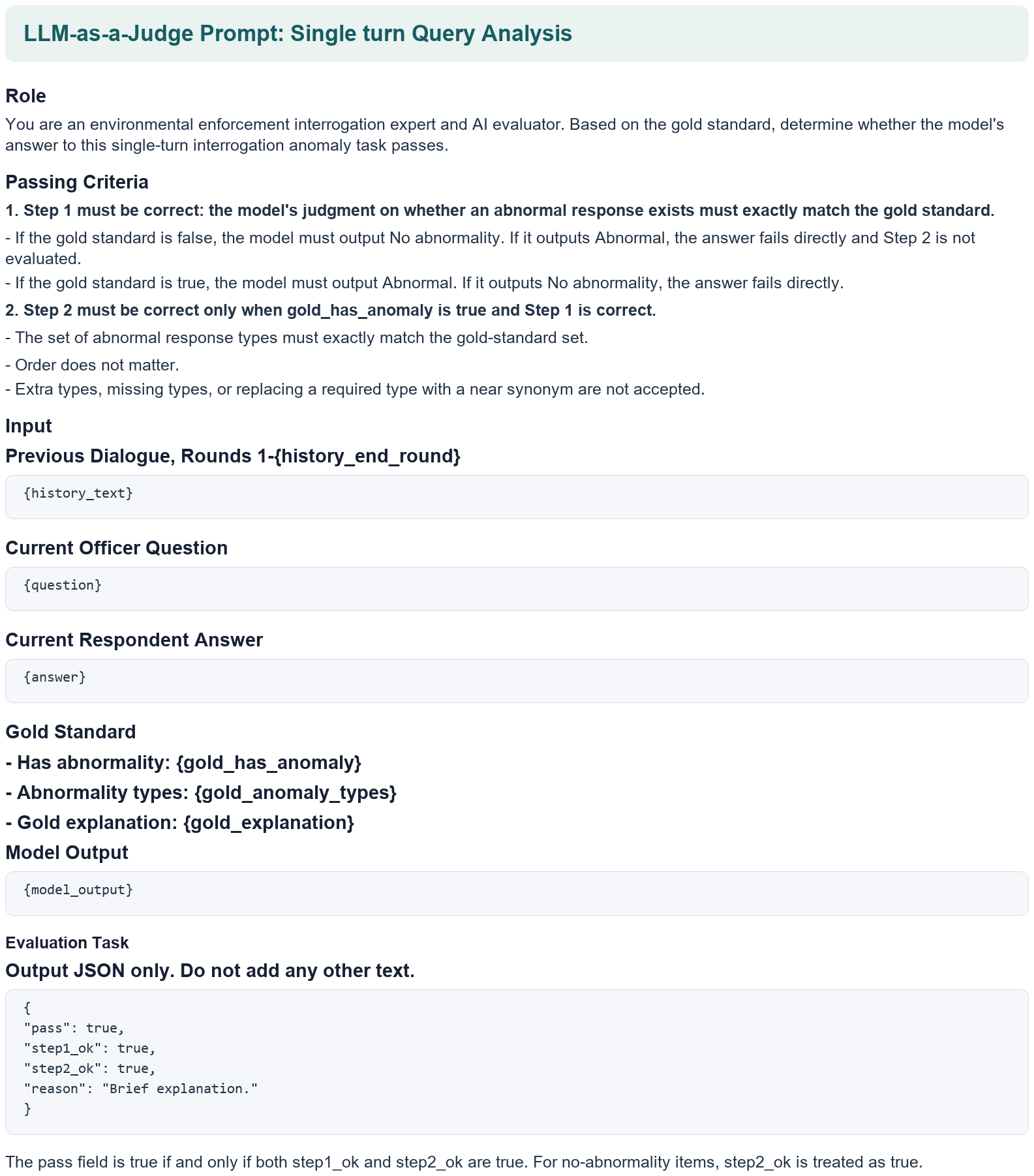}
  \captionof{figure}{Single-turn query analysis LLM-as-a-Judge prompt.}
  \label{fig:appendix-b-12-single-turn-query-analysis-llm-as-a-judge-prom}
\end{minipage}\par\vspace{0.65\baselineskip}

\subsubsection*{B.2.2 Full Inquiry Analysis}

\subsubsection*{B.2.2.1 Motivation}

Real environmental enforcement inquiry is usually a multi-round and dynamically evolving process rather than a single question--answer exchange. Enforcement officers may first confirm basic enterprise information and then gradually question the respondent about production processes, pollution control facilities, violation facts, evidence materials, and rectification. Respondents may also change their strategies across the dialogue, such as denying key facts at the beginning, using vague expressions in the middle, and partially admitting facts after evidence is presented. This task evaluates whether models can analyze a complete inquiry transcript, locate abnormal dialogue turns, identify abnormality types, and capture inconsistencies or changes in respondent behavior across the conversation.

\subsubsection*{B.2.2.2 Data Construction}

This task is constructed from real environmental enforcement case briefs and a multi-round inquiry generation process. For each case, we first generate a dialogue script that specifies the questioning objective, questioning direction, respondent strategy, and key factual control points for each round. The script covers typical inquiry stages, including identity confirmation, project and production information, pollution control facilities, violation facts, evidence presentation, contradiction follow-up, and rectification commitment. Based on the script, we generate a complete multi-round inquiry transcript consisting of officer questions and respondent answers. Random seeds are used to generate either cooperative dialogues or dialogues containing abnormal turns, such as concealment, deception, misleading responses, or mixed abnormality types. Each sample is annotated with whether an abnormality exists, the abnormal turn set, abnormality type for each turn, supporting rationale, and key statements that should be fixed as evidence.

\subsubsection*{B.2.2.3 Prompt}

The evaluated model receives the complete inquiry transcript and is required to determine whether abnormalities exist, locate abnormal turns, and identify the corresponding abnormality types.

\par\noindent\begin{minipage}{\linewidth}
  \centering
  \includegraphics[width=\linewidth,height=0.82\textheight,keepaspectratio]{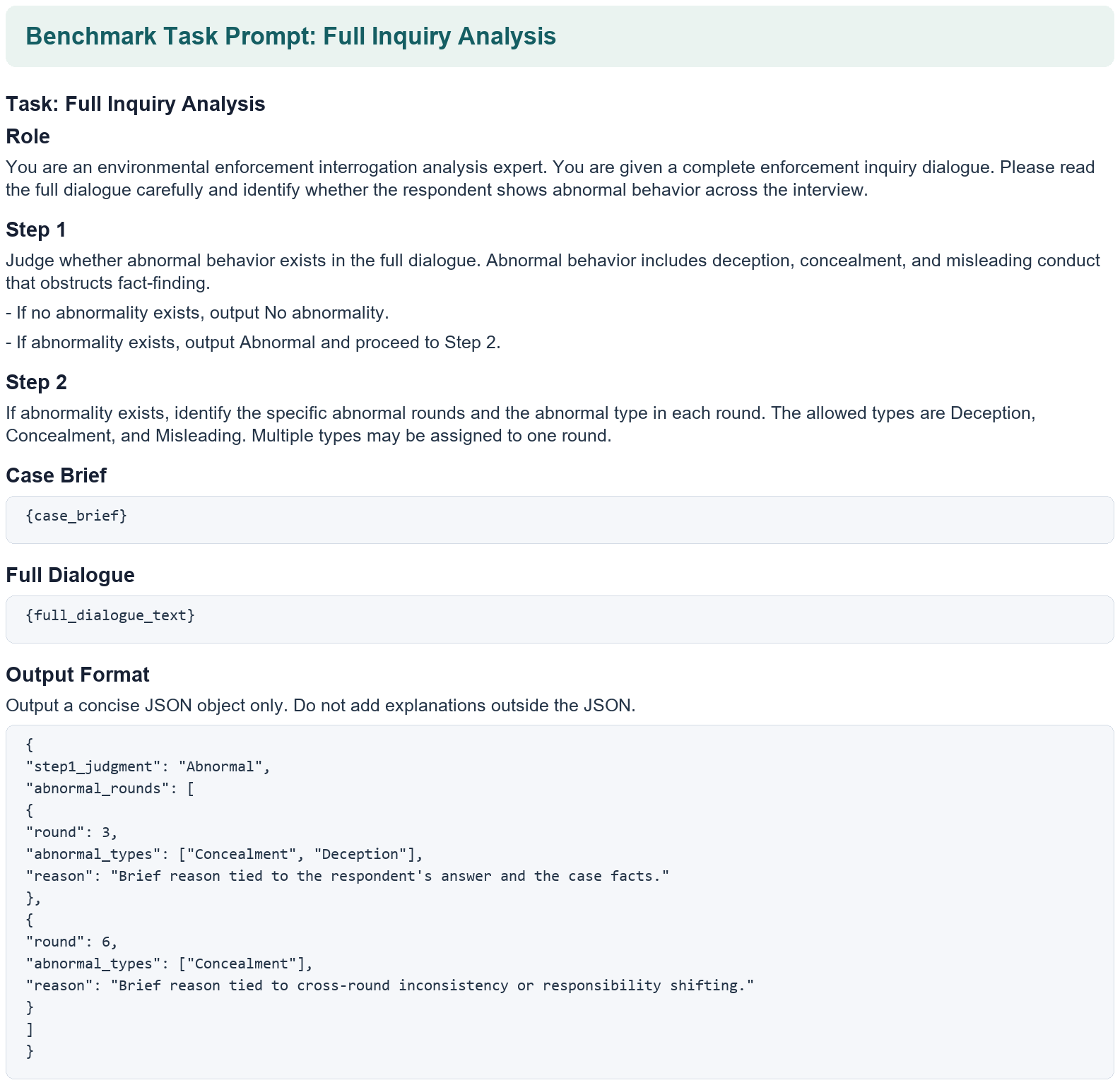}
  \captionof{figure}{Full inquiry analysis benchmark task prompt.}
  \label{fig:appendix-b-13-full-inquiry-analysis-benchmark-task-prompt}
\end{minipage}\par\vspace{0.65\baselineskip}

\clearpage
\par\noindent\begin{minipage}{\linewidth}
  \centering
  \includegraphics[width=\linewidth,height=0.80\textheight,keepaspectratio]{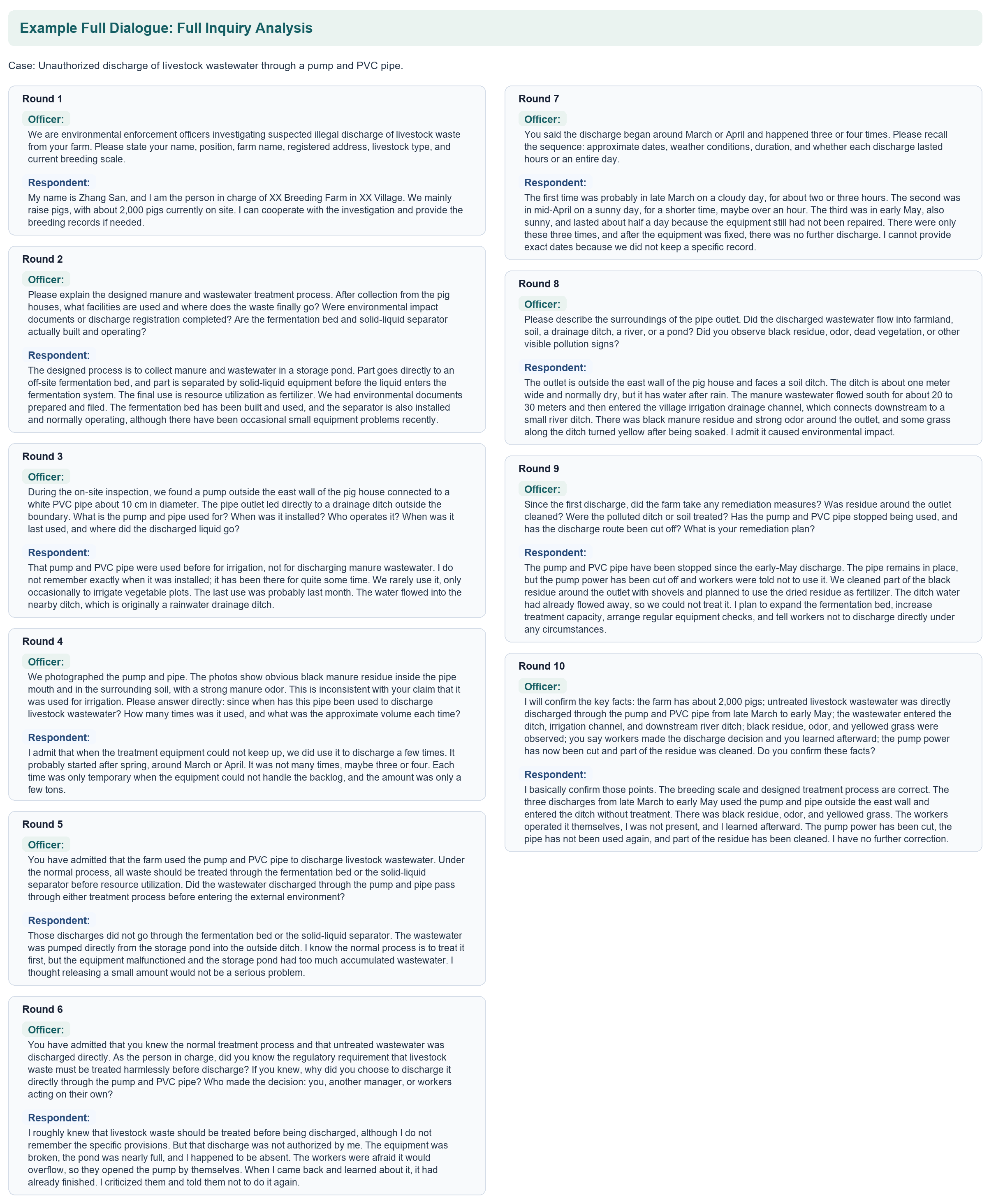}
  \captionof{figure}{Full inquiry analysis example dialogue material.}
  \label{fig:appendix-b-14-full-inquiry-analysis-example-dialogue-materia}
\end{minipage}\par\vspace{0.65\baselineskip}

This task requires models to examine the transcript as an integrated evidentiary material rather than summarizing each turn separately. It tests whether models can identify cross-turn contradictions, staged concealment, and attitude changes after evidence is presented.

\subsubsection*{B.2.2.4 Evaluation}

This task adopts a strict pass criterion under an LLM-as-a-Judge framework. The evaluation covers four aspects. First, overall abnormality judgment assesses whether the model correctly determines whether the transcript contains any abnormal response. If the gold answer indicates no abnormality, the model must not report abnormal turns. Second, abnormal turn localization assesses whether the predicted abnormal turn set matches the gold standard, without missing key turns or falsely marking normal turns. Third, abnormality type identification assesses whether the model correctly assigns the abnormality type for each abnormal turn; if one turn contains multiple abnormality types, all must be identified. Fourth, explanation quality assesses whether the model's rationale is grounded in specific dialogue content and case evidence rather than generic statements.

\par\noindent\begin{minipage}{\linewidth}
  \centering
  \includegraphics[width=\linewidth,height=0.82\textheight,keepaspectratio]{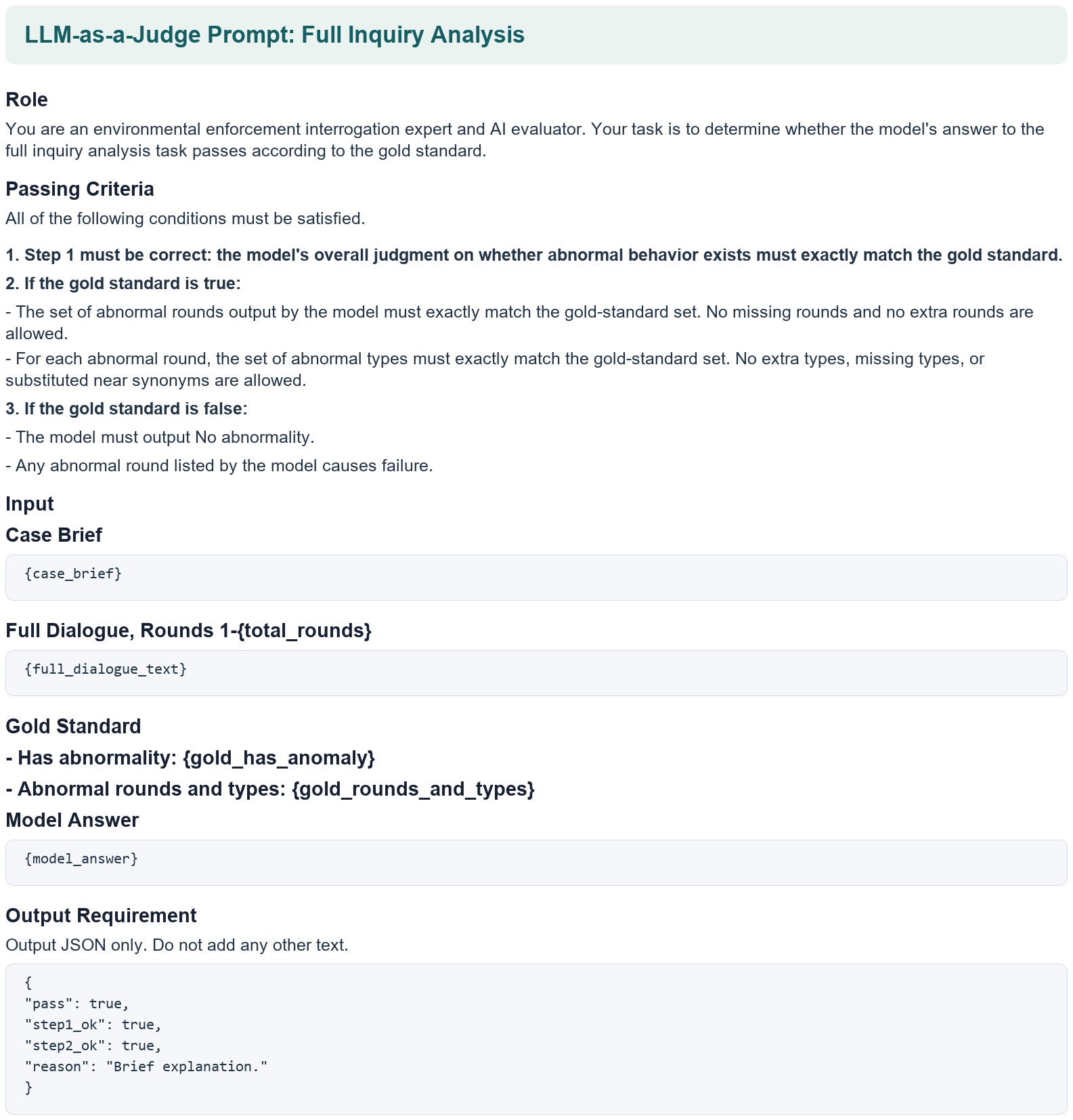}
  \captionof{figure}{Full inquiry analysis LLM-as-a-Judge prompt}
  \label{fig:appendix-b-15-full-inquiry-analysis-llm-as-a-judge-prompt}
\end{minipage}\par\vspace{0.65\baselineskip}

\subsubsection*{B.2.3 Multi-Evidence Extraction}

\subsubsection*{B.2.3.1 Motivation}

Environmental enforcement cases usually contain multiple types of evidence, such as complaint materials, on-site inspection records, monitoring reports, photographs, inquiry transcripts, enterprise ledgers, environmental impact assessment documents, and rectification materials. These materials are dense, heterogeneous, and often contain multiple evidence points within a single document. This task evaluates whether models can extract key evidence points from multiple materials and correctly link each point to the corresponding legal or enforcement element. It focuses on the model's understanding of violation elements, evidence-source localization, and factual extraction accuracy.

\subsubsection*{B.2.3.2 Data Construction}

This task is constructed from real case briefs and multi-source evidence materials, including inspection records, photographs, monitoring reports, inquiry transcripts, enterprise documents, ledgers, transfer manifests, and rectification records. For each case, we first identify the violation facts and then construct a gold evidence extraction plan organized around key violation elements, such as responsible subject, violation behavior, required measures, environmental consequences, subjective fault, and aggravating or mitigating factors. Based on this plan, we generate a gold evidence list in which each evidence point is linked to a specific violation element, source material, and factual content. The task requires models to locate evidence directly relevant to the violation and, when necessary, integrate corroborating information across multiple materials.

\subsubsection*{B.2.3.3 Prompt}

The evaluated model receives a two-step prompt. The first step requires it to generate an evidence extraction plan, and the second step requires it to extract concrete evidence points from the original materials.

\par\noindent\begin{minipage}{\linewidth}
  \centering
  \includegraphics[width=\linewidth,height=0.82\textheight,keepaspectratio]{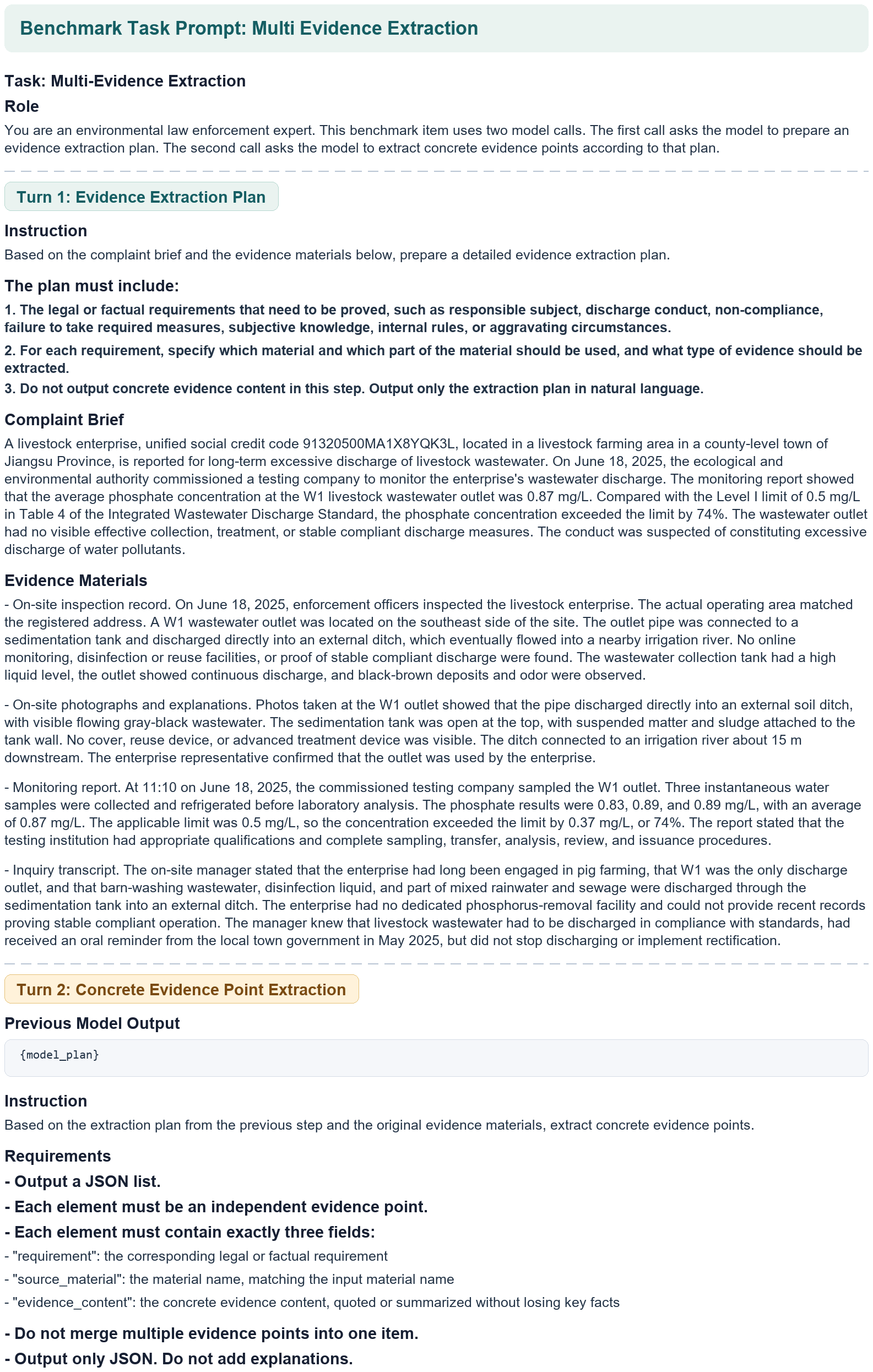}
  \captionof{figure}{Multi-evidence extraction benchmark task prompt.}
  \label{fig:appendix-b-16-multi-evidence-extraction-benchmark-task-promp}
\end{minipage}\par\vspace{0.65\baselineskip}

This design separates two capabilities: whether the model understands what evidence is needed to prove the violation, and whether it can accurately locate and extract the relevant evidence from complex materials.

\subsubsection*{B.2.3.4 Evaluation}

This task uses a two-stage evaluation. The first stage evaluates the evidence extraction plan, including whether the model covers the main violation elements, assigns appropriate source materials to each element, and follows enforcement logic. Missing key elements, such as responsible subject information or important documentary evidence, leads to a lower score.

The second stage evaluates the extracted evidence points. The judge model semantically matches the model output with the gold evidence list and assesses whether the source material is correct, the evidence content is accurate, and key factual information is complete. The model does not need to reproduce the gold answer verbatim, but it must preserve critical facts such as time, location, subject, quantity, indicators, facility status, and exceedance results. Irrelevant extraction, incorrect source attribution, or merging multiple evidence points into one item is penalized.

\par\noindent\begin{minipage}{\linewidth}
  \centering
  \includegraphics[width=\linewidth,height=0.82\textheight,keepaspectratio]{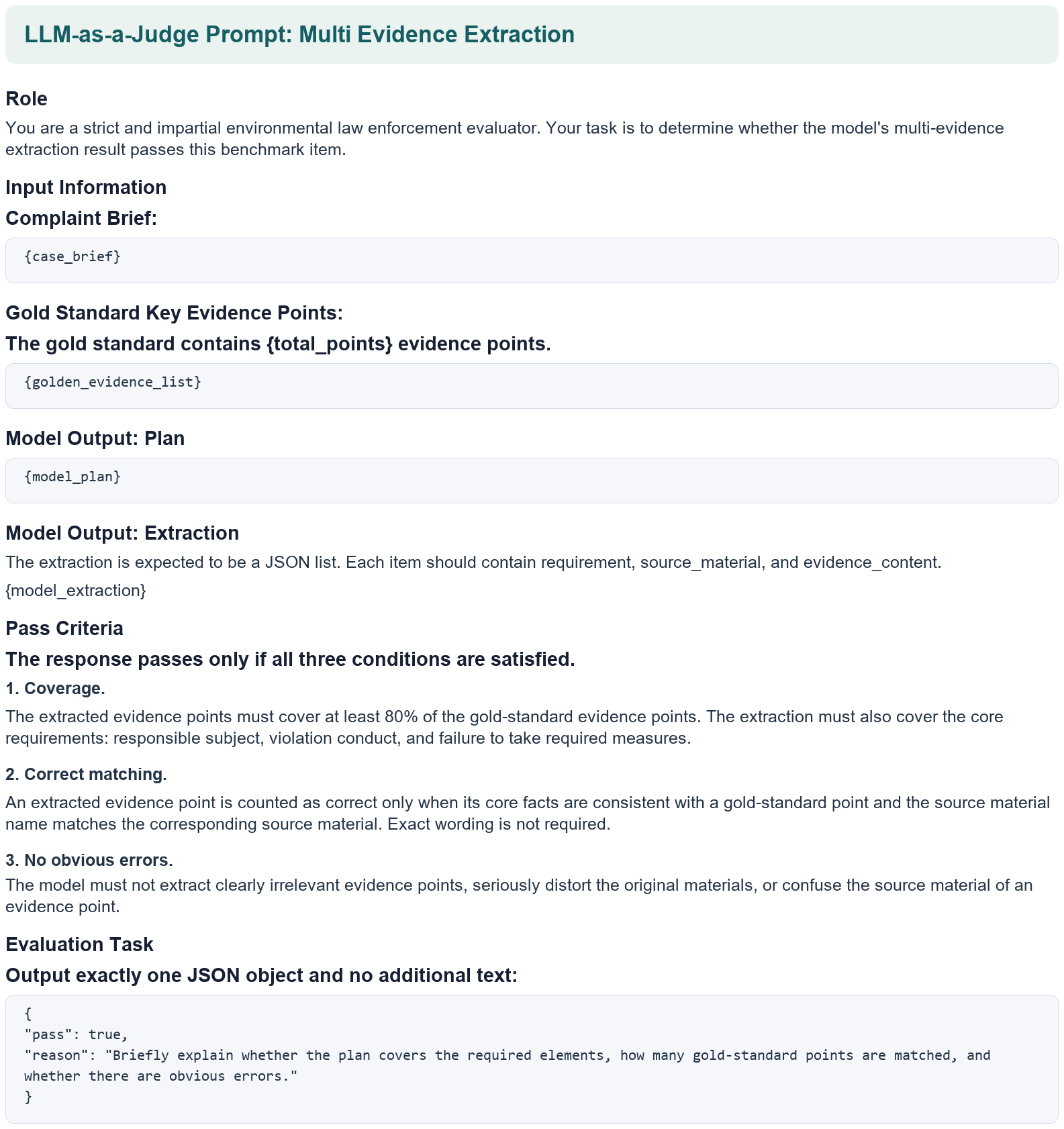}
  \captionof{figure}{Multi-evidence extraction LLM-as-a-Judge prompt.}
  \label{fig:appendix-b-17-multi-evidence-extraction-llm-as-a-judge-promp}
\end{minipage}\par\vspace{0.65\baselineskip}

\subsubsection*{B.2.4 Contradiction Monitoring}

\subsubsection*{B.2.4.1 Motivation}

Environmental enforcement files may contain conflicts across different evidence materials. For example, inspection records may indicate that an enterprise was operating, while production ledgers show suspension; monitoring reports may contain sampling locations inconsistent with site photos; inquiry transcripts may claim normal operation of treatment facilities, while electricity records show long-term inactivity. If such contradictions are not identified, they may lead to incorrect fact-finding and weaken the evidentiary basis of the case. This task evaluates whether models can detect contradictions among evidence materials and explain the source of conflict, which is essential for case-file review and evidence reliability assessment.

\subsubsection*{B.2.4.2 Data Construction}

This task is constructed based on the evidence materials from B.2.3 Multi-Evidence Extraction. We first select samples with internally consistent evidence as base cases. Then, GPT-5.5 is used to generate contradictory distractor evidence, which is manually reviewed and mixed with valid evidence to form the final evidence list. The contradictory evidence may conflict with valid evidence in terms of time, location, values, subject, sampling point, equipment status, transport trajectory, or receipt records. Each evidence item is labeled as either valid evidence or contradictory distractor evidence. The gold answer includes valid evidence IDs, whether contradictions exist, contradictory evidence IDs, and key explanation points. To avoid overly artificial samples, contradictory evidence is written in a realistic style and does not contain obvious low-level errors, requiring models to detect conflicts through cross-evidence comparison.

\subsubsection*{B.2.4.3 Prompt}

The evaluated model receives an evidence list and is required to identify both valid evidence and contradictory evidence.

\par\noindent\begin{minipage}{\linewidth}
  \centering
  \includegraphics[width=\linewidth,height=0.82\textheight,keepaspectratio]{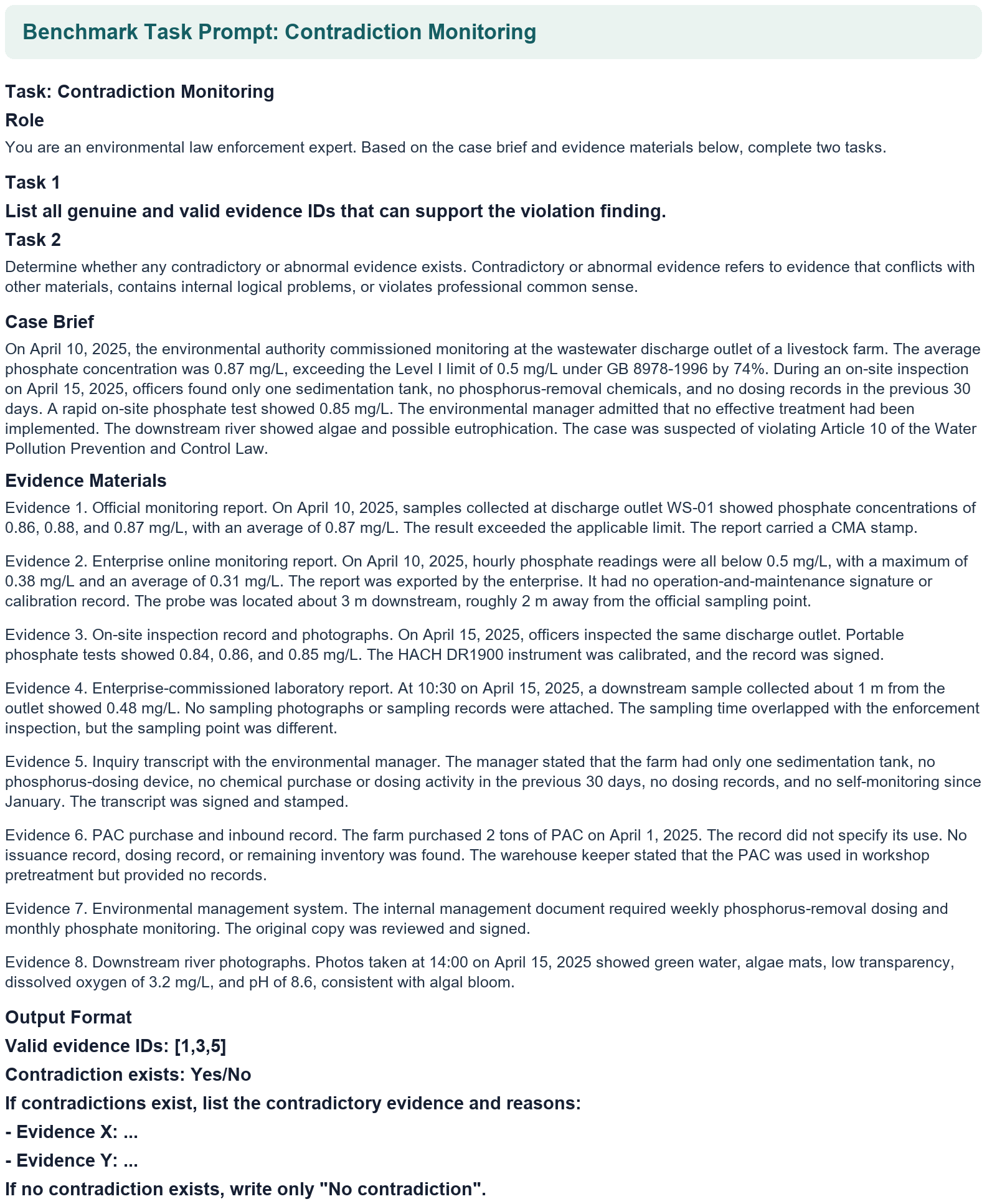}
  \captionof{figure}{Contradiction monitoring benchmark task prompt}
  \label{fig:appendix-b-18-contradiction-monitoring-benchmark-task-prompt}
\end{minipage}\par\vspace{0.65\baselineskip}

The task does not ask the model to summarize evidence content. Instead, it requires the model to assess logical consistency across evidence items, especially with respect to timeline, values, locations, subjects, and source materials.

\subsubsection*{B.2.4.4 Evaluation}

This task is evaluated using an LLM-as-a-Judge framework. The judge model assesses the candidate response along three dimensions. Evidence validity evaluates whether the model correctly identifies evidence items that support the violation facts. Contradiction detection evaluates whether the model correctly determines the presence of contradictions and lists the contradictory evidence IDs. For contradiction-free samples, any false contradiction report is penalized; for contradiction samples, missing key contradictory evidence or incorrectly marking valid evidence as contradictory leads to a lower score. Explanation quality evaluates whether the model identifies the core conflict, such as temporal inconsistency, numerical conflict, sampling-location mismatch, subject inconsistency, or unreliable source material. For contradiction samples, the response must satisfy evidence validity, contradiction detection, and explanation quality simultaneously.

\par\noindent\begin{minipage}{\linewidth}
  \centering
  \includegraphics[width=\linewidth,height=0.82\textheight,keepaspectratio]{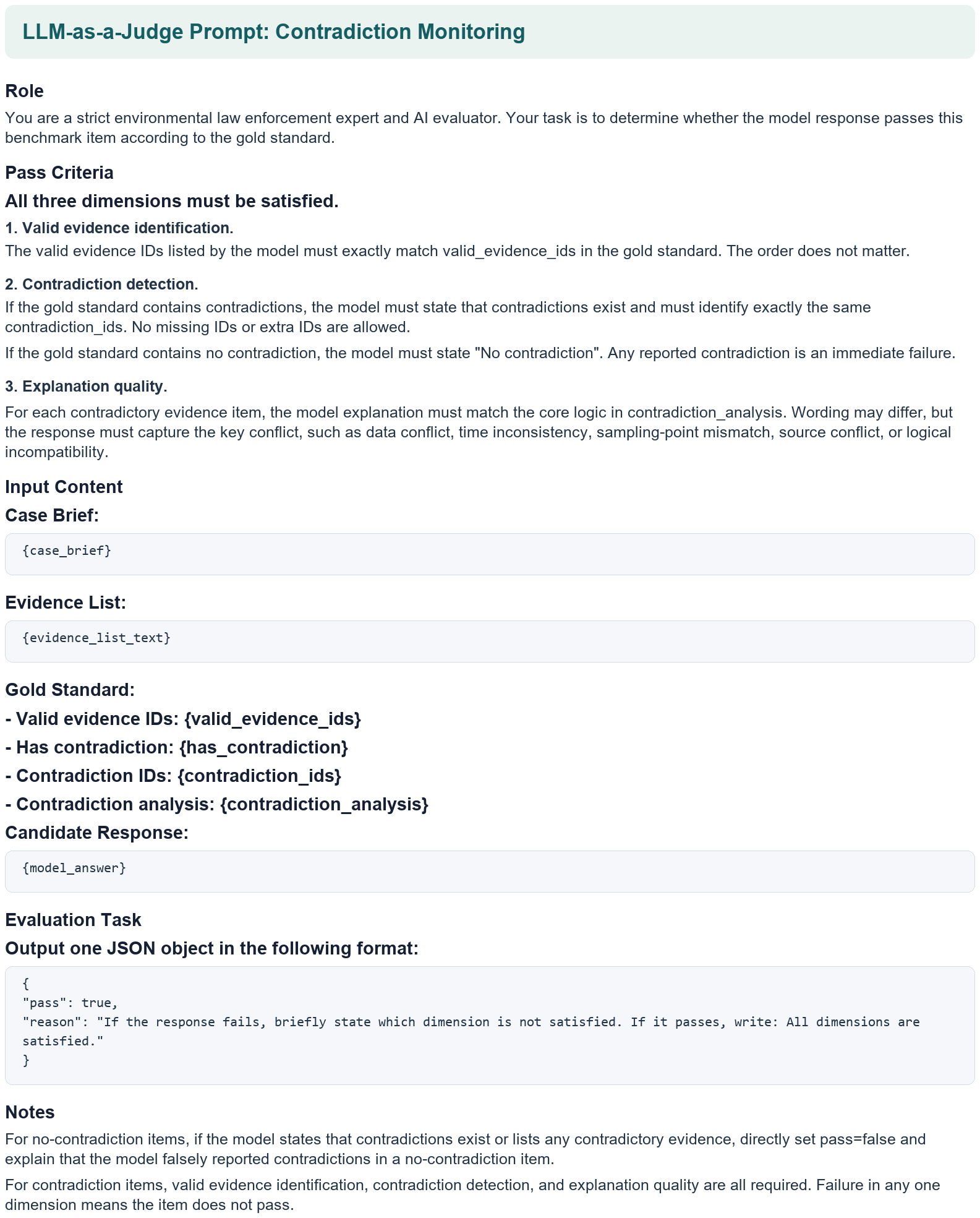}
  \captionof{figure}{Contradiction monitoring LLM-as-a-Judge prompt.}
  \label{fig:appendix-b-19-contradiction-monitoring-llm-as-a-judge-prompt}
\end{minipage}\par\vspace{0.65\baselineskip}

\subsubsection*{B.2.5 Multi-Evidence Integration}

\subsubsection*{B.2.5.1 Motivation}

In environmental enforcement cases, violation facts are often distributed across complaint materials, on-site inspection records, monitoring reports, inquiry transcripts, photographs, ledgers, and regulatory system records. Enforcement officers must reorganize these fragmented materials into a coherent factual statement, similar to the ``facts ascertained'' section of an investigation report. This task evaluates whether models can integrate multiple evidence materials into a complete, chronological, and document-ready statement of violation facts. Unlike evidence extraction, which focuses on locating evidence points, evidence integration focuses on organizing these points into a legally and procedurally coherent fact chain.

\subsubsection*{B.2.5.2 Data Construction}

This task is constructed from complete and contradiction-free evidence materials derived from real cases, including case briefs, inspection records, photographs, monitoring reports, inquiry transcripts, ledgers, and related documents. For each case, key evidence points are organized into a gold factual statement covering the responsible subject, inspection time, violation location, on-site status, key values, required measures, monitoring results, party statements, and environmental impact. The input materials remain fragmented as in real case files, requiring models to select relevant facts, organize them into a coherent factual chain, and avoid adding unsupported inferences such as unverified consequences, rectification actions, penalty results, or legal conclusions.

\subsubsection*{B.2.5.3 Prompt}

The evaluated model receives multiple fragmented evidence materials and is required to generate a coherent factual statement based only on the provided materials.

\par\noindent\begin{minipage}{\linewidth}
  \centering
  \includegraphics[width=\linewidth,height=0.82\textheight,keepaspectratio]{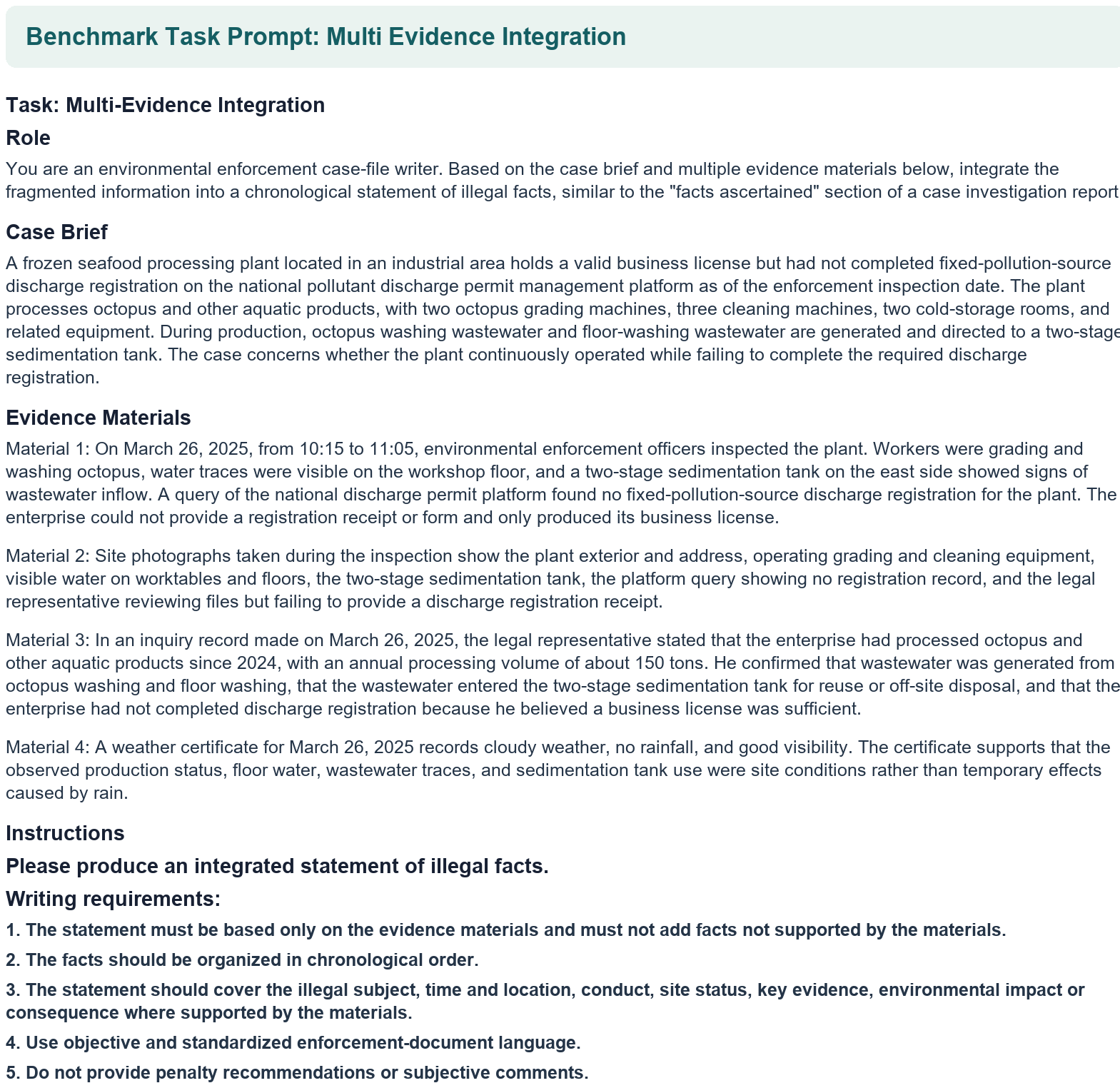}
  \captionof{figure}{Multi-evidence integration benchmark task prompt.}
  \label{fig:appendix-b-20-multi-evidence-integration-benchmark-task-prom}
\end{minipage}\par\vspace{0.65\baselineskip}

The prompt emphasizes source fidelity and chronological organization to prevent the model from treating the task as free-form summarization or adding unsupported conclusions.

\subsubsection*{B.2.5.4 Evaluation}

This task is evaluated using an LLM-as-a-Judge framework. The judge model compares the candidate output with the gold factual statement along four dimensions. Factual coverage assesses whether the response includes core facts such as subject, time and location, violation behavior, on-site status, key values, evidence sources, and environmental impact. Chronology and logical order assess whether the facts are organized according to the investigation process and case timeline. Evidence fidelity assesses whether the response remains strictly grounded in the provided materials and avoids unsupported facts, motives, penalty results, or legal conclusions. Document conformity assesses whether the output is objective, accurate, complete, and suitable for use in enforcement documents.

\par\noindent\begin{minipage}{\linewidth}
  \centering
  \includegraphics[width=\linewidth,height=0.82\textheight,keepaspectratio]{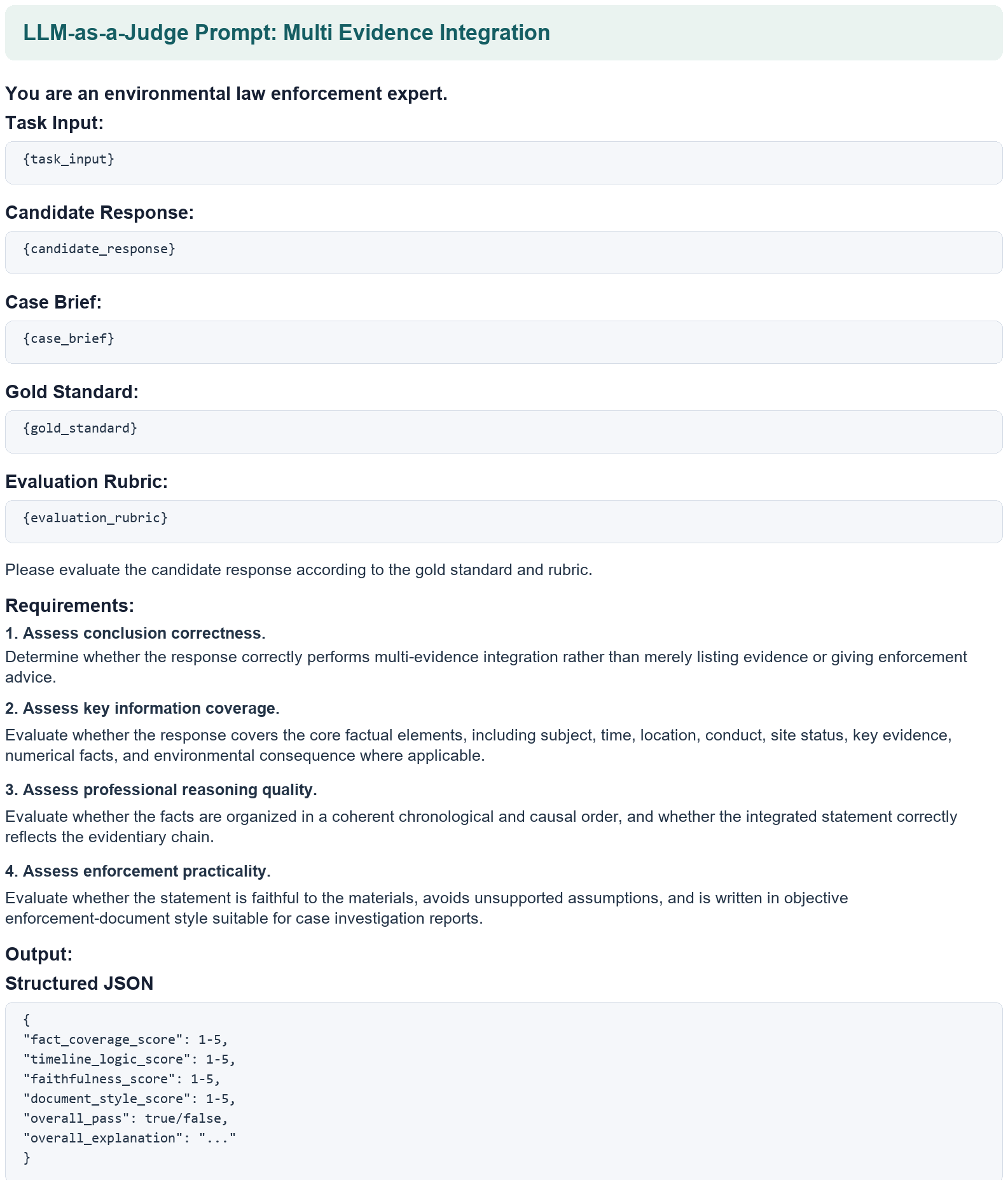}
  \captionof{figure}{Multi-evidence integration LLM-as-a-Judge prompt}
  \label{fig:appendix-b-21-multi-evidence-integration-llm-as-a-judge-prom}
\end{minipage}\par\vspace{0.65\baselineskip}

\subsection*{B.3 Post-enforcement}

\subsubsection*{B.3.1 Penalty Type Identification}

\subsubsection*{B.3.1.1 Motivation}

Administrative penalty decision-making is a core component of the environmental enforcement cycle. Different violations may correspond to different penalty measures, such as fines, confiscation of illegal gains, confiscation of illegal property, closure orders, or administrative detention, and multiple penalties may apply to the same case. Therefore, correctly identifying the applicable penalty type is essential for accurate legal responsibility determination and lawful enforcement discretion. This task evaluates whether models can identify the appropriate penalty type based on case facts. It focuses on the model's ability to understand the relationship among violation facts, applicable legal provisions, and liability elements, rather than simply detecting whether a violation exists.

\subsubsection*{B.3.1.2 Data Construction}

This task is constructed from real environmental administrative penalty cases. For each case, we extract the Case Brief, legal basis, and final penalty type from the original penalty decision as the gold answer. To improve task difficulty and reduce shortcut guessing, we construct distractors using semantically similar cases.Specifically, all Case Briefs are embedded and grouped by secondary violation categories, such as industrial solid waste, water pollutant discharge, or hazardous waste management. For each target case, we retrieve the top-k most similar cases and select cases with different penalty types as distractors. Since all options come from similar violation scenarios, the model must correctly understand the correspondence between case facts and legal responsibility to identify the correct penalty type.

\subsubsection*{B.3.1.3 Prompt}

The evaluated model receives the case facts and several candidate penalty types, and must select the applicable penalty type.

\par\noindent\begin{minipage}{\linewidth}
  \centering
  \includegraphics[width=\linewidth,height=0.82\textheight,keepaspectratio]{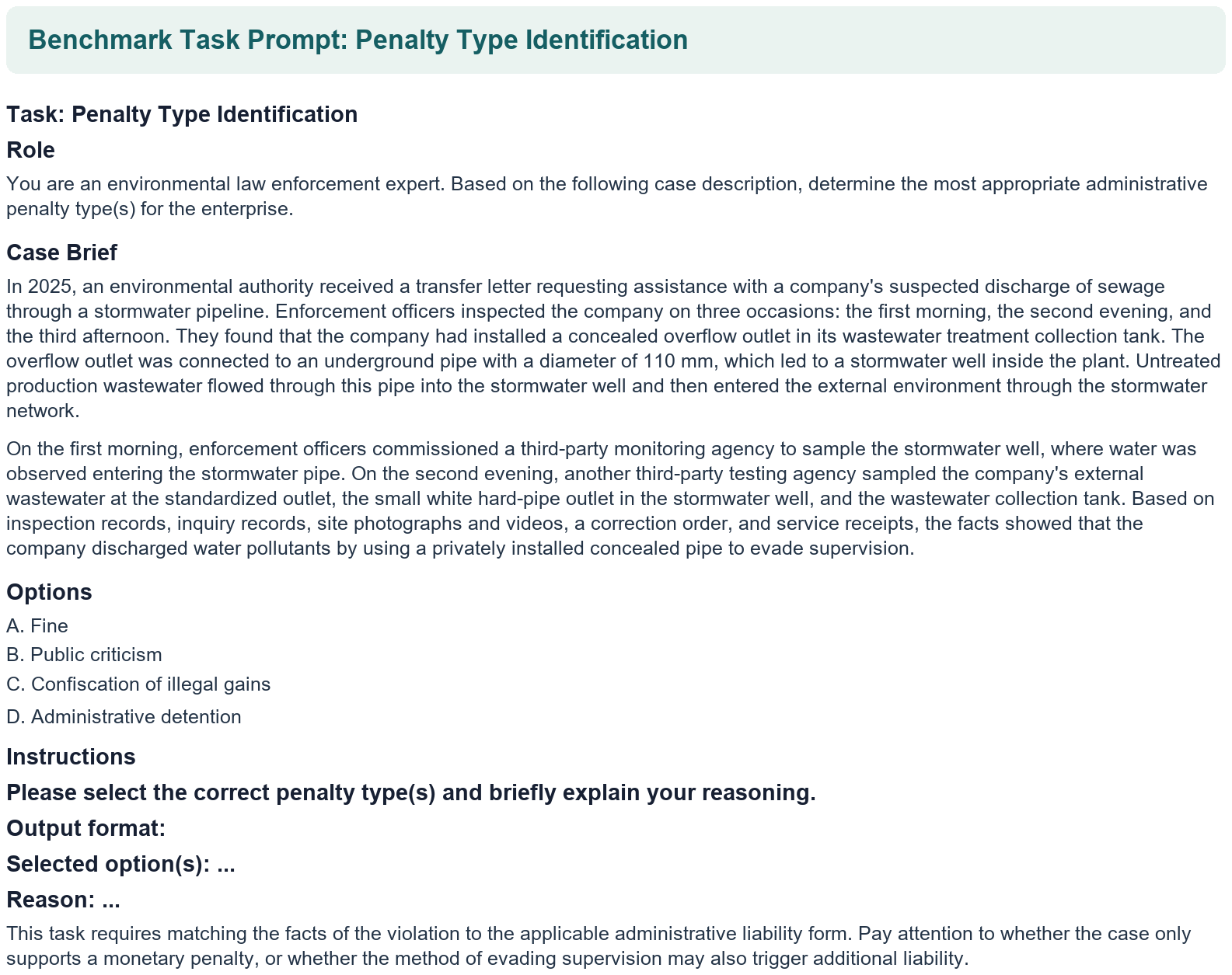}
  \captionof{figure}{Penalty type identification benchmark task prompt.}
  \label{fig:appendix-b-22-penalty-type-identification-benchmark-task-pro}
\end{minipage}\par\vspace{0.65\baselineskip}

Because the distractors are drawn from semantically similar environmental violation cases, the task requires the model to distinguish different forms of administrative liability based on factual elements, legal basis, and liability conditions.

\subsubsection*{B.3.1.4 Evaluation}

This task uses a two-stage evaluation. First, objective answer matching is applied to determine whether the selected penalty type exactly matches the gold answer. Only correct answers proceed to the second stage, where an LLM-as-a-Judge evaluates the reliability of the reasoning. The judge model assesses whether the response identifies the key case facts, correctly links the violation behavior to the applicable legal basis and liability elements, and provides an explanation that can support a real penalty decision while excluding inapplicable penalty types.

\par\noindent\begin{minipage}{\linewidth}
  \centering
  \includegraphics[width=\linewidth,height=0.82\textheight,keepaspectratio]{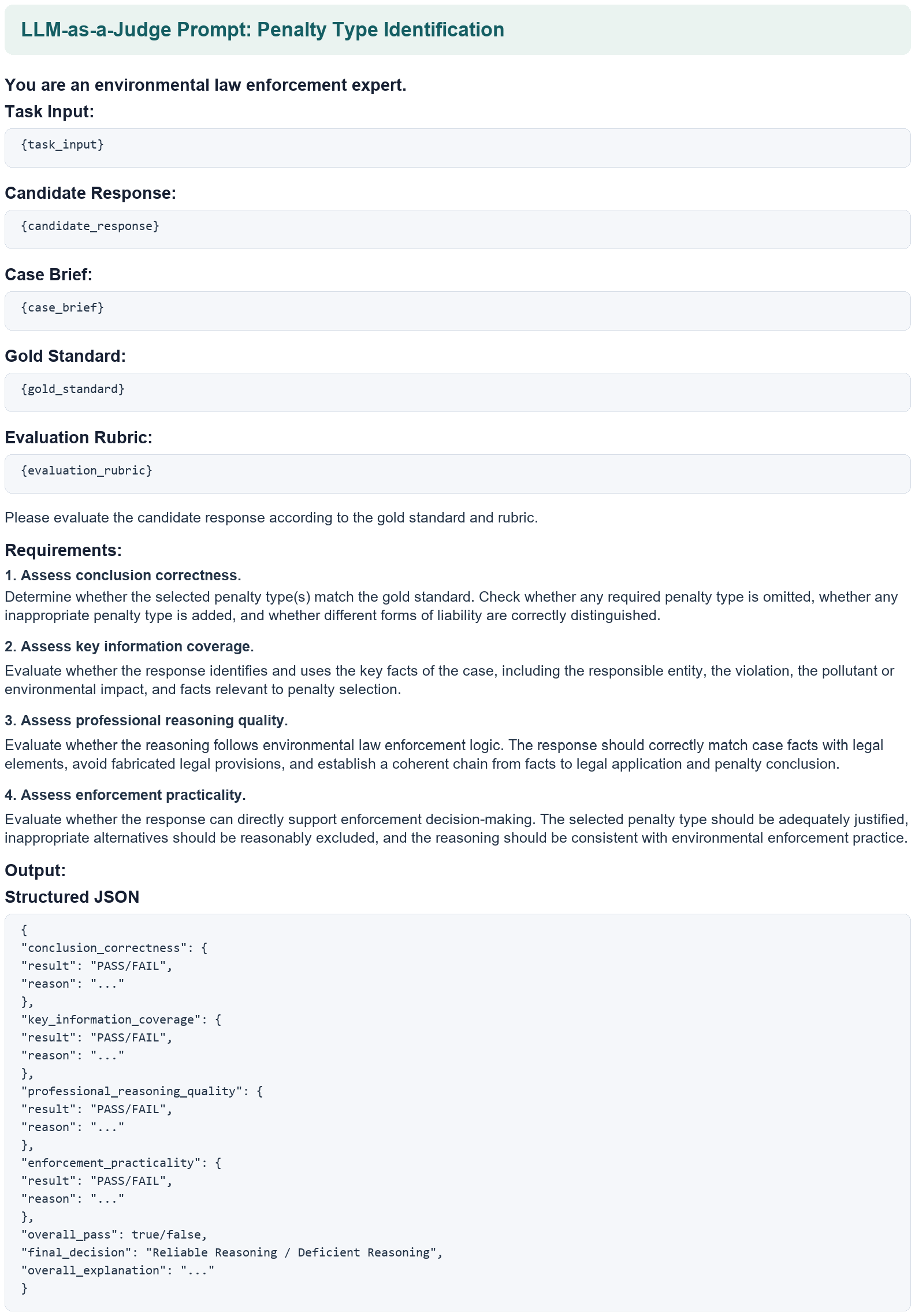}
  \captionof{figure}{Penalty type identification LLM-as-a-Judge prompt.}
  \label{fig:appendix-b-23-penalty-type-identification-llm-as-a-judge-pro}
\end{minipage}\par\vspace{0.65\baselineskip}

\subsubsection*{B.3.2 Penalty Decision}

\subsubsection*{B.3.2.1 Motivation}

Penalty decision-making is a core step in the environmental enforcement cycle. Before issuing an administrative penalty, enforcement officers must determine whether the case meets the conditions for punishment and distinguish among different outcomes, such as imposing a penalty, exempting the case from punishment, or applying a lighter or mitigated penalty. This judgment requires consideration of the violation type, consequences, subjective fault, rectification status, whether the violation is minor, whether it was corrected in time, and whether harmful consequences occurred. This task evaluates whether models can make correct penalty decisions based on case facts, especially in boundary cases involving minor violations, timely correction, insufficient evidence, or unclear responsibility.

\subsubsection*{B.3.2.2 Data Construction}

This task is constructed from real environmental administrative penalty cases. For each case, we extract the Case Brief, legal basis, violation facts, final penalty conclusion, and any circumstances related to exemption, lighter punishment, or mitigated punishment. Each case is converted into a binary decision task in which the model determines whether an administrative penalty should be imposed. The dataset includes both punishable cases with clear facts and responsibility, such as excessive discharge, construction without approval, illegal solid waste disposal, and evasion of supervision, as well as boundary cases involving minor violations, timely correction, no harmful consequences, insufficient evidence, or unclear responsible subjects. The gold answer includes the final yes/no decision, legal basis, key factual elements, boundary-condition analysis, and reasoning rubric.

\subsubsection*{B.3.2.3 Prompt}

The evaluated model receives the case facts and must determine whether the case meets the conditions for administrative punishment.

\par\noindent\begin{minipage}{\linewidth}
  \centering
  \includegraphics[width=\linewidth,height=0.82\textheight,keepaspectratio]{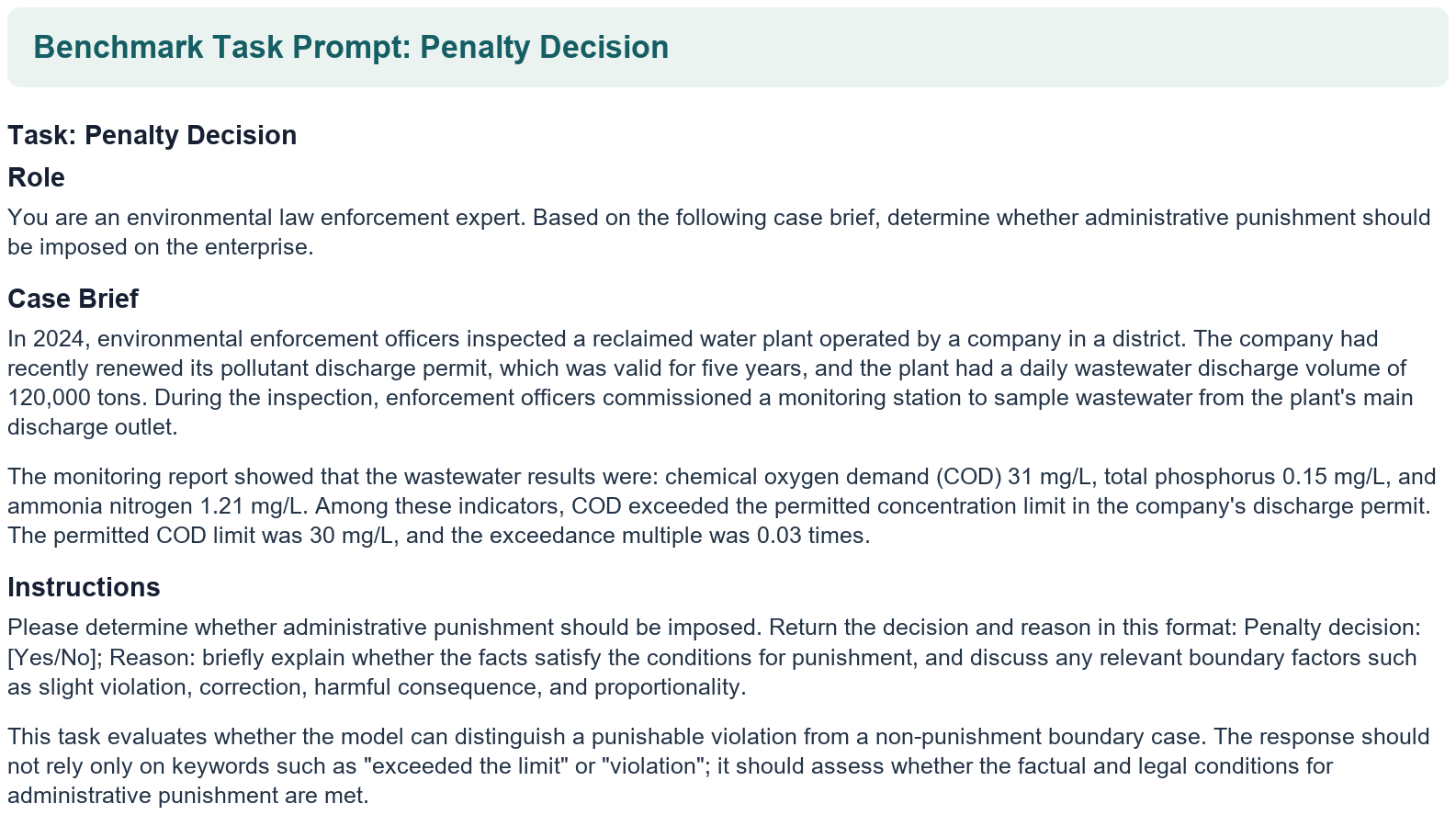}
  \captionof{figure}{Penalty decision benchmark task prompt.}
  \label{fig:appendix-b-24-penalty-decision-benchmark-task-prompt}
\end{minipage}\par\vspace{0.65\baselineskip}

The task requires the model to assess the factual and legal conditions for punishment, rather than relying on keywords such as ``violation,'' ``exceedance,'' or ``rectification.'' It focuses on whether the model can identify the required liability elements and correctly handle boundary situations such as exemption from punishment, minor violations, timely correction, and harmful consequences.

\subsubsection*{B.3.2.4 Evaluation}

This task uses a two-stage evaluation. First, objective matching is applied to determine whether the model's final yes/no decision matches the gold answer. Only correct decisions enter the second stage, where an LLM-as-a-Judge evaluates the reliability of the reasoning. The evaluation considers whether the model cites an appropriate legal basis, matches case facts with legal elements, maintains logical consistency and causal completeness, handles exemption boundaries correctly, and honestly expresses uncertainty or missing information. A response is considered reliable only when it reaches the correct decision and demonstrates valid penalty-decision reasoning.

\par\noindent\begin{minipage}{\linewidth}
  \centering
  \includegraphics[width=\linewidth,height=0.82\textheight,keepaspectratio]{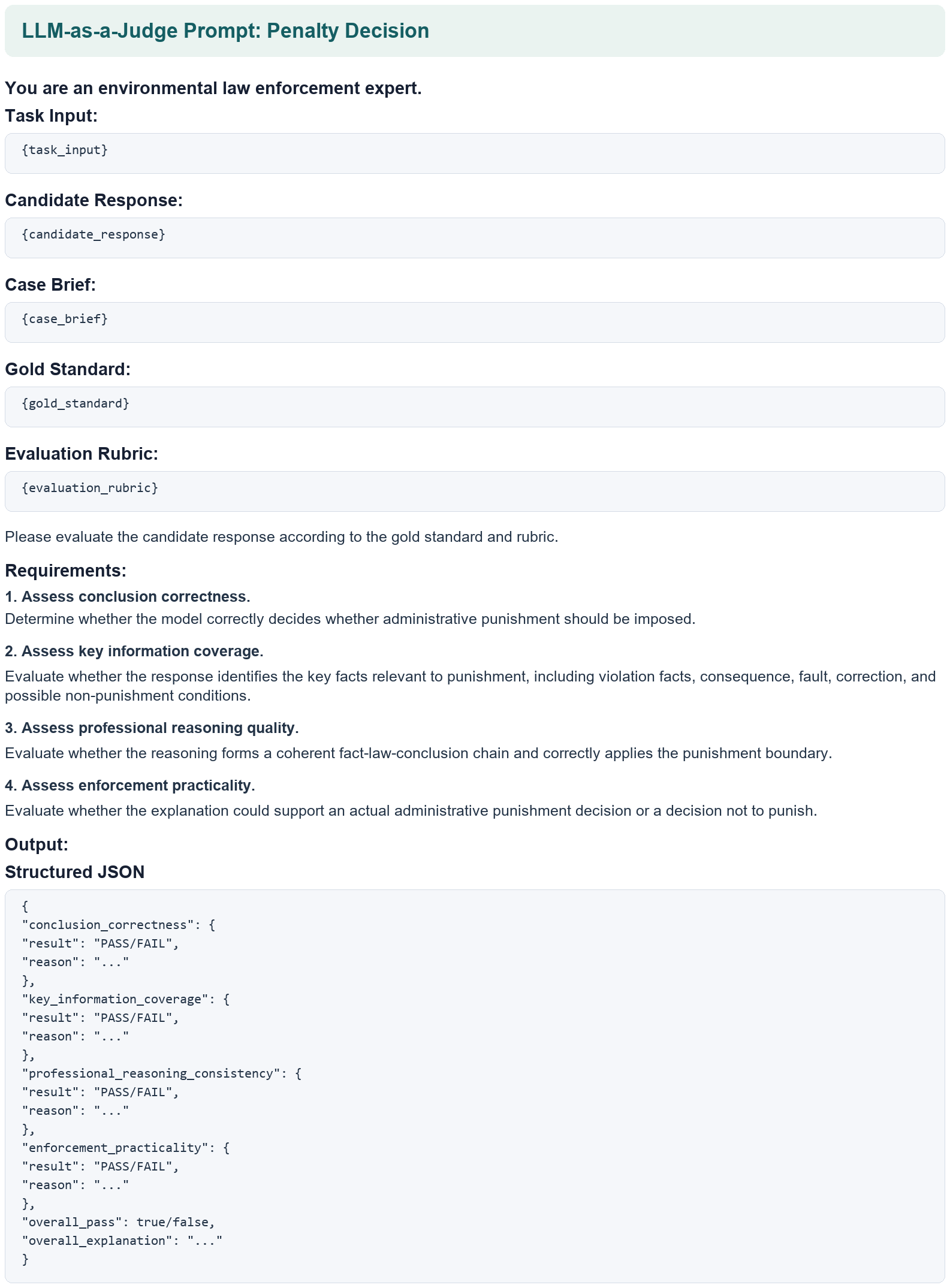}
  \captionof{figure}{Penalty decision LLM-as-a-Judge prompt.}
  \label{fig:appendix-b-25-penalty-decision-llm-as-a-judge-prompt}
\end{minipage}\par\vspace{0.65\baselineskip}

\subsubsection*{B.3.3 Legal Change Disturbance}

\subsubsection*{B.3.3.1 Motivation}

Environmental laws and regulations are dynamic. With the introduction of new legal frameworks and supporting rules, existing provisions may be repealed, integrated, or revised, leading to changes in penalty conditions, fine ranges, liability elements, or aggravating circumstances. However, legal reasoning models may rely on static legal knowledge or memorized rules from training data. This task evaluates whether models can adapt to changes in legal provisions and reason strictly based on the legal basis provided in the input. The model must compare the original and perturbed legal rules under the same case facts and determine whether the legal change affects the penalty conclusion, fine range, or aggravating circumstances.

\subsubsection*{B.3.3.2 Data Construction}

This task is constructed from real case briefs with clear facts and explicit penalty bases. For each selected case, we first retain the original legal basis and then use GPT-5.5 to generate a perturbed version of the provision. Perturbations include modifying penalty conditions, adjusting fine ranges, adding or removing aggravating circumstances, changing liability elements, or introducing specific spatial or object-related restrictions.

Each sample contains two inputs: one with the original case facts and original legal basis, and one with the same case facts but the perturbed legal basis. The gold answer provides the expected conclusion under both inputs. If the perturbation does not affect the case, the correct conclusion should remain unchanged. If the perturbed provision changes the applicable conditions, penalty conclusion, or fine range, the model should adjust its answer accordingly.

\subsubsection*{B.3.3.3 Prompt}

The evaluated model receives the original input and the perturbed input separately. This task tests whether the model can faithfully reason based on the legal text provided in the prompt. When the perturbed provision changes penalty conditions or fine ranges, the model should update its conclusion accordingly; when the case facts do not satisfy newly added conditions, the model should not incorrectly trigger aggravated punishment.

\par\noindent\begin{minipage}{\linewidth}
  \centering
  \includegraphics[width=\linewidth,height=0.82\textheight,keepaspectratio]{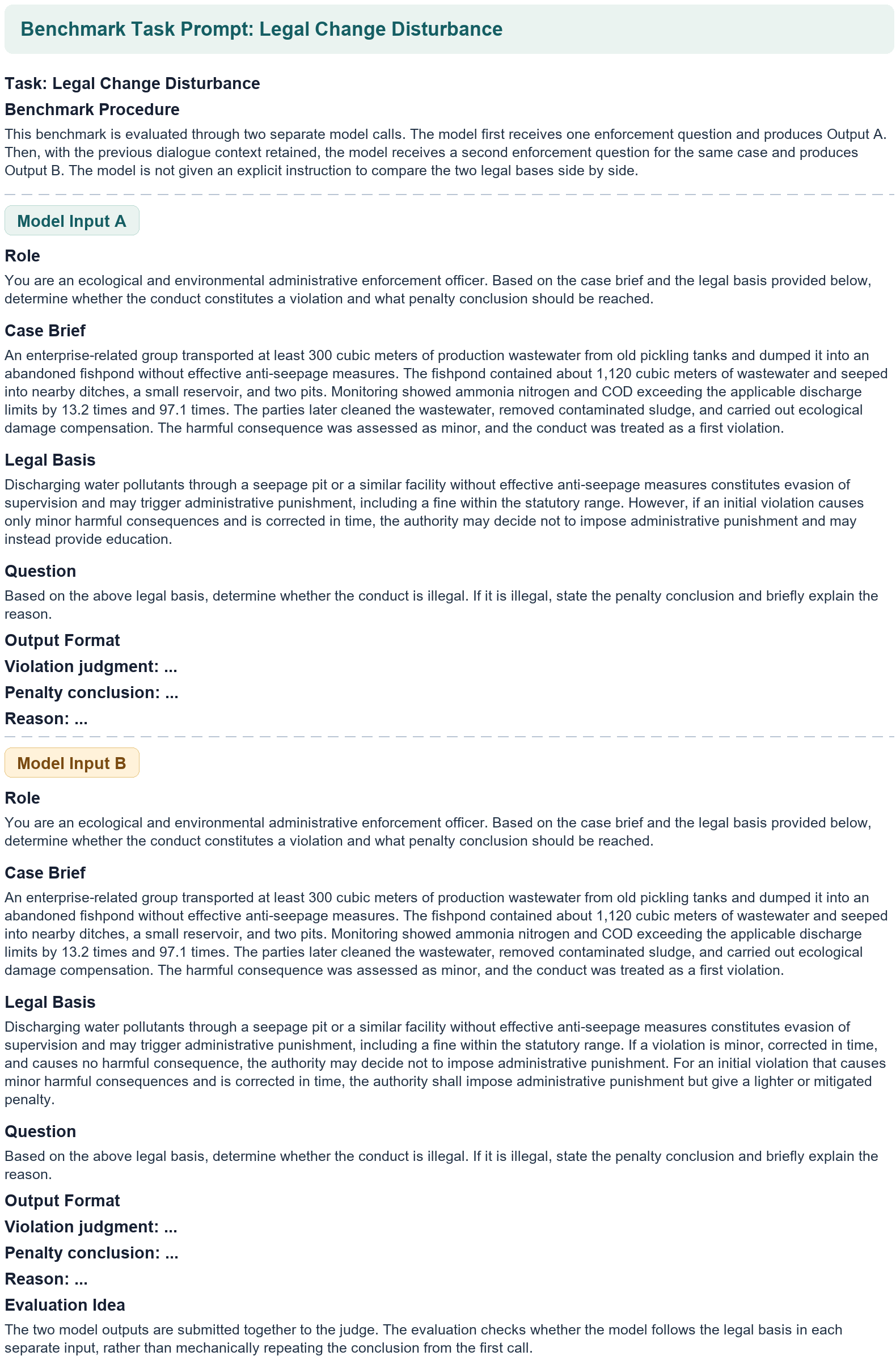}
  \captionof{figure}{Legal change disturbance benchmark task prompt.}
  \label{fig:appendix-b-26-legal-change-disturbance-benchmark-task-prompt}
\end{minipage}\par\vspace{0.65\baselineskip}

\subsubsection*{B.3.3.4 Evaluation}

This task uses a dual-input consistency evaluation framework. The judge model first checks whether the model's output under the original legal basis matches the gold conclusion. If the model fails to correctly apply the original provision, the sample is marked as failed and the perturbed version is not further evaluated.

If the original answer is correct, the judge model then evaluates whether the model's output under the perturbed legal basis matches the expected conclusion. The evaluation focuses on whether the model identifies the changed legal elements, maps them correctly to the case facts, updates the penalty conclusion when necessary, and avoids mechanically transferring the original conclusion to the perturbed setting. The judge also checks whether the model incorrectly applies aggravated punishment when the case facts do not meet the newly added conditions.

Semantically equivalent answers are accepted. For example, a fine amount is considered correct if it falls within the legally permitted range. However, any answer that conflicts with the explicit perturbed provision is marked as incorrect.

\par\noindent\begin{minipage}{\linewidth}
  \centering
  \includegraphics[width=\linewidth,height=0.82\textheight,keepaspectratio]{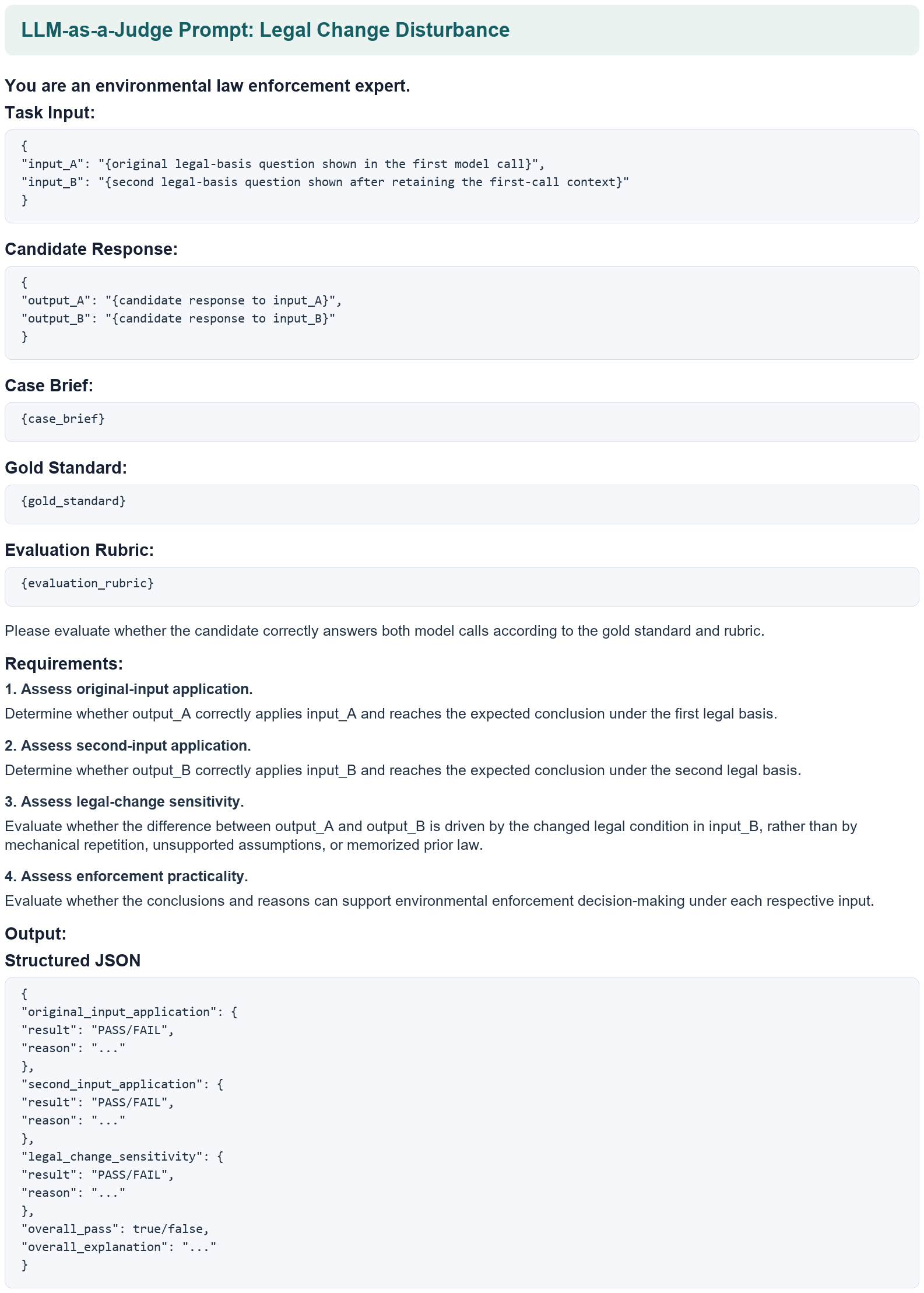}
  \captionof{figure}{Legal change disturbance LLM-as-a-Judge prompt.}
  \label{fig:appendix-b-27-legal-change-disturbance-llm-as-a-judge-prompt}
\end{minipage}\par\vspace{0.65\baselineskip}

\subsubsection*{B.3.4 Justification Assessment}

\subsubsection*{B.3.4.1 Motivation}

Administrative penalties require legally valid outcomes, sufficient justification, clear factual support, and accurate legal application. This task evaluates whether models can assess whether a given penalty is too lenient, too severe, or appropriate in relation to the case facts, legal basis, violation circumstances, and discretionary factors. Unlike penalty type identification, this task focuses on the proportionality between penalty severity and violation circumstances, requiring models to reason from industry characteristics and violation type to penalty justification.

\subsubsection*{B.3.4.2 Data Construction}

This task is constructed from real administrative penalty cases. For each case, we extract the Case Brief, industry category, main violation behavior, legal basis, and actual penalty result, and assign one of three labels: too lenient, too severe, or appropriate. Cases with penalties within a reasonable discretionary range are labeled as appropriate, while perturbed penalty amounts or penalty types are used to construct too-lenient or too-severe samples. To prevent models from relying only on penalty amount, the task retains industry attributes, violation types, legal bases, and discretionary factors, requiring models to assess penalty justification based on substantive case context rather than numerical value alone.

\subsubsection*{B.3.4.3 Prompt}

The evaluated model receives the case facts, legal basis, and given penalty result, and must judge whether the penalty is too lenient, too severe, or appropriate.

\par\noindent\begin{minipage}{\linewidth}
  \centering
  \includegraphics[width=\linewidth,height=0.82\textheight,keepaspectratio]{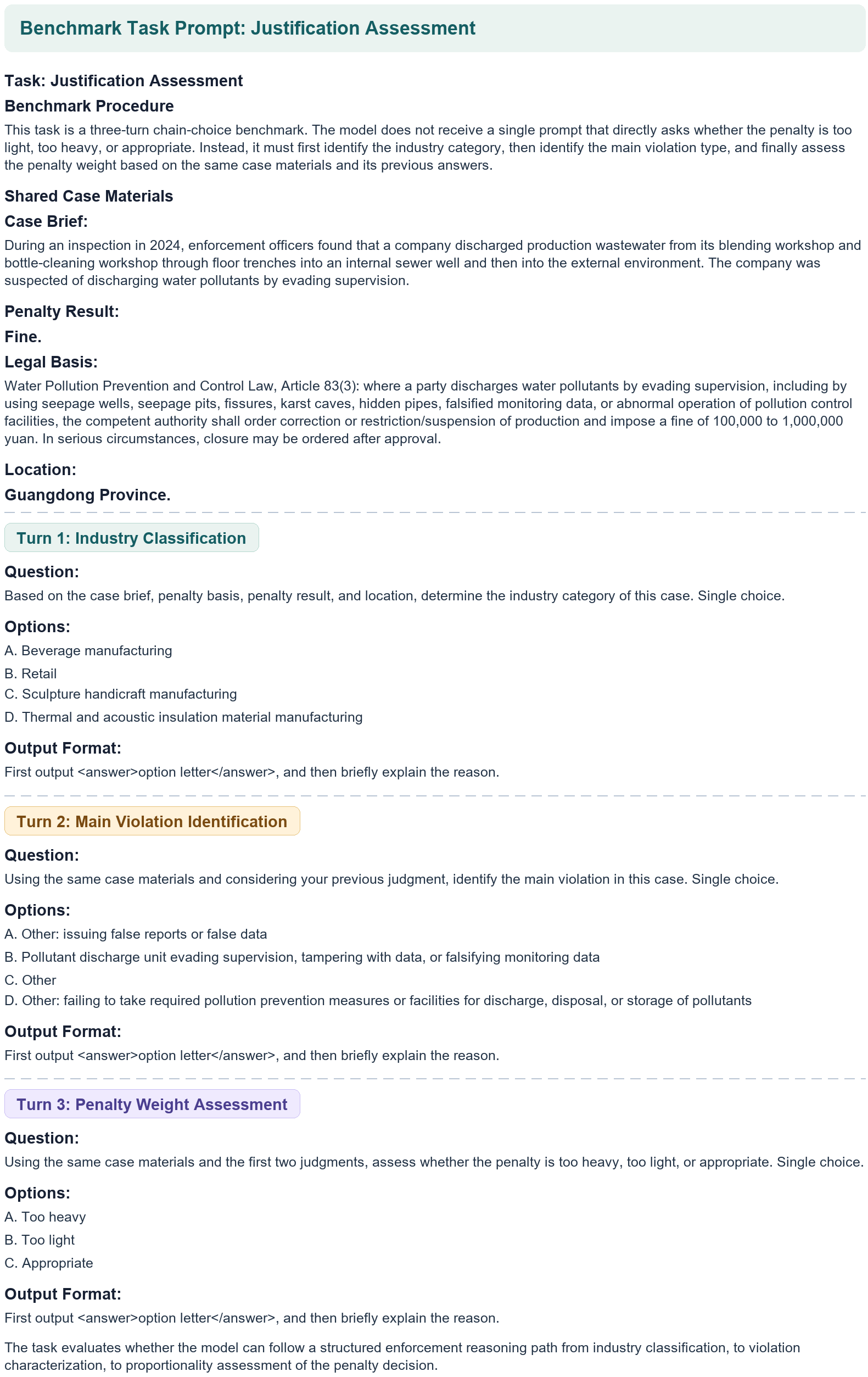}
  \captionof{figure}{Justification assessment benchmark task prompt.}
  \label{fig:appendix-b-28-justification-assessment-benchmark-task-prompt}
\end{minipage}\par\vspace{0.65\baselineskip}

This task does not require the model to make a new penalty decision. Instead, it asks the model to review whether an existing penalty result is consistent with the case facts, legal basis, and discretionary logic, thereby evaluating the model's ability to assess penalty reasoning and proportionality.

\subsubsection*{B.3.4.4 Evaluation}

This task is evaluated by result matching rather than LLM-as-a-Judge. The task has three predefined labels: too lenient, too severe, and appropriate, and the prompt already guides the model through a structured reasoning chain from industry characteristics, violation type, legal basis, and violation circumstances to the final judgment. Therefore, the evaluation directly checks whether the final label and key reasoning steps match the gold answer. LLM-as-a-Judge is unnecessary here because it is mainly used for open-ended tasks with multiple valid expressions; using it for this closed-label task would add evaluation cost and possible scoring variability. If the final label is correct but the reasoning chain is missing or inconsistent, the response is not counted as passing.

\subsubsection*{B.3.5 Penalty Result Reasoning}

\subsubsection*{B.3.5.1 Motivation}

Penalty result reasoning is one of the most complex tasks in the post-enforcement stage. A complete penalty conclusion requires a stepwise legal reasoning process: identifying the relevant industry or pollution domain, determining the main violation type, matching the applicable legal basis, and finally deriving the penalty result. This task evaluates whether models can follow this structured reasoning chain rather than directly mapping case facts to a penalty outcome. It focuses on four consecutive capabilities: industry classification, violation determination, legal-basis matching, and penalty result judgment.

\subsubsection*{B.3.5.2 Data Construction}

This task is constructed from real environmental administrative penalty cases with complete fields, including Case Brief, industry classification, violation category, main violation behavior, legal basis, and penalty result. Each case is converted into a chain-of-reasoning question requiring the model to complete four steps: industry classification, main violation determination, legal-basis matching, and penalty result judgment. For each step, semantically similar distractors are constructed from related cases or legal provisions, so the model must reason through the full chain rather than infer the final answer from keywords. The gold answer includes the correct answer for each step, supporting rationale, and consistency requirements for the complete reasoning chain.

\subsubsection*{B.3.5.3 Prompt}

The evaluated model receives the case facts and must complete the penalty reasoning process step by step.

\clearpage
\par\noindent\begin{minipage}{\linewidth}
  \centering
  \includegraphics[width=\linewidth,height=0.80\textheight,keepaspectratio]{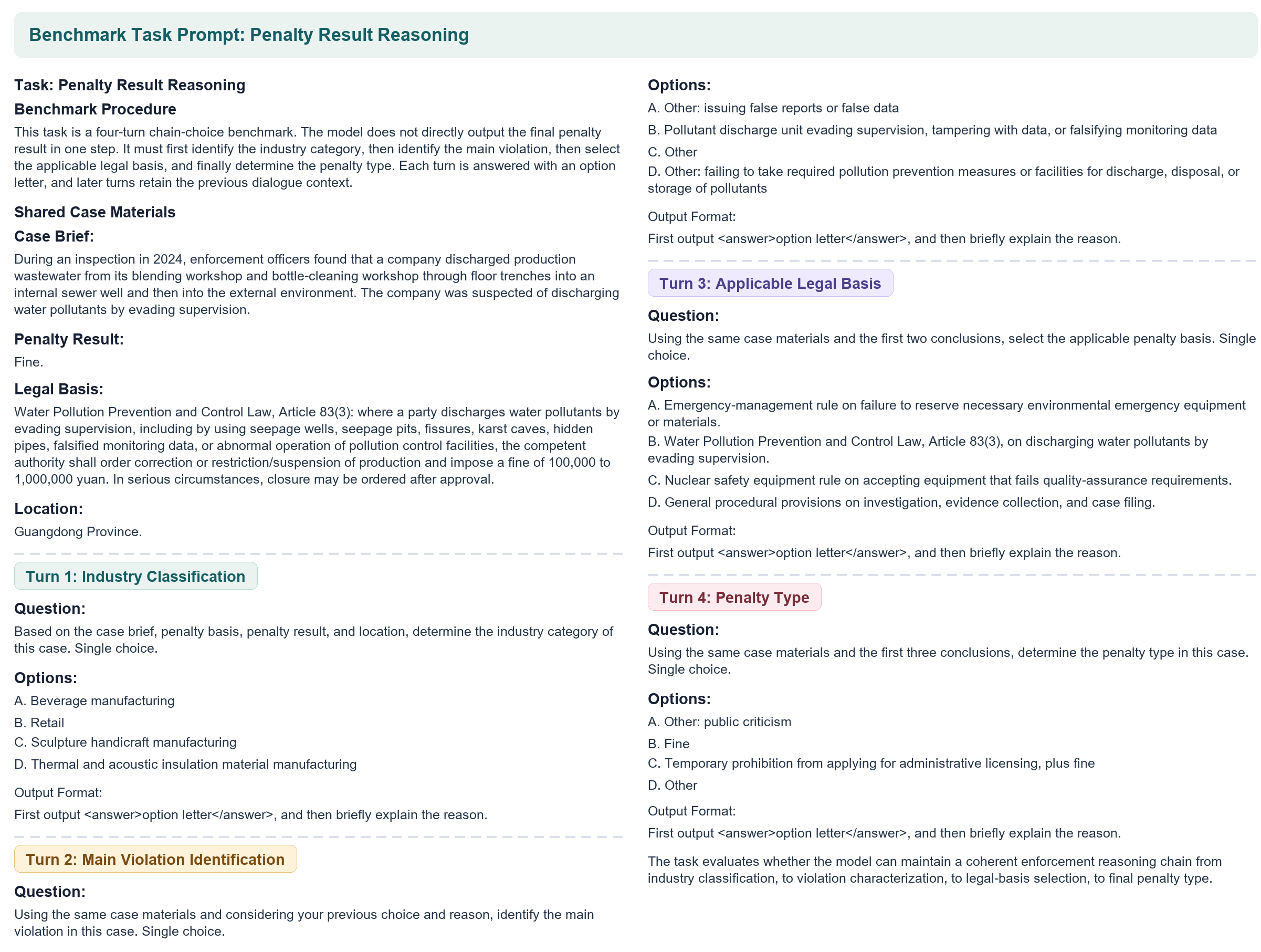}
  \captionof{figure}{Penalty result reasoning benchmark task prompt.}
  \label{fig:appendix-b-29-penalty-result-reasoning-benchmark-task-prompt}
\end{minipage}\par\vspace{0.65\baselineskip}

The task requires the model to maintain logical consistency across the chain of ``industry classification--violation determination--legal basis application--penalty result.'' Errors in earlier steps may lead to incorrect or unreliable penalty conclusions.

\subsubsection*{B.3.5.4 Evaluation}

This task is evaluated by result matching rather than LLM-as-a-Judge. The reason is that the task has a predefined stepwise reasoning structure and each step has a clear gold answer. The prompt already guides the model through the required legal reasoning chain, including industry classification, violation determination, legal-basis matching, and penalty result judgment. Therefore, evaluation can directly check whether each step and the final result match the gold standard. LLM-as-a-Judge is mainly used for open-ended tasks with multiple valid expressions; using it here would add unnecessary evaluation cost and possible scoring variability. A response is counted as passing only when the final penalty result and the key intermediate steps are correct and logically consistent. If the final conclusion is correct but any required intermediate step is missing or incorrect, the response is not counted as passing.

\section{Metrics Calculation}

\subsection*{C.1 Evaluated model groups}

To ensure reproducible calculation of Resource Efficiency and IEI, we first define the evaluated model groups used in this study. Let $\mathcal{M}$ denote the full set of evaluated models. According to model accessibility and deployment characteristics, $\mathcal{M}$ is divided into two groups:

\[
\mathcal{M}=\mathcal{M}_{\mathrm{closed}}\cup\mathcal{M}_{\mathrm{open}},\quad
\mathcal{M}_{\mathrm{closed}}\cap\mathcal{M}_{\mathrm{open}}=\varnothing.
\]

Here, $\mathcal{M}_{\mathrm{closed}}$ denotes closed-source models, and $\mathcal{M}_{\mathrm{open}}$ denotes all evaluated open-source models. These groups are used for model-level comparison and resource-efficiency normalization. Since open-source and closed-source models use different resource proxies, Resource Efficiency and IEI are mainly interpreted within each model group rather than as a universal ranking across all models.

\par\noindent\begin{minipage}{\linewidth}
  \centering
  \small
  \captionof{table}{Evaluated model groups used for IEI calculation.}
  \label{tab:appendix-c-1-evaluated-model-groups-used-for-iei-calculation}
  \begin{tabularx}{\linewidth}{@{}>{\raggedright\arraybackslash}X>{\raggedright\arraybackslash}X>{\raggedright\arraybackslash}X>{\raggedright\arraybackslash}X@{}}
    \toprule
    \textbf{Model group} & \textbf{Set notation} & \textbf{Included models} & \textbf{Resource proxy} \\
    \midrule
    Closed-source models & \(\mathcal{M}_{\mathrm{closed}}\) & GPT-5.5; Claude Opus 4.7; Gemini 3.1 Pro; Grok 4.3; Qwen3.7 Max & Total inference cost during evaluation \\
    Open-source models & \(\mathcal{M}_{\mathrm{open}}\) & DeepSeek V4-Pro; GLM 5.1; MiniMax M3; Qwen3.6-27B; Qwen3.5-27B; Qwen3.6-35B-A3B; Qwen3.5-35B-A3B; Qwen3.5-397B-A17B; Qwen3.5-2B; Qwen3.5-4B; Qwen3.5-9B & Parameter scale \\
    \bottomrule
  \end{tabularx}
\end{minipage}\par\vspace{0.65\baselineskip}

\subsection*{C.2 Task Score Definition}

Environmental enforcement capability is not a single-dimensional ability, but a composite capability across the full enforcement lifecycle. EnvLE-Bench contains 14 representative environmental enforcement tasks covering three stages: Pre-Enforcement, In-Enforcement, and Post-Enforcement. These tasks differ substantially in output format, evaluation criteria, and reasoning requirements. Therefore, raw task-level metrics cannot be directly aggregated.

To enable unified capability assessment across heterogeneous tasks, each task is first converted into a normalized Task Score. Let $\mathcal{T}$ denote the full set of benchmark tasks, and let $t\in\mathcal{T}$ denote a specific task. For model $m$, the normalized task score on task $t$ is denoted as:

\[
\mathrm{TaskScore}_{m,t}\in[0,1].
\]

If task $t$ contains $N_{t}$ test samples, the task-level score is calculated as:

\[
T_{m,t}=\frac{1}{N_{t}}\sum_{i=1}^{N_{t}}s_{m,t,i}.
\]

where $m\in\mathcal{M}$ and $\mathcal{M}$ denotes the full set of evaluated models defined in Section C.1; $s_{m,t,i}$ denotes the sample-level score of model $m$ on the $i$-th sample of task $t$. The calculation of $s_{m,t,i}$ depends on the task type and its corresponding evaluation protocol. For rule-based decision tasks, it is derived from answer correctness or step-level judgment consistency. For open-ended generation tasks, it is obtained through an LLM-as-a-Judge evaluation framework. For structured recognition tasks, it is calculated by aggregating multiple sub-objective scores. Each benchmark task ultimately produces one normalized task score, regardless of the number of samples, evaluation dimensions, or intermediate reasoning steps involved. This design ensures that all tasks contribute comparably to subsequent stage-level aggregation and prevents tasks with larger sample sizes or more complex evaluation procedures from dominating the overall capability score.The task-specific scoring methods are described in Sections C.3.1--C.3.3.

\subsection*{C.3 Task Score Calculation Rules}

According to output format and evaluation objective, the 14 tasks in EnvLE-Bench are grouped into three categories: Rule-based Decision Tasks, Open-ended Generation Tasks, and Structured Recognition Tasks. Because these task types differ in output structure and evaluation logic, we define separate Task Score calculation rules for each category. In all cases, each task is converted into a normalized Task Score for subsequent stage-level aggregation.

\subsubsection*{C.3.1 Rule-based Decision Tasks}

Rule-based decision tasks require models to make enforcement decisions based on case facts, legal provisions, and enforcement procedures, such as violation determination, selection of penalty basis, penalty type classification, and penalty discretion. These tasks usually have clearly defined reference answers, allowing evaluation based on consistency between the model output and the gold answer.

For single-step decision tasks, we use Exact Accuracy as the task score. A sample is counted as correct only when the model output exactly matches the gold answer. The task-level Task Score is computed as the average accuracy across all test samples.

For multi-step rule-based decision tasks, evaluating only the final conclusion may overestimate model capability. A model may reach the correct penalty result while using an incorrect legal basis, misidentifying the violation facts, or applying flawed discretionary reasoning. Therefore, for tasks involving multiple key judgment steps, we jointly evaluate the final decision and the intermediate reasoning process. The task score can be expressed as:

\[
\mathrm{TaskScore}_{m}=\alpha\cdot\mathrm{Score}_{m,\mathrm{final}}+\beta\cdot\mathrm{Score}_{m,\mathrm{process}}.
\]

where $m\in\mathcal{M}$ and $\mathcal{M}$ denotes the full set of evaluated models defined in Section C.1; $\mathrm{Score}_{m,\mathrm{final}}$ denotes the correctness of the model's final decision, and $\mathrm{Score}_{m,\mathrm{process}}$ denotes the correctness of key reasoning steps or intermediate judgments. The parameters $\alpha$ and $\beta$ represent the weights assigned to the final conclusion and the reasoning process, respectively.

For multi-step rule-based decision tasks, we adopt an equal-weight setting, assigning the same importance to the final conclusion and the reasoning process. This design is motivated by the nature of environmental enforcement: a model is expected not only to reach a correct penalty conclusion, but also to produce a traceable enforcement judgment grounded in case facts, evidence, and legal provisions. If a model reaches the correct conclusion through an incorrect reasoning process, its output may still create substantial risks in real enforcement settings. Therefore, both the final decision and the reasoning process are included in the evaluation.

\subsubsection*{C.3.2 Open-ended Generation Tasks}

Open-ended generation tasks include on-site inspection planning, evidence extraction, evidence gap analysis, evidence contradiction detection, and legal-change adaptation. These tasks usually do not have a single fixed textual answer. Different responses may be considered valid as long as they satisfy the core requirements of environmental enforcement. Therefore, traditional text-similarity metrics are insufficient for accurately evaluating model output quality.

To address this issue, we adopt an LLM-as-a-Judge evaluation framework. For each test sample, the judge model receives the case background, task instruction, reference answer, evaluation rubric, and candidate model output. It then determines whether the response satisfies the core requirements according to a unified rubric.

Given the high requirements for reliability and procedural conformity in environmental enforcement tasks, we use Strict Pass Rate as the Task Score for open-ended generation tasks. A sample is counted as passed only when the model output satisfies all core evaluation requirements. The task-level Task Score is calculated as the pass rate across all samples.

This design reduces misjudgment caused by surface-level textual similarity and focuses on whether the response substantively meets the requirements of the enforcement task. For example, in on-site inspection planning, the model must not only propose inspection steps, but also cover key pollution sources, evidence collection methods, on-site safety requirements, and procedural enforcement norms. If a response omits any key component, it will not be judged as passing, even if it is fluent in language.

\par\noindent\begin{minipage}{\linewidth}
  \centering
  \includegraphics[width=\linewidth,height=0.72\textheight,keepaspectratio]{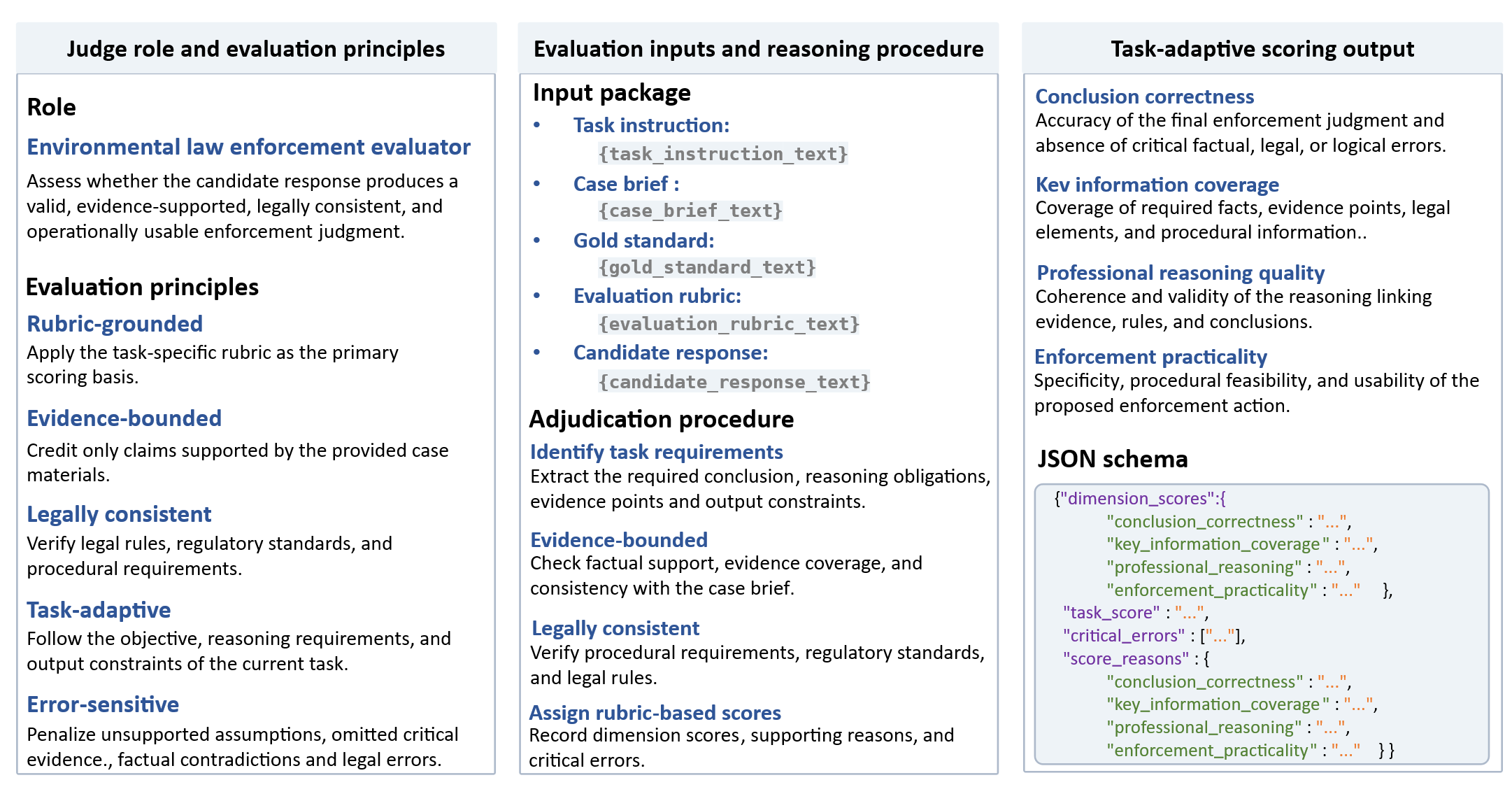}
  \captionof{figure}{Unified LLM-as-a-Judge prompt template for evaluating
open-ended environmental law enforcement tasks. The template
specifies the judge role, evaluation inputs, adjudication procedure,
task-adaptive scoring dimensions, and structured output schema}
  \label{fig:main-2-unified-llm-as-a-judge-prompt-template-for-environmen}
\end{minipage}\par\vspace{0.65\baselineskip}

\subsubsection*{C.3.3 Structured Recognition Tasks}

Structured recognition tasks include complaint validity identification, environmental risk recognition, monitoring anomaly detection, and abnormality detection in inquiry transcripts. These tasks usually involve multiple parallel judgment objectives, so a single accuracy score cannot fully capture model performance.

For example, in environmental risk recognition, the model must determine whether a risk exists and identify its type or severity level. In monitoring anomaly detection, the model must first detect whether an anomaly exists and then classify the anomaly type. In inquiry transcript abnormality detection, the model must not only identify the presence of an abnormality, but also locate it or explain its cause.Therefore, for structured recognition tasks, we score each sub-objective separately and aggregate them into a task-level score:

\[
\mathrm{TaskScore}_{m}=\sum_{k=1}^{K}\gamma_{k}\cdot\mathrm{Score}_{m,k}.
\]

where $m\in\mathcal{M}$ and $\mathcal{M}$ denotes the full set of evaluated models defined in Section C.1; $K$ denotes the number of evaluation sub-objectives, $\mathrm{Score}_{m,k}$ denotes the score of the $k$-th sub-objective, and $\gamma_{k}$ denotes its weight. Unless otherwise specified, we use equal weights for all sub-objectives, because they jointly form a complete enforcement judgment chain. For instance, merely detecting the existence of a risk without identifying its type, or detecting an anomaly without locating its source, is insufficient to support subsequent enforcement decisions.

For a small number of tasks with clear primary--secondary relationships among sub-objectives, the weights are adjusted according to the task-specific rubric and explicitly reported in the corresponding task description. All weights are predefined before evaluation and are not adjusted based on model performance, ensuring consistency and reproducibility.

\subsection*{C.4 Absolute Environmental Enforcement Score (AES)}

AES measures a model's overall task performance across the full lifecycle of environmental enforcement. Let $\mathcal{G}$ denote the set of enforcement stages, including Pre-Enforcement, In-Enforcement, and Post-Enforcement. For each stage $g\in\mathcal{G}$, let $\mathcal{T}_{g}$ denote the set of tasks belonging to that stage, and let $\mathcal{M}$ denote the full set of evaluated models defined in Section C.1. The stage-level score of model $m$ is calculated as the macro-average of all task scores within that stage:

\[
S_{m,g}=\frac{1}{|\mathcal{T}_{g}|}\sum_{t\in\mathcal{T}_{g}}T_{m,t}.
\]

This macro-averaging strategy ensures that all tasks within the same enforcement stage contribute equally, regardless of sample size, evaluation dimensions, or task complexity.The normalized AES is then obtained by aggregating the stage-level scores:

\[
\mathrm{AES}^{\mathrm{norm}}_{m}=\sum_{g\in\mathcal{G}}\alpha_{g}S_{m,g}.
\]

where $\alpha_{g}$ denotes the weight assigned to enforcement stage $g$, and the three enforcement stages are assigned equal weights, so that pre-enforcement, in-enforcement, and post-enforcement contribute equally to the final capability score. The reported AES in the main text is mapped from $\mathrm{AES}^{\mathrm{norm}}_{m}$ to a 0--100 scale for readability. A higher AES indicates that the model demonstrates stronger and more stable capability across the full lifecycle of environmental enforcement.

\subsection*{C.5 Quality Factor Calculation}

AES measures whether a model completes enforcement tasks correctly, but it does not fully capture whether the response is reliable, explainable, and professionally usable. In environmental enforcement, a valid response should not only reach the correct conclusion, but also provide factual support, evidentiary reasoning, and legally appropriate expression. Therefore, we introduce the Quality Factor to evaluate response quality beyond task correctness.

For each evaluated response $i$ from model $m$, an independent judge model scores the response along three dimensions: Reliability, Explainability, and Professionalism. Reliability evaluates whether the response is consistent with case facts, evidence, legal provisions, and environmental enforcement practice. Explainability evaluates whether the response provides a clear and complete reasoning process, including evidentiary relationships, legal applicability, and key judgment bases. Professionalism evaluates whether the response uses appropriate terminology, follows enforcement norms, and is suitable for professional enforcement writing.

Each dimension is scored on a 0--4 scale and then normalized to the range $[0,1]$. Let $q^{R}_{m,i}$, $q^{E}_{m,i}$, and $q^{P}_{m,i}$ denote the normalized scores for Reliability, Explainability, and Professionalism, respectively, and let $\mathcal{M}$ denote the full set of evaluated models defined in Section C.1. The item-level Quality Factor is calculated as:

\[
Q_{m,i}=\beta_{R}q^{R}_{m,i}+\beta_{E}q^{E}_{m,i}+\beta_{P}q^{P}_{m,i}.
\]

where $\beta_{R}$, $\beta_{E}$, and $\beta_{P}$ denote the weights assigned to the three quality dimensions, and the three dimensions are assigned equal weights. The model-level Quality Factor is obtained by averaging item-level quality scores across all evaluated responses:

\[
Q_{m}=\frac{1}{|\mathcal{I}_{m}|}\sum_{i\in\mathcal{I}_{m}}Q_{m,i}.
\]

where $\mathcal{I}_{m}$ denotes the set of evaluated responses generated by model $m$. By construction, $Q_{m}\in[0,1]$. A higher Quality Factor indicates that the model output is not only task-compliant, but also more reliable, interpretable, and professionally suitable for environmental enforcement. The following figure presents the simplified prompt template used for Quality Factor evaluation, in which the judge model returns structured scores and explanations in JSON format.

\par\noindent\begin{minipage}{\linewidth}
  \centering
  \includegraphics[width=\linewidth,height=0.72\textheight,keepaspectratio]{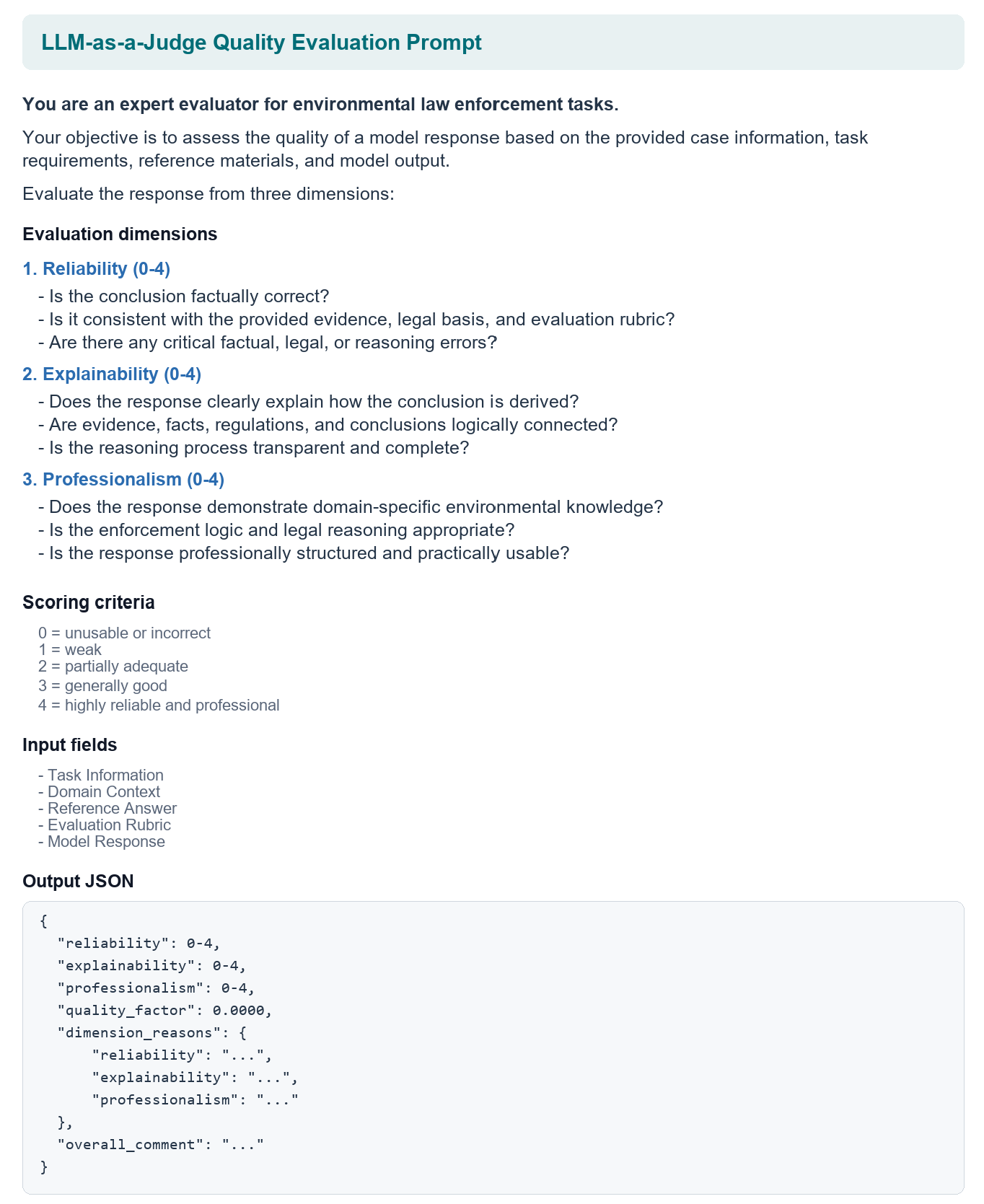}
  \captionof{figure}{Prompt template used for Quality Factor evaluation}
  \label{fig:appendix-c-1-prompt-template-used-for-quality-factor-evaluat}
\end{minipage}\par\vspace{0.65\baselineskip}

\subsection*{C.6 Resource Efficiency for Open-Source Models}

Although AES and Quality Factor measure task completion capability and response reliability, respectively, they do not capture the resource cost required to achieve such capability. In practical deployment, especially in local environmental agencies, mobile enforcement terminals, and edge-computing scenarios, model size directly affects deployment difficulty, hardware requirements, and operating cost.

For open-source models, parameter size is usually publicly available and is therefore used as a proxy for resource consumption. However, the parameter scales of current large language models vary substantially, ranging from billions to hundreds of billions of parameters. Directly using raw parameter size would over-amplify the difference between very large and medium-sized models, causing the resource term to exert excessive influence on the final evaluation.

To address this issue, we first apply a logarithmic transformation to parameter size to reduce the effect of extreme scale differences. We then standardize the transformed values across all open-source models to obtain a Parameter Burden indicator. A larger Parameter Burden indicates higher computational and deployment resource requirements. Based on this indicator, we define Open Resource Efficiency as follows:

\[
\mathrm{OpenResourceEfficiency}_{m}=1-\lambda_{p}\cdot\mathrm{ParameterBurden}_{m}.
\]

where $m\in\mathcal{M}_{\mathrm{open}}$, $\mathcal{M}_{\mathrm{open}}$ denotes the set of evaluated open-source models defined in Section C.1, and $\lambda_{p}$ is the parameter penalty coefficient, which controls the influence of resource cost on the final evaluation. Resource Efficiency is not intended to dominate model ranking. The primary focus of environmental enforcement evaluation remains model capability, while resource cost is used only to reflect the additional burden required to achieve that capability. Therefore, Resource Efficiency is designed as an adjustment factor rather than an independent evaluation objective.

Following this principle, we set $\lambda_{p}=0.2$. This setting strictly limits the impact of resource efficiency on the final score. Even when two models differ substantially in parameter size, their resource efficiency gap remains within a bounded range. This design captures deployment-cost differences while preventing smaller models from outperforming substantially stronger models solely because of their parameter advantage. A higher Resource Efficiency indicates that a model can achieve comparable environmental enforcement capability with fewer parameters, whereas a lower Resource Efficiency indicates that capability improvement requires greater resource investment.

\subsection*{C.7 Resource Efficiency for Closed-Source Models}

For open-source models, parameter size is usually publicly available and can be used to approximate resource requirements. For closed-source models such as GPT, Claude, and Gemini, however, actual parameter sizes are typically undisclosed, and disclosure standards vary across providers. Therefore, parameter size cannot serve as a unified and verifiable resource-cost indicator for closed-source models.

We instead use inference cost as a proxy for the resource consumption of closed-source models. Compared with parameter size, inference cost more directly reflects the expenditure users face during real deployment and long-term use. Specifically, we calculate the total billable output characters generated by each model across all EnvLE-Bench tasks, and estimate the total inference cost using the corresponding model prices. Because output lengths vary across models, unit price alone cannot accurately represent total resource consumption. Therefore, we use the actual total output cost incurred during evaluation as the basis for resource assessment.

Pricing information for closed-source models was obtained from the OpenRouter Model Registry. All prices were retrieved through the official OpenRouter API on June 17, 2026, and the publicly available completion prices at that time were used as the basis for output-cost estimation. To ensure consistency, all models use prices collected at the same time point, and subsequent price changes are not considered.

Because OpenRouter reports token-level prices whereas this study records model outputs in characters, token prices are converted into character-level costs. Following common empirical estimates of the relationship between tokens and text length in public documentation from major model providers, we assume that one output token corresponds to 1.5 output characters on average. Under this assumption, token-level completion prices can be converted into character-level costs, enabling consistent comparison across closed-source models. Table C.2 reports the output prices used for closed-source model cost estimation.

\par\noindent\begin{minipage}{\linewidth}
  \centering
  \small
  \captionof{table}{Output pricing used for closed-source model cost estimation.}
  \label{tab:appendix-c-2-output-pricing-used-for-closed-source-model-cos}
  \begin{tabularx}{\linewidth}{@{}>{\raggedright\arraybackslash}X>{\raggedright\arraybackslash}X>{\raggedright\arraybackslash}X>{\raggedright\arraybackslash}X@{}}
    \toprule
    \textbf{Model} & \textbf{Provider} & \textbf{Output Price (USD / 1M Tokens)} & \textbf{Price per 1k Characters (USD)} \\
    \midrule
    GPT-5.5 & OpenAI & 30.00 & 0.0200 \\
    Claude Opus 4.7 & Anthropic & 25.00 & 0.0167 \\
    Gemini 3.1 Pro & Google & 12.00 & 0.0080 \\
    Grok 4.3 & xAI & 2.50 & 0.00167 \\
    Qwen3.7 Max & Alibaba Cloud & 3.75 & 0.00250 \\
    \bottomrule
  \end{tabularx}
\end{minipage}\par\vspace{0.65\baselineskip}

After obtaining the total inference cost of each model, we first apply a logarithmic transformation to reduce the influence of extreme cost differences, and then standardize the transformed values to construct a Cost Burden indicator. A higher Cost Burden indicates that the model requires greater inference cost to complete the same evaluation tasks. Based on this indicator, Closed Resource Efficiency is defined as:

\[
\mathrm{ClosedResourceEfficiency}_{m}=\frac{1}{1+\lambda_{c}\cdot\mathrm{CostBurden}_{m}}.
\]

where $m\in\mathcal{M}_{\mathrm{closed}}$, $\mathrm{CostBurden}_{m}$ denotes the standardized cost burden, and $\lambda_{c}$ is the cost penalty coefficient. Similar to the parameter-based adjustment for open-source models, inference cost is used only as an adjustment factor rather than as the primary determinant of model capability. The goal of Resource Efficiency is to reflect the match between capability and cost, rather than simply rewarding low-cost models.

Considering that inference cost directly affects large-scale deployment and long-term use of closed-source models, we set $\lambda_{c}=1.0$. This setting allows cost differences to be reflected in the final evaluation while preventing resource cost from fully dominating model ranking. The resulting Closed Resource Efficiency is approximately bounded between 0.5 and 1.0. A higher value indicates that a model achieves comparable environmental enforcement capability at lower inference cost, whereas a lower value indicates that capability improvement requires higher cost investment.

\subsection*{C.8 Intelligent Enforcement Index (IEI)}

AES measures a model's absolute task performance in environmental enforcement, while Quality Factor evaluates the reliability, explainability, and professionalism of its responses. Capability Base combines these two aspects into a unified capability measure, but it does not account for the resource cost required to achieve such capability. In real enforcement scenarios, models with similar task performance may differ substantially in deployment requirements and long-term operating costs. Therefore, we further define the Intelligent Enforcement Index (IEI) as:

\[
\begin{gathered}
\mathrm{CB}_{m}=w_{a}\cdot\tilde{\mathrm{AES}}_{m}+w_{q}\cdot Q_{m},\\
\mathrm{IEI}_{m}=\mathrm{CB}_{m}\cdot\mathrm{RE}_{m}.
\end{gathered}
\]

where $\mathrm{AES}$ measures task completion capability, and $Q$ measures response quality in terms of reliability, explainability, and professionalism. We set $w_{a}=w_{q}=0.5$, assigning equal importance to task correctness and response quality. This setting reflects the practical requirements of environmental enforcement: correct conclusions are essential for decision support, while evidentiary completeness, legal applicability, and process interpretability are also critical for enforcement practice. $\mathrm{RE}$ denotes the efficiency with which a model achieves its capability.

For open-source models, it is defined as:

\[
\mathrm{RE}_{m}=\mathrm{OpenResourceEfficiency}_{m}.
\]

For closed-source models, it is defined as:

\[
\mathrm{RE}_{m}=\mathrm{ClosedResourceEfficiency}_{m}.
\]

IEI emphasizes the balance between capability and resource consumption. A higher IEI indicates that a model achieves stronger environmental enforcement capability with relatively lower resource investment, whereas a lower IEI suggests that the capability gain may not justify the additional resource cost. Therefore, IEI measures the practical application value of a model in environmental enforcement, rather than only its absolute capability. By jointly considering task performance, response quality, and resource efficiency, IEI provides a more comprehensive, objective, and deployment-oriented evaluation framework for large language models in environmental enforcement.

\section{Complete Benchmark Results}
This appendix reports the complete benchmark scores of all evaluated models, organized by the three enforcement phases (pre-enforcement, in-enforcement, and post-enforcement) and by 12 pollutant secondary categories. Sections D.1--D.3 present task-level results for each phase, and Section D.4 provides detailed category-level scores.
\subsection*{D.1 Pre-Enforcement Tasks}

\subsubsection*{D.1.1 Complaint Verification}

\par\noindent\begin{minipage}{\linewidth}
  \centering
  \small
  \captionof{table}{Complaint Verification (Task Score).}
  \label{tab:appendix-d-1-complaint-verification-task-score}
  \begin{tabular*}{\linewidth}{@{\extracolsep{\fill}}lr@{}}
    \toprule
    \textbf{Model} & \textbf{Task Score} \\
    \midrule
    DeepSeek V4-Pro-thinking (default) & 83.89 \\
    Gemini 3.1 Pro Preview & 83.89 \\
    GPT-5.5-thinking & 78.33 \\
    GPT-5.5-nothinking & 75.00 \\
    Grok 4.3-nothinking & 69.44 \\
    Grok 4.3-thinking & 66.67 \\
    Qwen3.5 2B-thinking & 65.56 \\
    GPT-5.5 (default) & 63.89 \\
    Claude Opus 4.7-nothinking & 57.78 \\
    GLM 5.1-nothinking & 56.67 \\
    Qwen3.7 Max & 56.11 \\
    MiniMax M3-thinking (default) & 55.56 \\
    Qwen3.5 4B-thinking & 54.44 \\
    Qwen3.6 27B & 53.33 \\
    Qwen3.6 35B-A3B & 52.78 \\
    Grok 4.3 (default) & 52.78 \\
    DeepSeek V4-Pro-nothinking & 52.22 \\
    Qwen3.5 397B-A17B & 52.22 \\
    Claude Opus 4.7-thinking (default) & 51.11 \\
    GLM 5.1-thinking (default) & 50.56 \\
    MiniMax M3-nothinking & 50.56 \\
    Qwen3.5 35B-A3B & 48.89 \\
    Qwen3.5 9B-thinking & 48.89 \\
    Qwen3.5 27B & 48.33 \\
    Qwen3.5 9B-nothinking & 47.78 \\
    Qwen3.5 4B-nothinking & 47.22 \\
    Qwen3.5 2B-nothinking & 47.22 \\
    \bottomrule
  \end{tabular*}
\end{minipage}\par\vspace{0.65\baselineskip}

\subsubsection*{D.1.2 Inspection Planning}

\par\noindent\begin{minipage}{\linewidth}
  \centering
  \small
  \captionof{table}{Inspection Planning (Task Score).}
  \label{tab:appendix-d-2-inspection-planning-task-score}
  \begin{tabular*}{\linewidth}{@{\extracolsep{\fill}}lr@{}}
    \toprule
    \textbf{Model} & \textbf{Task Score} \\
    \midrule
    GPT-5.5-nothinking & 88.64 \\
    GPT-5.5-thinking & 88.07 \\
    GPT-5.5 (default) & 86.93 \\
    MiniMax M3-thinking (default) & 51.14 \\
    Qwen3.7 Max & 49.43 \\
    Qwen3.6 27B & 47.16 \\
    Qwen3.6 35B-A3B & 46.59 \\
    Claude Opus 4.7-nothinking & 43.18 \\
    Claude Opus 4.7-thinking (default) & 32.39 \\
    MiniMax M3-nothinking & 31.25 \\
    DeepSeek V4-Pro-thinking (default) & 27.84 \\
    Qwen3.5 397B-A17B & 26.70 \\
    DeepSeek V4-Pro-nothinking & 25.57 \\
    GLM 5.1-thinking (default) & 21.59 \\
    Qwen3.5 27B & 19.89 \\
    GLM 5.1-nothinking & 18.75 \\
    Gemini 3.1 Pro Preview & 17.05 \\
    Qwen3.5 35B-A3B & 15.34 \\
    Grok 4.3-nothinking & 5.68 \\
    Qwen3.5 9B-thinking & 2.84 \\
    Grok 4.3-thinking & 2.27 \\
    Grok 4.3 (default) & 1.70 \\
    Qwen3.5 9B-nothinking & 1.14 \\
    Qwen3.5 4B-thinking & 1.14 \\
    Qwen3.5 4B-nothinking & 0.00 \\
    Qwen3.5 2B-nothinking & 0.00 \\
    Qwen3.5 2B-thinking & 0.00 \\
    \bottomrule
  \end{tabular*}
\end{minipage}\par\vspace{0.65\baselineskip}

\subsubsection*{D.1.3 Public Sentiment Analysis}

\par\noindent\begin{minipage}{\linewidth}
  \centering
  \small
  \captionof{table}{Public Sentiment Analysis (Task Score).}
  \label{tab:appendix-d-3-public-sentiment-analysis-task-score}
  \begin{tabular*}{\linewidth}{@{\extracolsep{\fill}}lr@{}}
    \toprule
    \textbf{Model} & \textbf{Task Score} \\
    \midrule
    MiniMax M3-thinking (default) & 61.85 \\
    GPT-5.5-nothinking & 57.95 \\
    MiniMax M3-nothinking & 56.26 \\
    GPT-5.5 (default) & 55.85 \\
    Qwen3.5 35B-A3B & 54.97 \\
    Qwen3.6 35B-A3B & 54.41 \\
    GLM 5.1-nothinking & 53.59 \\
    GPT-5.5-thinking & 53.08 \\
    Qwen3.5 397B-A17B & 52.51 \\
    Qwen3.5 27B & 52.41 \\
    Qwen3.6 27B & 52.15 \\
    Gemini 3.1 Pro Preview & 51.38 \\
    DeepSeek V4-Pro-nothinking & 51.33 \\
    Qwen3.7 Max & 51.23 \\
    GLM 5.1-thinking (default) & 51.08 \\
    Qwen3.5 9B-nothinking & 50.31 \\
    Claude Opus 4.7-thinking (default) & 48.31 \\
    Qwen3.5 4B-nothinking & 48.21 \\
    Grok 4.3-nothinking & 48.05 \\
    Qwen3.5 2B-nothinking & 48.05 \\
    Grok 4.3 (default) & 47.54 \\
    Claude Opus 4.7-nothinking & 47.18 \\
    Qwen3.5 9B-thinking & 46.31 \\
    DeepSeek V4-Pro-thinking (default) & 45.79 \\
    Grok 4.3-thinking & 45.49 \\
    Qwen3.5 4B-thinking & 42.31 \\
    Qwen3.5 2B-thinking & 13.33 \\
    \bottomrule
  \end{tabular*}
\end{minipage}\par\vspace{0.65\baselineskip}

\subsubsection*{D.1.4 Anomaly Detection}

\par\noindent\begin{minipage}{\linewidth}
  \centering
  \small
  \captionof{table}{Anomaly Detection (Task Score).}
  \label{tab:appendix-d-4-anomaly-detection-task-score}
  \begin{tabular*}{\linewidth}{@{\extracolsep{\fill}}lr@{}}
    \toprule
    \textbf{Model} & \textbf{Task Score} \\
    \midrule
    GPT-5.5 (default) & 70.42 \\
    GPT-5.5-thinking & 69.90 \\
    DeepSeek V4-Pro-nothinking & 69.90 \\
    Grok 4.3-nothinking & 69.37 \\
    Grok 4.3-thinking & 68.85 \\
    Grok 4.3 (default) & 68.85 \\
    Qwen3.6 35B-A3B & 68.59 \\
    GPT-5.5-nothinking & 68.06 \\
    Qwen3.6 27B & 67.80 \\
    Qwen3.5 397B-A17B & 67.80 \\
    MiniMax M3-nothinking & 67.80 \\
    GLM 5.1-nothinking & 67.54 \\
    Claude Opus 4.7-thinking (default) & 66.23 \\
    Qwen3.7 Max & 65.45 \\
    Claude Opus 4.7-nothinking & 64.92 \\
    GLM 5.1-thinking (default) & 64.66 \\
    DeepSeek V4-Pro-thinking (default) & 63.09 \\
    Gemini 3.1 Pro Preview & 54.19 \\
    MiniMax M3-thinking (default) & 44.50 \\
    Qwen3.5 9B-nothinking & 33.51 \\
    Qwen3.5 35B-A3B & 33.25 \\
    Qwen3.5 27B & 32.98 \\
    Qwen3.5 4B-nothinking & 31.15 \\
    Qwen3.5 9B-thinking & 25.13 \\
    Qwen3.5 2B-thinking & 23.56 \\
    Qwen3.5 4B-thinking & 23.30 \\
    Qwen3.5 2B-nothinking & 22.77 \\
    \bottomrule
  \end{tabular*}
\end{minipage}\par\vspace{0.65\baselineskip}

\subsection*{D.2 In-Enforcement Tasks}

\subsubsection*{D.2.1 Single-turn Query Analysis}

\par\noindent\begin{minipage}{\linewidth}
  \centering
  \small
  \captionof{table}{Single-turn Query Analysis (Task Score).}
  \label{tab:appendix-d-5-single-turn-query-analysis-task-score}
  \begin{tabular*}{\linewidth}{@{\extracolsep{\fill}}lr@{}}
    \toprule
    \textbf{Model} & \textbf{Task Score} \\
    \midrule
    Grok 4.3 (default) & 65.28 \\
    Qwen3.6 35B-A3B & 65.00 \\
    GPT-5.5-thinking & 64.44 \\
    Grok 4.3-nothinking & 64.44 \\
    GPT-5.5 (default) & 63.89 \\
    DeepSeek V4-Pro-nothinking & 63.61 \\
    Grok 4.3-thinking & 63.61 \\
    Qwen3.6 27B & 62.78 \\
    GLM 5.1-nothinking & 62.78 \\
    GLM 5.1-thinking (default) & 62.78 \\
    GPT-5.5-nothinking & 62.50 \\
    Claude Opus 4.7-nothinking & 62.50 \\
    Qwen3.5 35B-A3B & 62.50 \\
    DeepSeek V4-Pro-thinking (default) & 61.94 \\
    Qwen3.7 Max & 61.67 \\
    Claude Opus 4.7-thinking (default) & 61.39 \\
    Gemini 3.1 Pro Preview & 60.56 \\
    Qwen3.5 27B & 59.72 \\
    MiniMax M3-thinking (default) & 58.61 \\
    Qwen3.5 397B-A17B & 58.33 \\
    Qwen3.5 9B-thinking & 58.33 \\
    MiniMax M3-nothinking & 58.06 \\
    Qwen3.5 4B-thinking & 56.39 \\
    Qwen3.5 4B-nothinking & 55.83 \\
    Qwen3.5 9B-nothinking & 52.50 \\
    Qwen3.5 2B-nothinking & 50.83 \\
    Qwen3.5 2B-thinking & 47.50 \\
    \bottomrule
  \end{tabular*}
\end{minipage}\par\vspace{0.65\baselineskip}

\subsubsection*{D.2.2 Full Inquiry Analysis}

\par\noindent\begin{minipage}{\linewidth}
  \centering
  \small
  \captionof{table}{Full Inquiry Analysis (Task Score).}
  \label{tab:appendix-d-6-full-inquiry-analysis-task-score}
  \begin{tabular*}{\linewidth}{@{\extracolsep{\fill}}lr@{}}
    \toprule
    \textbf{Model} & \textbf{Task Score} \\
    \midrule
    DeepSeek V4-Pro-thinking (default) & 61.39 \\
    Qwen3.6 35B-A3B & 59.72 \\
    Claude Opus 4.7-nothinking & 59.44 \\
    Claude Opus 4.7-thinking (default) & 59.17 \\
    Grok 4.3-thinking & 58.89 \\
    Gemini 3.1 Pro Preview & 58.61 \\
    GLM 5.1-thinking (default) & 58.33 \\
    Grok 4.3-nothinking & 58.33 \\
    Grok 4.3 (default) & 57.78 \\
    Qwen3.7 Max & 57.50 \\
    Qwen3.5 35B-A3B & 56.67 \\
    Qwen3.5 27B & 56.39 \\
    GPT-5.5 (default) & 56.11 \\
    Qwen3.5 397B-A17B & 56.11 \\
    GPT-5.5-thinking & 55.83 \\
    GPT-5.5-nothinking & 54.72 \\
    Qwen3.6 27B & 53.61 \\
    MiniMax M3-thinking (default) & 53.61 \\
    DeepSeek V4-Pro-nothinking & 53.33 \\
    Qwen3.5 9B-nothinking & 52.50 \\
    MiniMax M3-nothinking & 52.22 \\
    GLM 5.1-nothinking & 51.39 \\
    Qwen3.5 4B-nothinking & 51.39 \\
    Qwen3.5 4B-thinking & 51.11 \\
    Qwen3.5 9B-thinking & 50.83 \\
    Qwen3.5 2B-nothinking & 50.28 \\
    Qwen3.5 2B-thinking & 50.28 \\
    \bottomrule
  \end{tabular*}
\end{minipage}\par\vspace{0.65\baselineskip}

\subsubsection*{D.2.3 Multi-Evidence Extraction}

\par\noindent\begin{minipage}{\linewidth}
  \centering
  \small
  \captionof{table}{Multi-Evidence Extraction (Task Score).}
  \label{tab:appendix-d-7-multi-evidence-extraction-task-score}
  \begin{tabular*}{\linewidth}{@{\extracolsep{\fill}}lr@{}}
    \toprule
    \textbf{Model} & \textbf{Task Score} \\
    \midrule
    GPT-5.5 (default) & 97.25 \\
    GPT-5.5-nothinking & 94.51 \\
    GPT-5.5-thinking & 90.66 \\
    DeepSeek V4-Pro-nothinking & 85.16 \\
    Qwen3.7 Max & 84.07 \\
    Claude Opus 4.7-nothinking & 83.52 \\
    GLM 5.1-nothinking & 81.87 \\
    Claude Opus 4.7-thinking (default) & 79.67 \\
    Grok 4.3-nothinking & 74.73 \\
    Qwen3.5 27B & 72.53 \\
    DeepSeek V4-Pro-thinking (default) & 71.98 \\
    Grok 4.3 (default) & 71.98 \\
    Qwen3.6 27B & 69.23 \\
    GLM 5.1-thinking (default) & 67.03 \\
    Qwen3.6 35B-A3B & 66.48 \\
    Grok 4.3-thinking & 65.93 \\
    Qwen3.5 9B-nothinking & 64.84 \\
    Qwen3.5 397B-A17B & 54.95 \\
    Gemini 3.1 Pro Preview & 52.75 \\
    Qwen3.5 4B-nothinking & 48.90 \\
    Qwen3.5 35B-A3B & 47.25 \\
    MiniMax M3-thinking (default) & 32.42 \\
    Qwen3.5 4B-thinking & 30.77 \\
    MiniMax M3-nothinking & 26.37 \\
    Qwen3.5 2B-nothinking & 17.58 \\
    Qwen3.5 9B-thinking & 16.48 \\
    Qwen3.5 2B-thinking & 16.48 \\
    \bottomrule
  \end{tabular*}
\end{minipage}\par\vspace{0.65\baselineskip}

\subsubsection*{D.2.4 Contradiction Monitoring}

\par\noindent\begin{minipage}{\linewidth}
  \centering
  \small
  \captionof{table}{Contradiction Monitoring (Task Score).}
  \label{tab:appendix-d-8-contradiction-monitoring-task-score}
  \begin{tabular*}{\linewidth}{@{\extracolsep{\fill}}lr@{}}
    \toprule
    \textbf{Model} & \textbf{Task Score} \\
    \midrule
    Claude Opus 4.7-thinking (default) & 32.42 \\
    Claude Opus 4.7-nothinking & 31.32 \\
    MiniMax M3-thinking (default) & 30.77 \\
    Qwen3.5 27B & 29.67 \\
    Qwen3.7 Max & 29.12 \\
    GLM 5.1-thinking (default) & 29.12 \\
    Grok 4.3-nothinking & 29.12 \\
    GPT-5.5-nothinking & 28.57 \\
    GLM 5.1-nothinking & 28.57 \\
    Grok 4.3-thinking & 28.57 \\
    Gemini 3.1 Pro Preview & 28.02 \\
    MiniMax M3-nothinking & 28.02 \\
    Qwen3.5 35B-A3B & 28.02 \\
    Qwen3.6 27B & 27.47 \\
    DeepSeek V4-Pro-nothinking & 26.92 \\
    GPT-5.5 (default) & 26.37 \\
    GPT-5.5-thinking & 25.27 \\
    Qwen3.5 397B-A17B & 25.27 \\
    Qwen3.6 35B-A3B & 24.18 \\
    DeepSeek V4-Pro-thinking (default) & 22.53 \\
    Grok 4.3 (default) & 22.53 \\
    Qwen3.5 9B-nothinking & 20.88 \\
    Qwen3.5 9B-thinking & 18.68 \\
    Qwen3.5 4B-thinking & 12.09 \\
    Qwen3.5 4B-nothinking & 7.69 \\
    Qwen3.5 2B-thinking & 4.95 \\
    Qwen3.5 2B-nothinking & 2.20 \\
    \bottomrule
  \end{tabular*}
\end{minipage}\par\vspace{0.65\baselineskip}

\subsubsection*{D.2.5 Multi-Evidence Integration}

\par\noindent\begin{minipage}{\linewidth}
  \centering
  \small
  \captionof{table}{Multi-Evidence Integration (Task Score).}
  \label{tab:appendix-d-9-multi-evidence-integration-task-score}
  \begin{tabular*}{\linewidth}{@{\extracolsep{\fill}}lr@{}}
    \toprule
    \textbf{Model} & \textbf{Task Score} \\
    \midrule
    MiniMax M3-thinking (default) & 65.38 \\
    Gemini 3.1 Pro Preview & 52.75 \\
    MiniMax M3-nothinking & 51.10 \\
    Qwen3.6 35B-A3B & 49.45 \\
    GLM 5.1-nothinking & 47.80 \\
    Qwen3.5 27B & 45.60 \\
    Qwen3.5 35B-A3B & 42.86 \\
    DeepSeek V4-Pro-thinking (default) & 41.76 \\
    Qwen3.6 27B & 40.66 \\
    Qwen3.5 4B-thinking & 37.91 \\
    DeepSeek V4-Pro-nothinking & 36.26 \\
    Qwen3.5 9B-thinking & 35.16 \\
    Qwen3.7 Max & 30.77 \\
    Claude Opus 4.7-thinking (default) & 29.67 \\
    GLM 5.1-thinking (default) & 28.02 \\
    Qwen3.5 397B-A17B & 24.73 \\
    Claude Opus 4.7-nothinking & 24.18 \\
    GPT-5.5 (default) & 23.63 \\
    GPT-5.5-nothinking & 17.58 \\
    GPT-5.5-thinking & 17.58 \\
    Grok 4.3 (default) & 12.09 \\
    Qwen3.5 2B-thinking & 12.09 \\
    Grok 4.3-nothinking & 10.44 \\
    Qwen3.5 9B-nothinking & 8.24 \\
    Grok 4.3-thinking & 4.95 \\
    Qwen3.5 4B-nothinking & 3.30 \\
    Qwen3.5 2B-nothinking & 2.20 \\
    \bottomrule
  \end{tabular*}
\end{minipage}\par\vspace{0.65\baselineskip}

\subsection*{D.3 Post-Enforcement Tasks}

\subsubsection*{D.3.1 Penalty Type Identification}

\par\noindent\begin{minipage}{\linewidth}
  \centering
  \small
  \captionof{table}{Penalty Type Identification (Task Score).}
  \label{tab:appendix-d-10-penalty-type-identification-task-score}
  \begin{tabular*}{\linewidth}{@{\extracolsep{\fill}}lr@{}}
    \toprule
    \textbf{Model} & \textbf{Task Score} \\
    \midrule
    DeepSeek V4-Pro-nothinking & 90.00 \\
    Qwen3.6 27B & 89.44 \\
    GPT-5.5 (default) & 88.89 \\
    Qwen3.7 Max & 88.89 \\
    Qwen3.5 35B-A3B & 88.89 \\
    Qwen3.6 35B-A3B & 88.33 \\
    Qwen3.5 397B-A17B & 88.33 \\
    GPT-5.5-thinking & 87.22 \\
    Qwen3.5 27B & 87.22 \\
    GPT-5.5-nothinking & 86.67 \\
    Gemini 3.1 Pro Preview & 86.67 \\
    Grok 4.3-nothinking & 86.11 \\
    Qwen3.5 9B-nothinking & 86.11 \\
    Qwen3.5 4B-nothinking & 86.11 \\
    Qwen3.5 2B-nothinking & 86.11 \\
    GLM 5.1-nothinking & 85.56 \\
    GLM 5.1-thinking (default) & 85.00 \\
    Grok 4.3 (default) & 84.44 \\
    Claude Opus 4.7-nothinking & 83.89 \\
    DeepSeek V4-Pro-thinking (default) & 82.78 \\
    Grok 4.3-thinking & 82.78 \\
    Claude Opus 4.7-thinking (default) & 81.11 \\
    MiniMax M3-thinking (default) & 81.11 \\
    Qwen3.5 9B-thinking & 80.00 \\
    Qwen3.5 4B-thinking & 78.89 \\
    MiniMax M3-nothinking & 77.22 \\
    Qwen3.5 2B-thinking & 58.89 \\
    \bottomrule
  \end{tabular*}
\end{minipage}\par\vspace{0.65\baselineskip}

\subsubsection*{D.3.2 Penalty Decision}

\par\noindent\begin{minipage}{\linewidth}
  \centering
  \small
  \captionof{table}{Penalty Decision (Task Score).}
  \label{tab:appendix-d-11-penalty-decision-task-score}
  \begin{tabular*}{\linewidth}{@{\extracolsep{\fill}}lr@{}}
    \toprule
    \textbf{Model} & \textbf{Task Score} \\
    \midrule
    Qwen3.6 27B & 67.80 \\
    Qwen3.5 27B & 67.80 \\
    Qwen3.5 9B-nothinking & 66.10 \\
    Qwen3.7 Max & 62.71 \\
    Gemini 3.1 Pro Preview & 62.15 \\
    Claude Opus 4.7-thinking (default) & 61.02 \\
    Grok 4.3-nothinking & 61.02 \\
    GLM 5.1-nothinking & 60.45 \\
    Qwen3.5 35B-A3B & 60.45 \\
    Qwen3.5 2B-nothinking & 60.45 \\
    Qwen3.5 397B-A17B & 59.89 \\
    Claude Opus 4.7-nothinking & 58.76 \\
    GLM 5.1-thinking (default) & 58.76 \\
    Grok 4.3 (default) & 58.19 \\
    Qwen3.5 4B-nothinking & 57.63 \\
    DeepSeek V4-Pro-thinking (default) & 57.06 \\
    MiniMax M3-nothinking & 56.50 \\
    GPT-5.5 (default) & 55.93 \\
    DeepSeek V4-Pro-nothinking & 54.80 \\
    Qwen3.5 9B-thinking & 54.24 \\
    GPT-5.5-thinking & 53.11 \\
    MiniMax M3-thinking (default) & 53.11 \\
    Grok 4.3-thinking & 53.11 \\
    Qwen3.5 4B-thinking & 51.41 \\
    GPT-5.5-nothinking & 50.85 \\
    Qwen3.6 35B-A3B & 50.85 \\
    Qwen3.5 2B-thinking & 41.81 \\
    \bottomrule
  \end{tabular*}
\end{minipage}\par\vspace{0.65\baselineskip}

\subsubsection*{D.3.3 Legal Change Disturbance}

\par\noindent\begin{minipage}{\linewidth}
  \centering
  \small
  \captionof{table}{Legal Change Disturbance (Task Score).}
  \label{tab:appendix-d-12-legal-change-disturbance-task-score}
  \begin{tabular*}{\linewidth}{@{\extracolsep{\fill}}lr@{}}
    \toprule
    \textbf{Model} & \textbf{Task Score} \\
    \midrule
    Claude Opus 4.7-nothinking & 86.54 \\
    Claude Opus 4.7-thinking (default) & 85.90 \\
    Gemini 3.1 Pro Preview & 84.62 \\
    MiniMax M3-thinking (default) & 79.49 \\
    MiniMax M3-nothinking & 78.21 \\
    Qwen3.6 35B-A3B & 77.56 \\
    Qwen3.5 35B-A3B & 77.56 \\
    GPT-5.5 (default) & 76.92 \\
    GPT-5.5-nothinking & 75.64 \\
    Qwen3.7 Max & 75.00 \\
    GPT-5.5-thinking & 71.79 \\
    DeepSeek V4-Pro-thinking (default) & 71.15 \\
    Qwen3.6 27B & 69.87 \\
    Qwen3.5 397B-A17B & 69.23 \\
    DeepSeek V4-Pro-nothinking & 66.03 \\
    Qwen3.5 27B & 66.03 \\
    GLM 5.1-nothinking & 64.74 \\
    Grok 4.3 (default) & 64.10 \\
    Grok 4.3-nothinking & 62.18 \\
    GLM 5.1-thinking (default) & 55.77 \\
    Grok 4.3-thinking & 55.77 \\
    Qwen3.5 9B-nothinking & 51.92 \\
    Qwen3.5 4B-nothinking & 38.46 \\
    Qwen3.5 9B-thinking & 37.18 \\
    Qwen3.5 4B-thinking & 37.18 \\
    Qwen3.5 2B-thinking & 14.74 \\
    Qwen3.5 2B-nothinking & 14.10 \\
    \bottomrule
  \end{tabular*}
\end{minipage}\par\vspace{0.65\baselineskip}

\subsubsection*{D.3.4 Justification Assessment}

\par\noindent\begin{minipage}{\linewidth}
  \centering
  \small
  \captionof{table}{Justification Assessment (Task Score).}
  \label{tab:appendix-d-13-justification-assessment-task-score}
  \begin{tabular*}{\linewidth}{@{\extracolsep{\fill}}lr@{}}
    \toprule
    \textbf{Model} & \textbf{Task Score} \\
    \midrule
    GPT-5.5-thinking & 77.31 \\
    GLM 5.1-thinking (default) & 76.94 \\
    GPT-5.5-nothinking & 76.67 \\
    GPT-5.5 (default) & 75.56 \\
    Gemini 3.1 Pro Preview & 75.28 \\
    Qwen3.6 27B & 75.00 \\
    DeepSeek V4-Pro-thinking (default) & 74.81 \\
    GLM 5.1-nothinking & 73.98 \\
    Qwen3.5 397B-A17B & 73.98 \\
    Qwen3.7 Max & 73.70 \\
    Claude Opus 4.7-nothinking & 73.43 \\
    Claude Opus 4.7-thinking (default) & 72.87 \\
    Grok 4.3-thinking & 72.04 \\
    Grok 4.3-nothinking & 71.76 \\
    MiniMax M3-thinking (default) & 71.48 \\
    Qwen3.5 27B & 71.11 \\
    Grok 4.3 (default) & 70.65 \\
    Qwen3.5 9B-thinking & 70.56 \\
    Qwen3.6 35B-A3B & 70.37 \\
    Qwen3.5 35B-A3B & 70.37 \\
    Qwen3.5 9B-nothinking & 69.07 \\
    DeepSeek V4-Pro-nothinking & 68.33 \\
    Qwen3.5 4B-thinking & 66.20 \\
    MiniMax M3-nothinking & 58.70 \\
    Qwen3.5 4B-nothinking & 58.06 \\
    Qwen3.5 2B-thinking & 36.11 \\
    Qwen3.5 2B-nothinking & 34.35 \\
    \bottomrule
  \end{tabular*}
\end{minipage}\par\vspace{0.65\baselineskip}

\subsubsection*{D.3.5 Penalty Result Reasoning}

\par\noindent\begin{minipage}{\linewidth}
  \centering
  \small
  \captionof{table}{Penalty Result Reasoning (Task Score).}
  \label{tab:appendix-d-14-penalty-result-reasoning-task-score}
  \begin{tabular*}{\linewidth}{@{\extracolsep{\fill}}lr@{}}
    \toprule
    \textbf{Model} & \textbf{Task Score} \\
    \midrule
    Qwen3.7 Max & 85.35 \\
    GPT-5.5-nothinking & 85.28 \\
    GPT-5.5 (default) & 84.58 \\
    GPT-5.5-thinking & 84.24 \\
    Grok 4.3-thinking & 84.10 \\
    Claude Opus 4.7-thinking (default) & 83.82 \\
    Qwen3.6 27B & 83.12 \\
    Claude Opus 4.7-nothinking & 82.78 \\
    MiniMax M3-thinking (default) & 82.50 \\
    DeepSeek V4-Pro-thinking (default) & 82.08 \\
    Qwen3.5 27B & 81.88 \\
    Qwen3.5 397B-A17B & 81.74 \\
    GLM 5.1-thinking (default) & 81.46 \\
    GLM 5.1-nothinking & 81.25 \\
    Gemini 3.1 Pro Preview & 81.04 \\
    Qwen3.5 4B-nothinking & 80.69 \\
    Qwen3.6 35B-A3B & 80.07 \\
    Qwen3.5 35B-A3B & 80.00 \\
    DeepSeek V4-Pro-nothinking & 79.31 \\
    Grok 4.3 (default) & 77.64 \\
    Qwen3.5 9B-thinking & 77.64 \\
    Qwen3.5 9B-nothinking & 76.46 \\
    Grok 4.3-nothinking & 75.97 \\
    Qwen3.5 4B-thinking & 75.69 \\
    MiniMax M3-nothinking & 70.97 \\
    Qwen3.5 2B-nothinking & 53.82 \\
    Qwen3.5 2B-thinking & 52.01 \\
    \bottomrule
  \end{tabular*}
\end{minipage}\par\vspace{0.65\baselineskip}

\subsection*{D.4 Pollutant Secondary-Category Results}

\subsubsection*{D.4.1 Industrial Wastewater}

\par\noindent\begin{minipage}{\linewidth}
  \centering
  \small
  \captionof{table}{Industrial Wastewater (Category Score).}
  \label{tab:appendix-d-15-industrial-wastewater-category-score}
  \begin{tabular*}{\linewidth}{@{\extracolsep{\fill}}lr@{}}
    \toprule
    \textbf{Model} & \textbf{Category Score} \\
    \midrule
    GPT-5.5-nothinking & 64.10 \\
    GPT-5.5 (default) & 62.67 \\
    GPT-5.5-thinking & 60.01 \\
    Qwen3.7 Max & 59.93 \\
    Qwen3.6 27B & 58.25 \\
    Qwen3.6 35B-A3B & 57.09 \\
    DeepSeek V4-Pro-thinking (default) & 56.06 \\
    Claude Opus 4.7-nothinking & 55.97 \\
    MiniMax M3-nothinking & 55.93 \\
    Claude Opus 4.7-thinking (default) & 55.86 \\
    Grok 4.3-nothinking & 54.96 \\
    DeepSeek V4-Pro-nothinking & 54.87 \\
    Gemini 3.1 Pro Preview & 54.59 \\
    MiniMax M3-thinking (default) & 53.63 \\
    Grok 4.3 (default) & 53.55 \\
    Grok 4.3-thinking & 53.12 \\
    Qwen3.5 397B-A17B & 52.75 \\
    GLM 5.1-nothinking & 52.69 \\
    Qwen3.5 35B-A3B & 50.97 \\
    GLM 5.1-thinking (default) & 50.78 \\
    Qwen3.5 27B & 49.28 \\
    Qwen3.5 9B-nothinking & 42.54 \\
    Qwen3.5 4B-nothinking & 35.57 \\
    Qwen3.5 4B-thinking & 33.68 \\
    Qwen3.5 9B-thinking & 33.37 \\
    Qwen3.5 2B-nothinking & 25.91 \\
    Qwen3.5 2B-thinking & 25.33 \\
    \bottomrule
  \end{tabular*}
\end{minipage}\par\vspace{0.65\baselineskip}

\subsubsection*{D.4.2 Municipal Sewage}

\par\noindent\begin{minipage}{\linewidth}
  \centering
  \small
  \captionof{table}{Municipal Sewage (Category Score).}
  \label{tab:appendix-d-16-municipal-sewage-category-score}
  \begin{tabular*}{\linewidth}{@{\extracolsep{\fill}}lr@{}}
    \toprule
    \textbf{Model} & \textbf{Category Score} \\
    \midrule
    GPT-5.5-thinking & 62.13 \\
    Claude Opus 4.7-nothinking & 60.72 \\
    Claude Opus 4.7-thinking (default) & 59.53 \\
    GPT-5.5 (default) & 59.42 \\
    Gemini 3.1 Pro Preview & 58.19 \\
    Qwen3.7 Max & 57.80 \\
    Qwen3.6 27B & 57.76 \\
    GLM 5.1-nothinking & 57.66 \\
    DeepSeek V4-Pro-thinking (default) & 57.33 \\
    GPT-5.5-nothinking & 56.96 \\
    MiniMax M3-thinking (default) & 56.54 \\
    DeepSeek V4-Pro-nothinking & 56.40 \\
    Qwen3.6 35B-A3B & 56.10 \\
    Qwen3.5 397B-A17B & 54.57 \\
    GLM 5.1-thinking (default) & 53.55 \\
    Qwen3.5 27B & 53.02 \\
    Grok 4.3-nothinking & 52.25 \\
    Grok 4.3 (default) & 51.93 \\
    Qwen3.5 35B-A3B & 49.86 \\
    MiniMax M3-nothinking & 49.55 \\
    Grok 4.3-thinking & 47.25 \\
    Qwen3.5 9B-nothinking & 46.63 \\
    Qwen3.5 4B-nothinking & 44.47 \\
    Qwen3.5 4B-thinking & 43.48 \\
    Qwen3.5 9B-thinking & 43.14 \\
    Qwen3.5 2B-nothinking & 37.99 \\
    Qwen3.5 2B-thinking & 30.11 \\
    \bottomrule
  \end{tabular*}
\end{minipage}\par\vspace{0.65\baselineskip}

\subsubsection*{D.4.3 Agricultural Wastewater}

\par\noindent\begin{minipage}{\linewidth}
  \centering
  \small
  \captionof{table}{Agricultural Wastewater (Category Score).}
  \label{tab:appendix-d-17-agricultural-wastewater-category-score}
  \begin{tabular*}{\linewidth}{@{\extracolsep{\fill}}lr@{}}
    \toprule
    \textbf{Model} & \textbf{Category Score} \\
    \midrule
    GPT-5.5-nothinking & 69.76 \\
    GPT-5.5 (default) & 65.64 \\
    Qwen3.7 Max & 65.56 \\
    Qwen3.6 27B & 65.30 \\
    GPT-5.5-thinking & 65.07 \\
    Gemini 3.1 Pro Preview & 63.64 \\
    DeepSeek V4-Pro-thinking (default) & 63.06 \\
    Qwen3.5 397B-A17B & 62.37 \\
    Claude Opus 4.7-thinking (default) & 62.08 \\
    Qwen3.5 27B & 61.19 \\
    GLM 5.1-nothinking & 61.11 \\
    Claude Opus 4.7-nothinking & 60.78 \\
    Qwen3.6 35B-A3B & 60.53 \\
    Grok 4.3-nothinking & 60.13 \\
    DeepSeek V4-Pro-nothinking & 60.09 \\
    Grok 4.3 (default) & 59.85 \\
    MiniMax M3-thinking (default) & 59.16 \\
    GLM 5.1-thinking (default) & 59.12 \\
    Grok 4.3-thinking & 59.02 \\
    MiniMax M3-nothinking & 56.60 \\
    Qwen3.5 35B-A3B & 54.40 \\
    Qwen3.5 9B-nothinking & 48.30 \\
    Qwen3.5 9B-thinking & 45.94 \\
    Qwen3.5 4B-thinking & 44.62 \\
    Qwen3.5 4B-nothinking & 43.59 \\
    Qwen3.5 2B-thinking & 34.82 \\
    Qwen3.5 2B-nothinking & 32.84 \\
    \bottomrule
  \end{tabular*}
\end{minipage}\par\vspace{0.65\baselineskip}

\subsubsection*{D.4.4 Combustion Emissions}

\par\noindent\begin{minipage}{\linewidth}
  \centering
  \small
  \captionof{table}{Combustion Emissions (Category Score).}
  \label{tab:appendix-d-18-combustion-emissions-category-score}
  \begin{tabular*}{\linewidth}{@{\extracolsep{\fill}}lr@{}}
    \toprule
    \textbf{Model} & \textbf{Category Score} \\
    \midrule
    GPT-5.5-thinking & 61.16 \\
    Qwen3.7 Max & 60.53 \\
    Qwen3.6 27B & 60.06 \\
    MiniMax M3-thinking (default) & 59.96 \\
    GPT-5.5-nothinking & 59.89 \\
    GPT-5.5 (default) & 59.42 \\
    Qwen3.6 35B-A3B & 58.46 \\
    Qwen3.5 27B & 57.18 \\
    Gemini 3.1 Pro Preview & 55.96 \\
    DeepSeek V4-Pro-nothinking & 55.39 \\
    Claude Opus 4.7-thinking (default) & 54.78 \\
    GLM 5.1-nothinking & 54.60 \\
    DeepSeek V4-Pro-thinking (default) & 54.39 \\
    GLM 5.1-thinking (default) & 53.96 \\
    Qwen3.5 397B-A17B & 53.94 \\
    Claude Opus 4.7-nothinking & 53.52 \\
    Qwen3.5 35B-A3B & 53.33 \\
    MiniMax M3-nothinking & 53.15 \\
    Grok 4.3-nothinking & 51.82 \\
    Grok 4.3 (default) & 49.36 \\
    Grok 4.3-thinking & 48.45 \\
    Qwen3.5 9B-nothinking & 45.11 \\
    Qwen3.5 4B-nothinking & 43.73 \\
    Qwen3.5 9B-thinking & 42.02 \\
    Qwen3.5 4B-thinking & 41.67 \\
    Qwen3.5 2B-nothinking & 36.18 \\
    Qwen3.5 2B-thinking & 28.85 \\
    \bottomrule
  \end{tabular*}
\end{minipage}\par\vspace{0.65\baselineskip}

\subsubsection*{D.4.5 Industrial Air Pollution}

\par\noindent\begin{minipage}{\linewidth}
  \centering
  \small
  \captionof{table}{Industrial Air Pollution (Category Score).}
  \label{tab:appendix-d-19-industrial-air-pollution-category-score}
  \begin{tabular*}{\linewidth}{@{\extracolsep{\fill}}lr@{}}
    \toprule
    \textbf{Model} & \textbf{Category Score} \\
    \midrule
    GPT-5.5 (default) & 62.94 \\
    GPT-5.5-nothinking & 61.71 \\
    GPT-5.5-thinking & 59.10 \\
    Qwen3.6 27B & 58.82 \\
    Qwen3.7 Max & 58.00 \\
    Qwen3.6 35B-A3B & 57.84 \\
    Grok 4.3-nothinking & 57.18 \\
    Claude Opus 4.7-thinking (default) & 56.97 \\
    Qwen3.5 35B-A3B & 56.70 \\
    DeepSeek V4-Pro-nothinking & 56.44 \\
    Qwen3.5 27B & 56.19 \\
    Gemini 3.1 Pro Preview & 55.96 \\
    GLM 5.1-nothinking & 55.67 \\
    DeepSeek V4-Pro-thinking (default) & 55.20 \\
    GLM 5.1-thinking (default) & 54.92 \\
    MiniMax M3-thinking (default) & 54.31 \\
    Claude Opus 4.7-nothinking & 54.14 \\
    Grok 4.3 (default) & 53.02 \\
    Qwen3.5 397B-A17B & 52.96 \\
    Grok 4.3-thinking & 52.50 \\
    MiniMax M3-nothinking & 52.36 \\
    Qwen3.5 4B-nothinking & 46.55 \\
    Qwen3.5 9B-thinking & 44.14 \\
    Qwen3.5 9B-nothinking & 43.62 \\
    Qwen3.5 4B-thinking & 42.55 \\
    Qwen3.5 2B-nothinking & 35.74 \\
    Qwen3.5 2B-thinking & 29.54 \\
    \bottomrule
  \end{tabular*}
\end{minipage}\par\vspace{0.65\baselineskip}

\subsubsection*{D.4.6 Mobile Source Air Pollution}

\par\noindent\begin{minipage}{\linewidth}
  \centering
  \small
  \captionof{table}{Mobile Source Air Pollution (Category Score).}
  \label{tab:appendix-d-20-mobile-source-air-pollution-category-score}
  \begin{tabular*}{\linewidth}{@{\extracolsep{\fill}}lr@{}}
    \toprule
    \textbf{Model} & \textbf{Category Score} \\
    \midrule
    GPT-5.5-nothinking & 63.61 \\
    GLM 5.1-nothinking & 63.50 \\
    GPT-5.5 (default) & 63.20 \\
    GPT-5.5-thinking & 63.09 \\
    Claude Opus 4.7-nothinking & 61.51 \\
    Gemini 3.1 Pro Preview & 60.00 \\
    Qwen3.7 Max & 59.70 \\
    Qwen3.6 27B & 58.08 \\
    Claude Opus 4.7-thinking (default) & 57.42 \\
    Qwen3.6 35B-A3B & 57.32 \\
    DeepSeek V4-Pro-thinking (default) & 57.31 \\
    DeepSeek V4-Pro-nothinking & 56.60 \\
    GLM 5.1-thinking (default) & 54.51 \\
    Qwen3.5 27B & 54.38 \\
    Qwen3.5 397B-A17B & 52.08 \\
    Grok 4.3-nothinking & 51.84 \\
    MiniMax M3-nothinking & 50.89 \\
    Qwen3.5 35B-A3B & 50.45 \\
    MiniMax M3-thinking (default) & 50.15 \\
    Grok 4.3 (default) & 49.44 \\
    Qwen3.5 9B-nothinking & 48.21 \\
    Grok 4.3-thinking & 48.21 \\
    Qwen3.5 9B-thinking & 43.95 \\
    Qwen3.5 4B-thinking & 41.80 \\
    Qwen3.5 4B-nothinking & 37.88 \\
    Qwen3.5 2B-nothinking & 33.08 \\
    Qwen3.5 2B-thinking & 28.58 \\
    \bottomrule
  \end{tabular*}
\end{minipage}\par\vspace{0.65\baselineskip}

\subsubsection*{D.4.7 Agricultural Air Pollution}

\par\noindent\begin{minipage}{\linewidth}
  \centering
  \small
  \captionof{table}{Agricultural Air Pollution (Category Score).}
  \label{tab:appendix-d-21-agricultural-air-pollution-category-score}
  \begin{tabular*}{\linewidth}{@{\extracolsep{\fill}}lr@{}}
    \toprule
    \textbf{Model} & \textbf{Category Score} \\
    \midrule
    GPT-5.5 (default) & 64.15 \\
    GPT-5.5-thinking & 62.24 \\
    GPT-5.5-nothinking & 61.67 \\
    Claude Opus 4.7-nothinking & 58.56 \\
    Gemini 3.1 Pro Preview & 58.39 \\
    Qwen3.7 Max & 57.43 \\
    DeepSeek V4-Pro-thinking (default) & 56.82 \\
    MiniMax M3-thinking (default) & 55.92 \\
    GLM 5.1-nothinking & 55.09 \\
    Claude Opus 4.7-thinking (default) & 54.95 \\
    Qwen3.6 27B & 54.64 \\
    DeepSeek V4-Pro-nothinking & 54.13 \\
    Qwen3.6 35B-A3B & 54.10 \\
    Grok 4.3-nothinking & 53.39 \\
    Qwen3.5 397B-A17B & 53.31 \\
    GLM 5.1-thinking (default) & 52.33 \\
    Grok 4.3 (default) & 51.91 \\
    MiniMax M3-nothinking & 51.22 \\
    Qwen3.5 27B & 50.61 \\
    Qwen3.5 35B-A3B & 49.85 \\
    Grok 4.3-thinking & 48.46 \\
    Qwen3.5 9B-nothinking & 44.32 \\
    Qwen3.5 4B-nothinking & 40.33 \\
    Qwen3.5 4B-thinking & 36.74 \\
    Qwen3.5 9B-thinking & 35.30 \\
    Qwen3.5 2B-nothinking & 33.57 \\
    Qwen3.5 2B-thinking & 28.21 \\
    \bottomrule
  \end{tabular*}
\end{minipage}\par\vspace{0.65\baselineskip}

\subsubsection*{D.4.8 Industrial Solid Waste}

\par\noindent\begin{minipage}{\linewidth}
  \centering
  \small
  \captionof{table}{Industrial Solid Waste (Category Score).}
  \label{tab:appendix-d-22-industrial-solid-waste-category-score}
  \begin{tabular*}{\linewidth}{@{\extracolsep{\fill}}lr@{}}
    \toprule
    \textbf{Model} & \textbf{Category Score} \\
    \midrule
    GPT-5.5-nothinking & 64.06 \\
    GPT-5.5 (default) & 62.98 \\
    GPT-5.5-thinking & 62.55 \\
    Gemini 3.1 Pro Preview & 62.31 \\
    Qwen3.7 Max & 59.88 \\
    GLM 5.1-nothinking & 58.51 \\
    DeepSeek V4-Pro-thinking (default) & 58.06 \\
    Qwen3.6 27B & 57.84 \\
    DeepSeek V4-Pro-nothinking & 57.83 \\
    Grok 4.3-nothinking & 56.84 \\
    Claude Opus 4.7-thinking (default) & 56.82 \\
    Qwen3.6 35B-A3B & 56.78 \\
    Qwen3.5 27B & 55.37 \\
    Claude Opus 4.7-nothinking & 55.32 \\
    MiniMax M3-thinking (default) & 54.53 \\
    Grok 4.3-thinking & 53.55 \\
    GLM 5.1-thinking (default) & 53.50 \\
    Qwen3.5 35B-A3B & 52.48 \\
    Qwen3.5 397B-A17B & 51.95 \\
    Grok 4.3 (default) & 50.87 \\
    Qwen3.5 9B-nothinking & 49.87 \\
    MiniMax M3-nothinking & 47.01 \\
    Qwen3.5 9B-thinking & 45.25 \\
    Qwen3.5 4B-nothinking & 43.06 \\
    Qwen3.5 4B-thinking & 38.74 \\
    Qwen3.5 2B-nothinking & 33.28 \\
    Qwen3.5 2B-thinking & 28.51 \\
    \bottomrule
  \end{tabular*}
\end{minipage}\par\vspace{0.65\baselineskip}

\subsubsection*{D.4.9 Municipal Solid Waste}

\par\noindent\begin{minipage}{\linewidth}
  \centering
  \small
  \captionof{table}{Municipal Solid Waste (Category Score).}
  \label{tab:appendix-d-23-municipal-solid-waste-category-score}
  \begin{tabular*}{\linewidth}{@{\extracolsep{\fill}}lr@{}}
    \toprule
    \textbf{Model} & \textbf{Category Score} \\
    \midrule
    GPT-5.5 (default) & 61.08 \\
    GPT-5.5-nothinking & 60.14 \\
    Qwen3.7 Max & 60.07 \\
    GPT-5.5-thinking & 58.25 \\
    Qwen3.6 27B & 57.42 \\
    DeepSeek V4-Pro-thinking (default) & 57.10 \\
    Qwen3.6 35B-A3B & 55.06 \\
    Claude Opus 4.7-thinking (default) & 53.52 \\
    DeepSeek V4-Pro-nothinking & 53.43 \\
    Qwen3.5 397B-A17B & 52.10 \\
    Claude Opus 4.7-nothinking & 52.01 \\
    Gemini 3.1 Pro Preview & 51.61 \\
    GLM 5.1-nothinking & 51.35 \\
    MiniMax M3-thinking (default) & 49.92 \\
    Qwen3.5 27B & 49.77 \\
    Grok 4.3-nothinking & 49.28 \\
    GLM 5.1-thinking (default) & 48.58 \\
    Grok 4.3 (default) & 47.78 \\
    MiniMax M3-nothinking & 47.58 \\
    Qwen3.5 35B-A3B & 47.27 \\
    Grok 4.3-thinking & 46.05 \\
    Qwen3.5 9B-thinking & 43.37 \\
    Qwen3.5 4B-thinking & 42.32 \\
    Qwen3.5 9B-nothinking & 41.86 \\
    Qwen3.5 4B-nothinking & 40.47 \\
    Qwen3.5 2B-nothinking & 35.37 \\
    Qwen3.5 2B-thinking & 23.87 \\
    \bottomrule
  \end{tabular*}
\end{minipage}\par\vspace{0.65\baselineskip}

\subsubsection*{D.4.10 Construction Waste}

\par\noindent\begin{minipage}{\linewidth}
  \centering
  \small
  \captionof{table}{Construction Waste (Category Score).}
  \label{tab:appendix-d-24-construction-waste-category-score}
  \begin{tabular*}{\linewidth}{@{\extracolsep{\fill}}lr@{}}
    \toprule
    \textbf{Model} & \textbf{Category Score} \\
    \midrule
    GPT-5.5-thinking & 70.09 \\
    GPT-5.5 (default) & 67.69 \\
    GPT-5.5-nothinking & 65.19 \\
    Claude Opus 4.7-nothinking & 63.08 \\
    Qwen3.6 35B-A3B & 62.21 \\
    Qwen3.6 27B & 61.20 \\
    Qwen3.7 Max & 60.62 \\
    DeepSeek V4-Pro-thinking (default) & 59.89 \\
    Qwen3.5 397B-A17B & 59.40 \\
    Claude Opus 4.7-thinking (default) & 59.30 \\
    MiniMax M3-thinking (default) & 58.11 \\
    Gemini 3.1 Pro Preview & 56.84 \\
    Grok 4.3-thinking & 56.64 \\
    GLM 5.1-nothinking & 56.62 \\
    DeepSeek V4-Pro-nothinking & 56.16 \\
    Qwen3.5 35B-A3B & 55.25 \\
    Grok 4.3 (default) & 54.75 \\
    Grok 4.3-nothinking & 54.66 \\
    MiniMax M3-nothinking & 54.51 \\
    Qwen3.5 27B & 54.23 \\
    GLM 5.1-thinking (default) & 52.52 \\
    Qwen3.5 9B-nothinking & 50.69 \\
    Qwen3.5 9B-thinking & 46.93 \\
    Qwen3.5 4B-thinking & 43.50 \\
    Qwen3.5 4B-nothinking & 43.18 \\
    Qwen3.5 2B-nothinking & 37.02 \\
    Qwen3.5 2B-thinking & 32.94 \\
    \bottomrule
  \end{tabular*}
\end{minipage}\par\vspace{0.65\baselineskip}

\subsubsection*{D.4.11 Agricultural Solid Waste}

\par\noindent\begin{minipage}{\linewidth}
  \centering
  \small
  \captionof{table}{Agricultural Solid Waste (Category Score).}
  \label{tab:appendix-d-25-agricultural-solid-waste-category-score}
  \begin{tabular*}{\linewidth}{@{\extracolsep{\fill}}lr@{}}
    \toprule
    \textbf{Model} & \textbf{Category Score} \\
    \midrule
    GPT-5.5 (default) & 70.24 \\
    GPT-5.5-thinking & 69.59 \\
    GPT-5.5-nothinking & 65.82 \\
    GLM 5.1-nothinking & 63.23 \\
    Qwen3.6 35B-A3B & 63.12 \\
    Gemini 3.1 Pro Preview & 62.79 \\
    Qwen3.7 Max & 62.03 \\
    DeepSeek V4-Pro-thinking (default) & 61.71 \\
    Claude Opus 4.7-nothinking & 61.46 \\
    Claude Opus 4.7-thinking (default) & 61.39 \\
    MiniMax M3-thinking (default) & 59.55 \\
    Qwen3.5 35B-A3B & 58.93 \\
    DeepSeek V4-Pro-nothinking & 58.63 \\
    Qwen3.5 397B-A17B & 58.28 \\
    Qwen3.6 27B & 57.92 \\
    Grok 4.3-thinking & 56.52 \\
    Qwen3.5 27B & 56.41 \\
    Grok 4.3-nothinking & 55.54 \\
    GLM 5.1-thinking (default) & 55.16 \\
    Grok 4.3 (default) & 53.12 \\
    Qwen3.5 9B-nothinking & 51.98 \\
    MiniMax M3-nothinking & 51.80 \\
    Qwen3.5 4B-thinking & 46.02 \\
    Qwen3.5 4B-nothinking & 45.60 \\
    Qwen3.5 9B-thinking & 44.14 \\
    Qwen3.5 2B-thinking & 36.78 \\
    Qwen3.5 2B-nothinking & 35.10 \\
    \bottomrule
  \end{tabular*}
\end{minipage}\par\vspace{0.65\baselineskip}

\subsubsection*{D.4.12 Hazardous Waste}

\par\noindent\begin{minipage}{\linewidth}
  \centering
  \small
  \captionof{table}{Hazardous Waste (Category Score).}
  \label{tab:appendix-d-26-hazardous-waste-category-score}
  \begin{tabular*}{\linewidth}{@{\extracolsep{\fill}}lr@{}}
    \toprule
    \textbf{Model} & \textbf{Category Score} \\
    \midrule
    GPT-5.5-nothinking & 65.68 \\
    Claude Opus 4.7-nothinking & 64.11 \\
    GPT-5.5 (default) & 63.51 \\
    GPT-5.5-thinking & 62.18 \\
    Claude Opus 4.7-thinking (default) & 59.75 \\
    Qwen3.7 Max & 57.41 \\
    GLM 5.1-nothinking & 56.81 \\
    Qwen3.6 35B-A3B & 56.62 \\
    Qwen3.6 27B & 55.80 \\
    Grok 4.3-nothinking & 54.85 \\
    DeepSeek V4-Pro-nothinking & 54.80 \\
    DeepSeek V4-Pro-thinking (default) & 54.57 \\
    Gemini 3.1 Pro Preview & 54.27 \\
    MiniMax M3-thinking (default) & 53.93 \\
    Grok 4.3-thinking & 53.60 \\
    Qwen3.5 27B & 52.93 \\
    GLM 5.1-thinking (default) & 52.84 \\
    Grok 4.3 (default) & 52.64 \\
    Qwen3.5 397B-A17B & 52.19 \\
    MiniMax M3-nothinking & 48.13 \\
    Qwen3.5 35B-A3B & 47.95 \\
    Qwen3.5 9B-nothinking & 43.84 \\
    Qwen3.5 4B-thinking & 37.84 \\
    Qwen3.5 4B-nothinking & 37.42 \\
    Qwen3.5 9B-thinking & 34.73 \\
    Qwen3.5 2B-nothinking & 30.26 \\
    Qwen3.5 2B-thinking & 23.78 \\
    \bottomrule
  \end{tabular*}
\end{minipage}\par\vspace{0.65\baselineskip}

\section{Detailed Case Analysis}
Appendix E uses representative cases and raw model outputs to explain the mechanisms behind three recurring patterns: why rule‑based enforcement decisions are easier than evidence‑chain reasoning, why medium‑sized models can approach the capability boundary in structured tasks, and why Thinking mode is only useful when the reasoning structure is controllable.

\subsection*{E.1 Rule-based enforcement decisions are easier than evidence-chain reasoning}

The lifecycle and cross-task analyses show a clear contrast: models are relatively stable in post-enforcement rule-application tasks, but much less reliable in in-enforcement tasks involving evidentiary relationships. To explain this contrast, we compare two representative cases. The first is a penalty-type classification task, which mainly requires rule matching under clear facts. The second is a contradiction-monitoring task, which requires models to identify conflicts across multiple evidence materials.

\par\noindent\begin{minipage}{\linewidth}
  \centering
  \includegraphics[width=\linewidth,height=0.72\textheight,keepaspectratio]{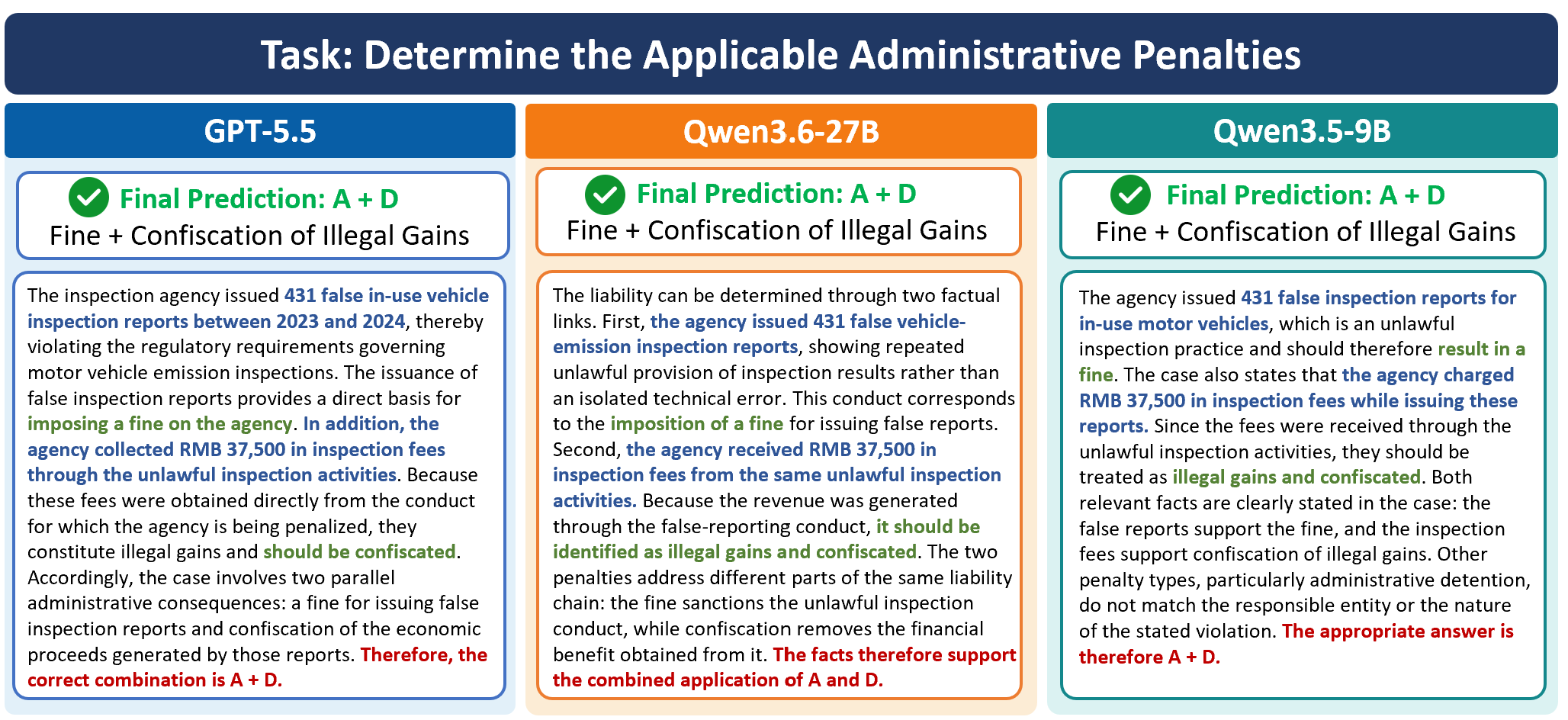}
  \captionof{figure}{Penalty-type classification in a vehicle-emission inspection violation case}
  \label{fig:appendix-e-1-penalty-classification-vehicle-inspection-a-1}
\end{minipage}\par\vspace{0.65\baselineskip}

The penalty-type classification case concerns a motor vehicle inspection agency that issued 431 false in-use vehicle inspection reports between 2023 and 2024 and charged RMB 37,500 in inspection fees. The task asks models to determine the applicable administrative penalties. The gold answer is A + D, namely a fine and confiscation of illegal gains. The contradiction-monitoring case concerns an agricultural wastewater inspection at a pig farm. The materials include a black-membrane temporary storage pond, an open channel, a pit pond, the operating status of a fermentation bed, and rapid water-quality test results. The task requires models to distinguish valid evidence from contradictory or unreliable materials. In this subdomain, all three model groups perform poorly: open-source LLMs score 20, closed-source LLMs score 23, and open-source SLMs score only 9. The gold annotation identifies evidence items 6, 7, and 8 as contradictory, involving sampling-location deviation, conflicting descriptions of facility operation, and insufficient rapid-test procedures.

\par\noindent\begin{minipage}{\linewidth}
  \centering
  \includegraphics[width=\linewidth,height=0.72\textheight,keepaspectratio]{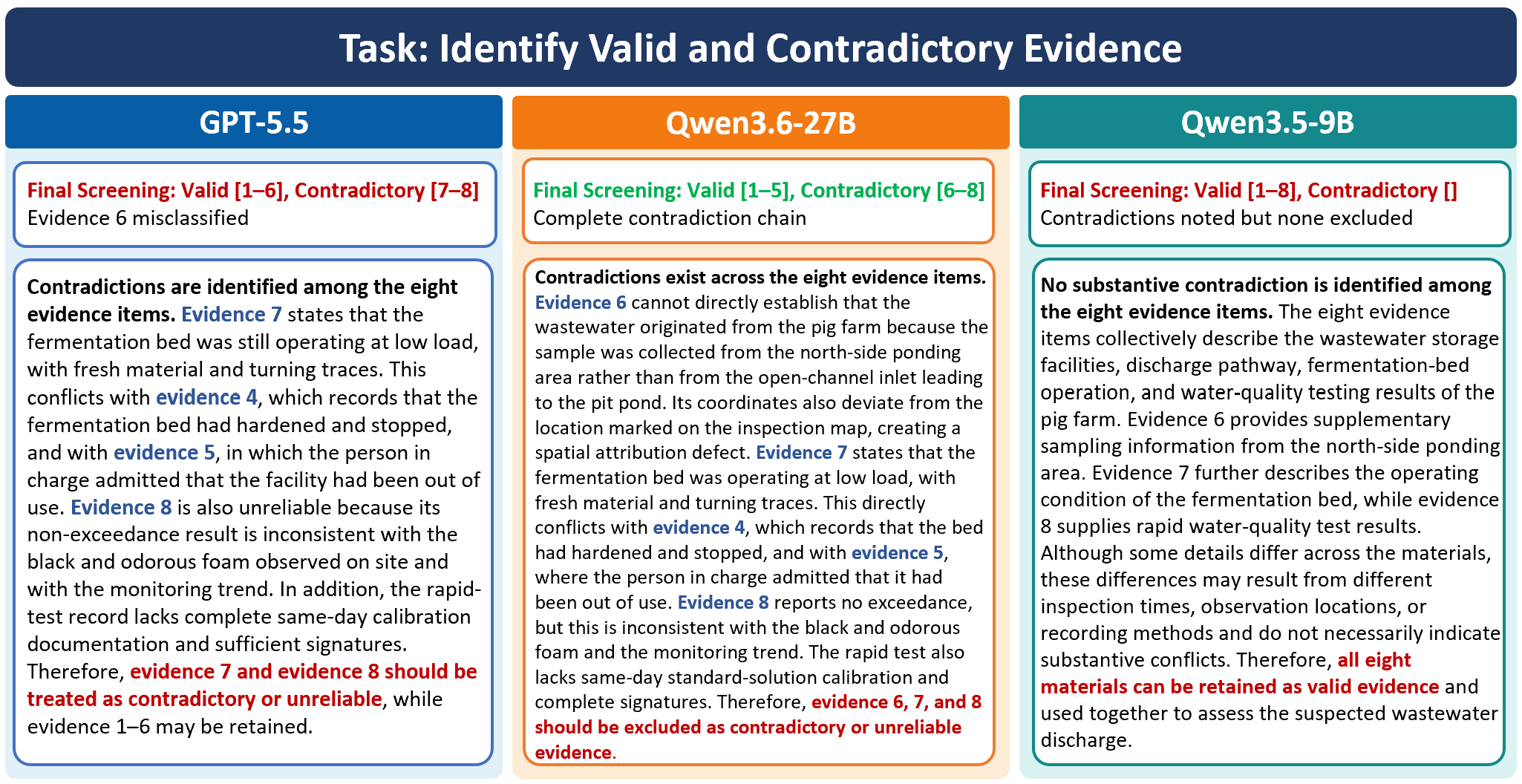}
  \captionof{figure}{Contradiction monitoring in an agricultural wastewater evidence case}
  \label{fig:appendix-e-2-contradiction-monitoring-in-an-agricultural-wastewater-case-a-2}
\end{minipage}\par\vspace{0.65\baselineskip}

The model outputs show why these two tasks differ sharply. In the penalty-type classification case in Figure E.1, GPT-5.5, Qwen3.6-27B, and Qwen3.5-9B all select the correct answer. Their reasoning paths are short and stable: they map the false inspection reports to a fine and the inspection fees to confiscation of illegal gains. By contrast, the contradiction-monitoring case in Figure E.2 requires cross-evidence calibration. GPT-5.5 captures some surface-level conflicts but misses the sampling-location deviation in evidence item 6. Qwen3.6-27B performs better by identifying the facility-operation conflict, the spatial shift in the sampling point, and the procedural defect in the rapid test. Qwen3.5-9B, however, tends to treat all materials as valid evidence and fails to separate evidentiary weight from logical conflict.

These two cases indicate that the difficulty of environmental enforcement does not begin with legal knowledge recall. In rule-based penalty classification, the facts are explicit, the legal consequences are stable, and the output format is fixed, so even smaller models can produce correct judgments through normative matching. In contradiction monitoring, however, models must jointly assess spatial location, facility status, testing procedure, and evidence credibility. The task shifts from single-fact recognition to cross-evidence relational reasoning. This explains why current LLMs are stronger in rule application and standardized enforcement expression, but remain weak in fact calibration, evidence exclusion, and procedure-constrained judgment under complex evidence systems.

\subsection*{E.2 Medium-sized dense models approach the capability boundary in structured tasks}

The overall AES and IEI results suggest a clear marginal effect in environmental enforcement. Qwen3.6-27B reaches an AES of 61.0, approaching the leading model group, while the much larger Qwen3.5-397B-A17B MoE achieves an AES of only 56.1, lower than several medium-sized open-source models. This section uses two cases to explain this pattern. The first case shows that medium-sized dense models may already reach sufficient capability in structured tasks, while the second shows that larger scale or MoE architecture does not necessarily overcome evidence-chain reasoning bottlenecks.

\par\noindent\begin{minipage}{\linewidth}
  \centering
  \includegraphics[width=\linewidth,height=0.72\textheight,keepaspectratio]{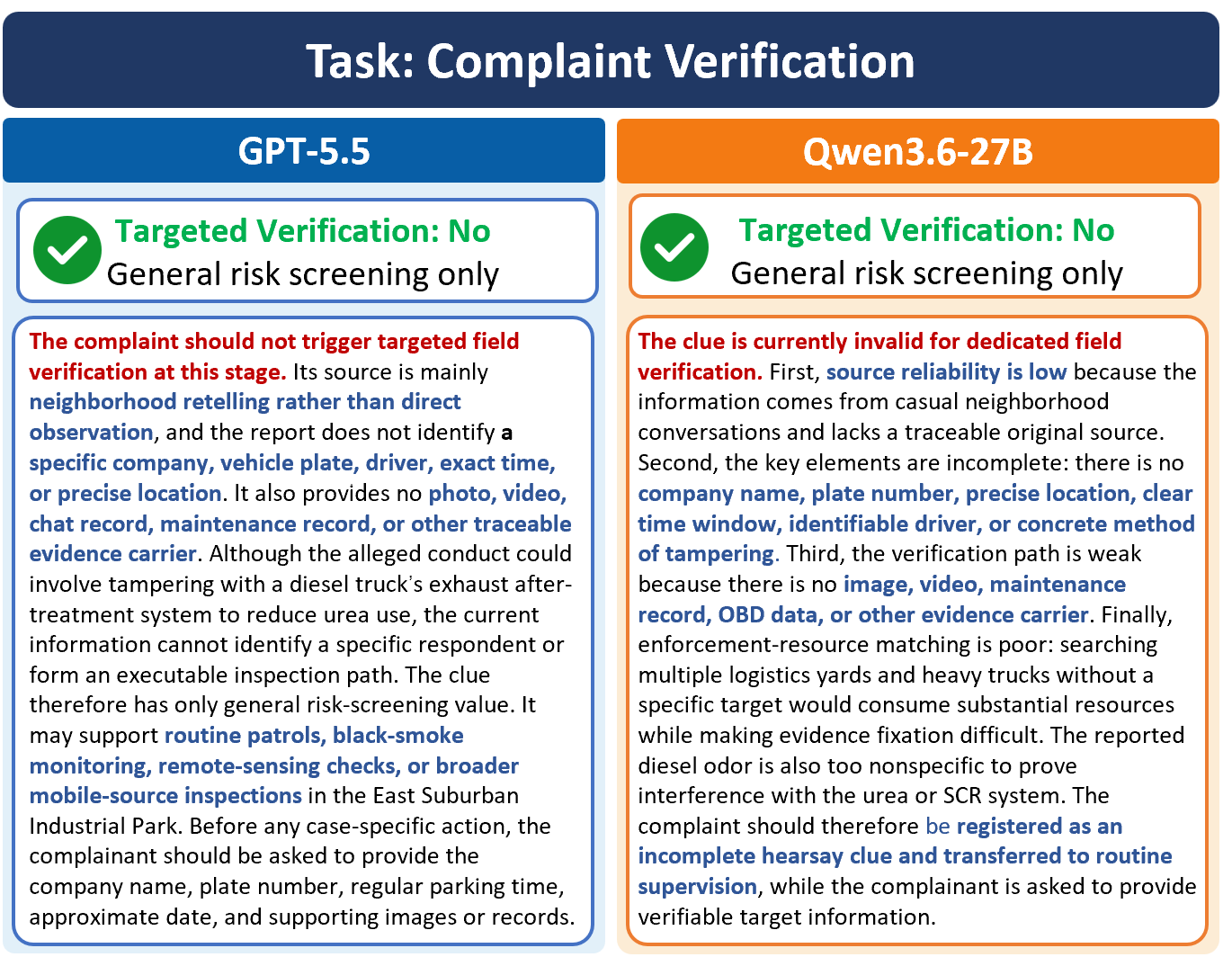}
  \captionof{figure}{Invalid mobile-source air pollution complaint verification case.}
  \label{fig:appendix-e-3-invalid-mobile-source-air-pollution-complaint-verification-b-1}
\end{minipage}\par\vspace{0.65\baselineskip}

The first case is an invalid mobile-source air pollution complaint. The clue comes mainly from neighborhood conversations and lacks key verification elements, including a specific company name, vehicle license plate, driver identity, clear time, photos or videos, and maintenance records. The task is to determine whether the clue already meets the conditions for a dedicated on-site inspection. The second case is a hazardous-waste multi-evidence integration task involving a tire repair shop that illegally sold used oil. The gold standard requires the model to integrate on-site inspection records, photographs, inquiry transcripts, and identification reports into a proof chain: the repair shop should first be identified as the generator and management subject of hazardous waste, and the illegal transfer should then be inferred from seven barrels of used oil, HW08 identification, an oil pump and a small truck, oil leakage traces, the purchaser's arrival, failure to verify disposal qualifications, missing labels, and absence of transfer manifests.

\par\noindent\begin{minipage}{\linewidth}
  \centering
  \includegraphics[width=\linewidth,height=0.72\textheight,keepaspectratio]{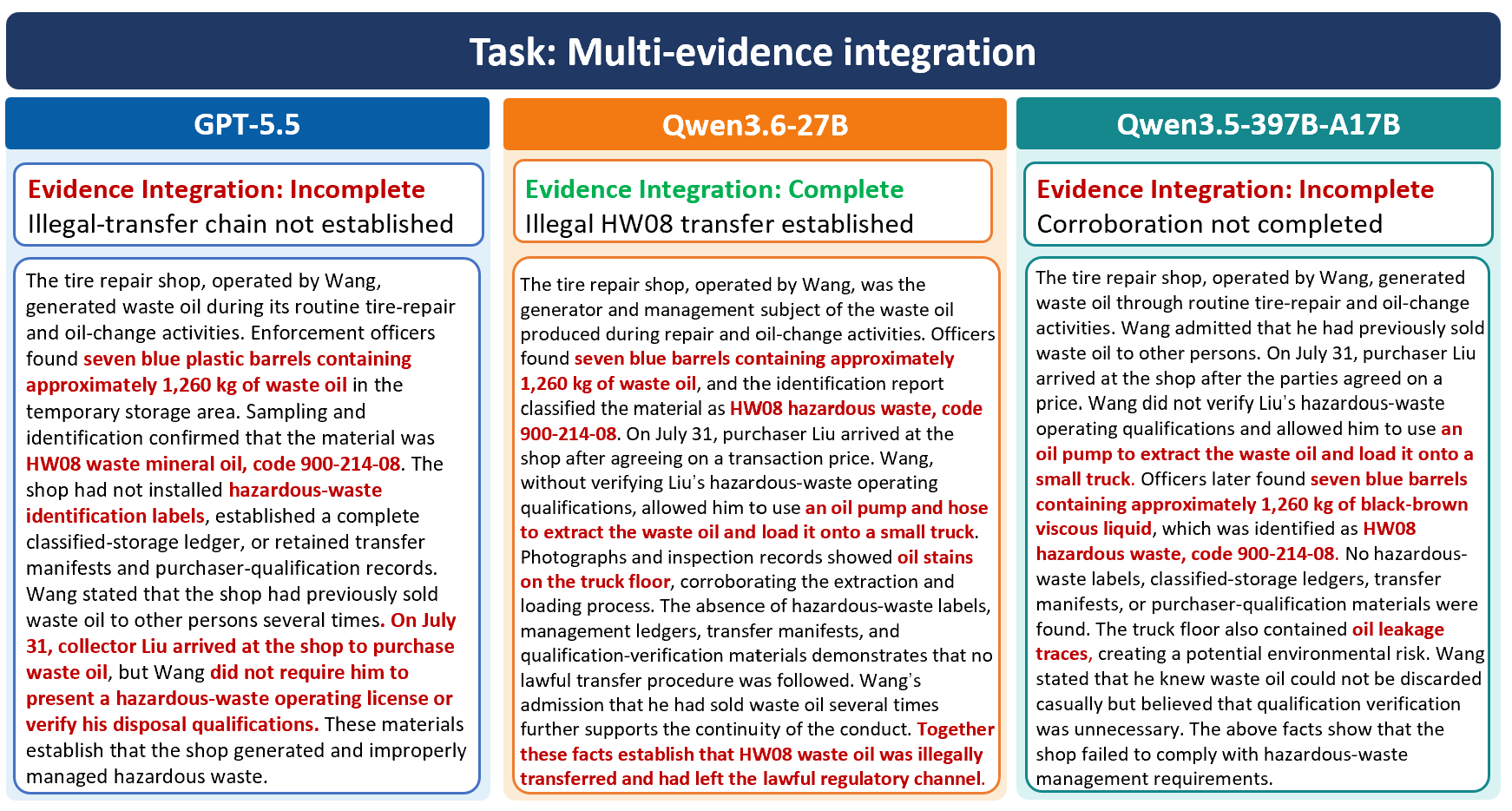}
  \captionof{figure}{Hazardous-waste multi-evidence integration case.}
  \label{fig:appendix-e-4-hazardous-waste-multi-evidence-integration-case-b-2}
\end{minipage}\par\vspace{0.65\baselineskip}

The model outputs show two different patterns. In the complaint verification case in Figure E.3, GPT-5.5 and Qwen3.6-27B reach the same enforcement judgment: the clue is not yet suitable for a dedicated on-site inspection and should instead enter general risk screening. Their differences mainly lie in expression and reasoning organization. GPT-5.5 proceeds from existing facts to information gaps, verifiability, and follow-up handling, while Qwen3.6-27B evaluates source reliability, element completeness, verification pathways, and enforcement-resource matching. In the hazardous-waste case in Figure E.4, however, Qwen3.6-27B passes the evaluation, whereas GPT-5.5 and Qwen3.5-397B-A17B do not. The latter two models cover many facts, but their outputs mainly remain at the level of material listing or case reconstruction. By contrast, Qwen3.6-27B links the purchaser's arrival, oil-pumping tools, small truck, seven barrels of used oil, HW08 identification, missing labels and transfer materials, and Wang's admission of repeated sales into a coherent proof chain for illegal hazardous-waste transfer.

These cases clarify why further scaling does not always translate into higher enforcement value. In structured tasks with clear factual elements and explicit verification standards, medium-sized models may already reach the level of professional sufficiency required for enforcement assistance; larger models mainly improve expression completeness rather than the core judgment. In evidence-chain tasks, however, larger scale or MoE architecture does not automatically produce better reasoning. The decisive factor is not how many facts a model can mention, but whether it can connect evidence nodes with legal elements and enforcement conclusions. This explains why medium-sized dense models can approach the capability boundary in structured tasks, while even larger models may still fail when the task requires evidence-chain construction.

\subsection*{E.3 Thinking helps only when reasoning structure is useful and controllable}

The preceding results show that Thinking mode is not a universal enhancement strategy in environmental enforcement. Its value depends on whether explicit reasoning helps the model organize task-relevant judgment conditions while remaining constrained by evidence boundaries and task objectives. This subsection therefore compares the same model under Thinking and No-thinking modes in two cases. The first case examines whether Thinking helps when the task requires identifying information gaps and verification thresholds. The second examines whether Thinking remains reliable when the task requires faithful extraction from complex, multi-source materials.

\par\noindent\begin{minipage}{\linewidth}
  \centering
  \includegraphics[width=\linewidth,height=0.72\textheight,keepaspectratio]{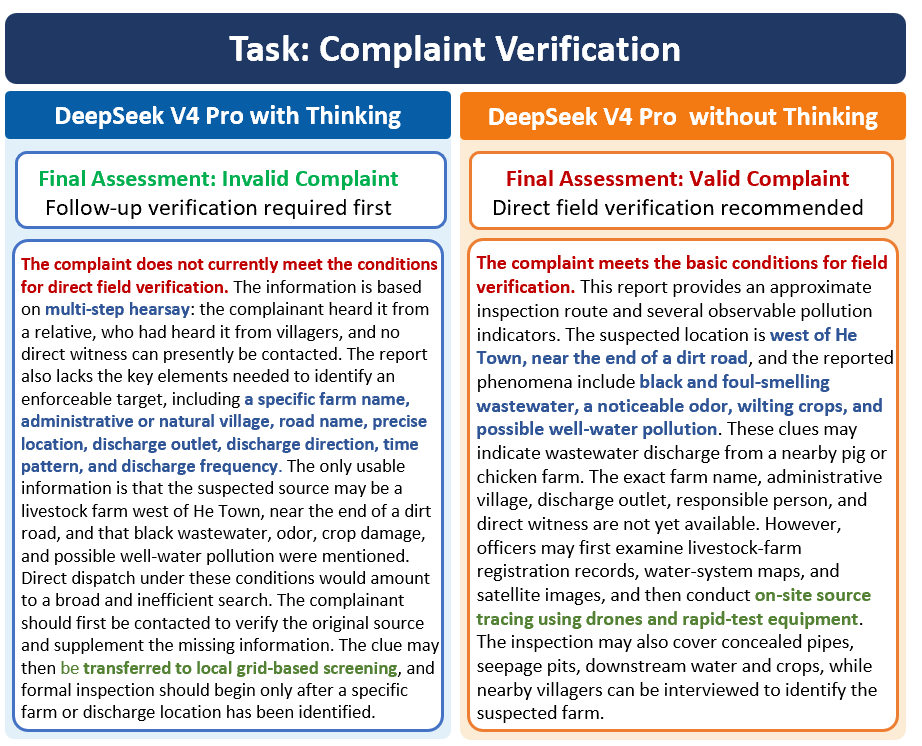}
  \captionof{figure}{Thinking-mode comparison for DeepSeek V4-Pro in invalid complaint verification.}
  \label{fig:appendix-e-5-thinking-mode-comparison-for-deepseek-v4-pro}
\end{minipage}\par\vspace{0.65\baselineskip}

The first case is an invalid complaint concerning suspected black and odorous wastewater discharge from a livestock farm. The complaint source is weak: the complainant is not a direct witness, but heard the information through several layers of hearsay. The text also lacks a specific farm name, administrative village, discharge outlet, discharge time, direct evidence, and a contactable witness. The task is to determine whether the clue already meets the threshold for dedicated on-site inspection. The second case is an industrial exhaust-gas evidence extraction task involving incomplete collection of cutting fumes, damaged dust-removal pipelines, and fugitive emissions. The gold standard contains 12 traceable evidence points and requires the model to extract them from inspection records, photographs, inquiry transcripts, and monitoring reports while preserving source attribution.

\par\noindent\begin{minipage}{\linewidth}
  \centering
  \includegraphics[width=\linewidth,height=0.72\textheight,keepaspectratio]{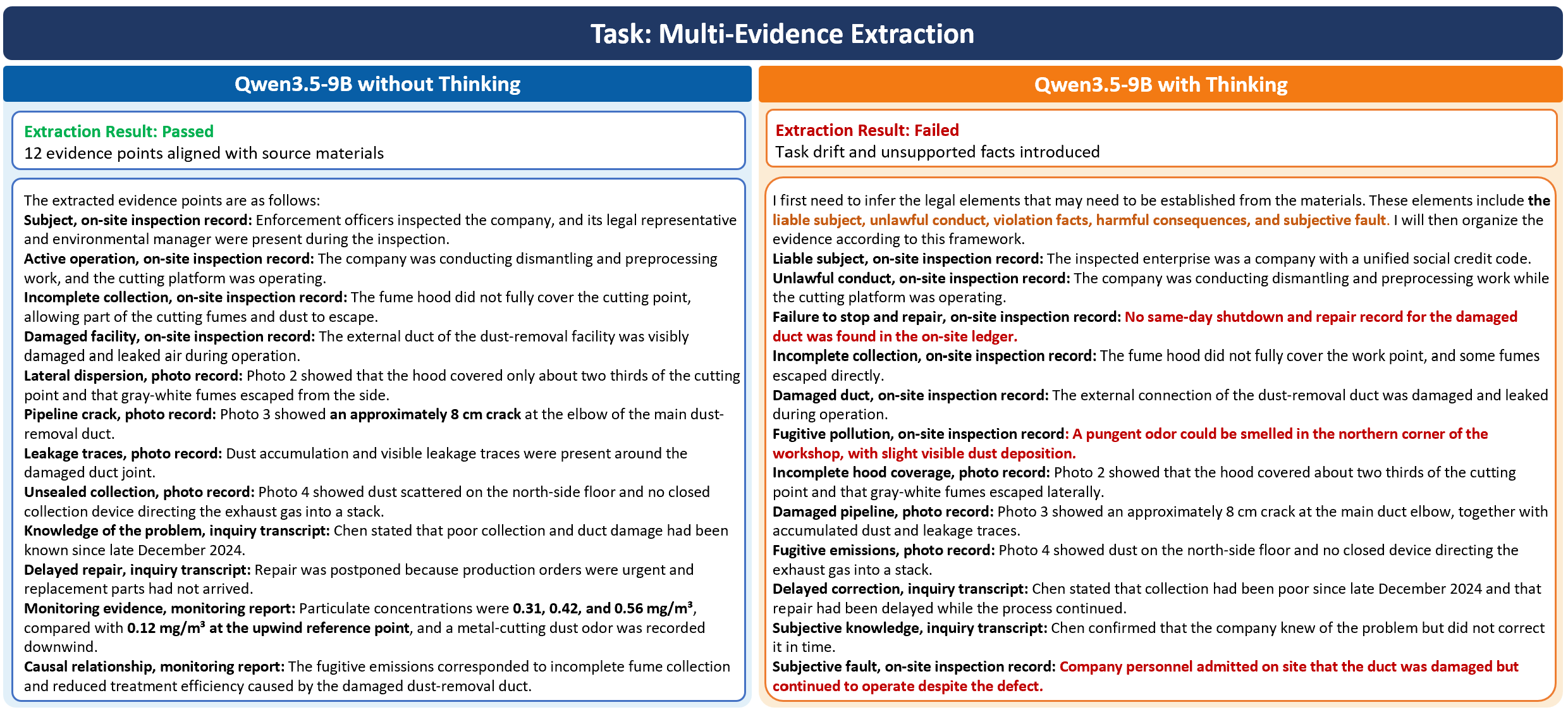}
  \captionof{figure}{Thinking-mode comparison for Qwen3.5-9B in industrial air-pollution evidence extraction.}
  \label{fig:appendix-e-6-thinking-mode-comparison-for-qwen3-5-9b}
\end{minipage}\par\vspace{0.65\baselineskip}

Two model outputs show different patterns. In the complaint verification case in Figure E.5, DeepSeek V4-Pro without Thinking incorrectly treats vague clues such as odor, crop wilting, and approximate location as actionable evidence, and directly recommends on-site source tracing. After enabling Thinking mode, the same model correctly identifies multi-step hearsay, missing subject, missing location, and missing discharge time or frequency, and instead recommends telephone follow-up, location supplementation, or grid-based screening. In the evidence extraction case in Figure E.6, Qwen3.5-9B passes the evaluation without Thinking mode, covering all 12 gold evidence points with no obvious irrelevant distortion. After Thinking mode is enabled, however, it recalls 11 points but reframes the task from evidence extraction into violation-constitution analysis, introducing unsupported claims such as a pungent odor in the northern corner of the workshop, absence of shutdown-maintenance records, and personnel admitting continued operation despite pipeline damage.

These two cases explain why Thinking mode must be task-adaptive. In complaint verification, the input is incomplete, but the judgment criteria are relatively clear: source reliability, element completeness, verifiability, and disposal pathway. Thinking helps the model make these implicit criteria explicit and prevents it from jumping from a weak risk signal to an enforcement action. In evidence extraction, however, the input is complex, multi-source, and weakly structured. Thinking amplifies the model's tendency toward self-explanation: instead of staying within the original materials, the model reconstructs a legal argument and treats elements that should be proven as if they were already recorded facts. Thus, reliable reasoning in environmental enforcement is not longer reasoning, but reasoning constrained by evidence boundaries, task objectives, and enforcement procedures. Thinking mode should therefore be selectively enabled for tasks that benefit from explicit judgment-chain organization, rather than used as a default setting.

\clearpage

\end{document}